\documentclass{article}

\usepackage{arxiv}

\usepackage[utf8]{inputenc} 
\usepackage[T1]{fontenc}    
\usepackage{hyperref}       
\usepackage{url}            
\usepackage{booktabs}       
\usepackage{amsfonts}       
\usepackage{nicefrac}       
\usepackage{microtype}      
\usepackage{lipsum}		
\usepackage{graphicx}
\usepackage{amsmath}
\usepackage{array}

\title{A Comprehensive Overview and Survey of Recent Advances in Meta-Learning}

\date{} 					

\author{ \href{https://orcid.org/0000-0001-7431-1619}{\includegraphics[scale=0.06]{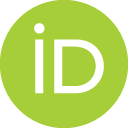}\hspace{1mm}Huimin Peng}\thanks{Thank you for all helpful comments! Feel free to leave a message about comments on this manuscript. In case I did not receive email, my personal email is \texttt{974630998@qq.com}. Thanks to github.com/kourgeorge/arxiv-style for this pdf latex template. I realize that this manuscript may be better named as recent advances in few-shot meta-learning.} \\
	\texttt{hpeng2@ncsu.edu} \\
	\texttt{peng.huimin.pennie@gmail.com} \\
}

\begin{document}
\maketitle

\begin{abstract}
This article reviews meta-learning also known as learning-to-learn which seeks rapid and accurate model adaptation to unseen tasks with applications in highly automated AI, few-shot learning, natural language processing and robotics. 
Unlike deep learning, meta-learning can be applied to few-shot high-dimensional datasets and considers further improving model generalization to unseen tasks. Deep learning is focused upon in-sample prediction and meta-learning concerns model adaptation for out-of-sample prediction. 
Meta-learning can continually perform self-improvement to achieve highly autonomous AI.
Meta-learning may serve as an additional generalization block complementary for original deep learning model. 
Meta-learning seeks adaptation of machine learning models to unseen tasks which are vastly different from trained tasks.
Meta-learning with coevolution between agent and environment provides solutions for complex tasks unsolvable by training from scratch.
Meta-learning methodology covers a wide range of great minds and thoughts. We briefly introduce meta-learning methodologies in the following categories: black-box meta-learning, metric-based meta-learning, layered meta-learning and Bayesian meta-learning framework.  
Recent applications concentrate upon the
integration of meta-learning with other machine learning framework to provide feasible integrated problem solutions. 
We briefly present recent meta-learning advances and discuss potential future research directions. 
\end{abstract}


\section{Introduction}
	\label{intro}
	
	\subsection{Background}
	\label{Background}

Meta-learning was first proposed by Jürgen Schmidhuber in 1987 \cite{Schmidthuber} which considers the interaction between agent and environment driving self-improvement in agent. It integrates evolutionary algorithms and Turing machine to achieve continual self-improvement so that agents adapt to dynamic environment conditions rapidly and precisely.
RNN (Recurrent Neural Network) \cite{Schmidhuber2015} 
contains self-connected recursive neurons that include memory cells to store model experiences and can be updated sequentially to model self-improvement. 
LSTM (Long Short-Term Memory) is a main form of RNN that can be used to account for long-range dependence in sequential data. 
Forget gates \cite{Gers1999} in LSTM are crucial to avoiding memory explosion and improving model generalization capability.
LSTM itself can be applied as a meta-learning system \cite{Younger2001} and parameters in LSTM can be updated to reflect upon base learner and meta-learners. LSTM can also be applied as a meta-learner \cite{Ravi2019a} under the framework of layered meta-learning to account for all base learners optimized with stochastic gradient descent.
Coevolution \cite{Clune2019} considers cooperation and competition between agents, and interaction between agents and evolving environments. 
Coevolution \cite{Wang2019a} generalizes deep model solutions to complex environments which cannot be solved by training from scratch. It demonstrates the necessity of meta-learning since meta-learning with coevolution is capable of providing solutions unreachable by training from scratch.
Most early developments in meta-learning are from meta-reinforcement learning seeking to improve degree of autonomy and smartness in robots. Training a reinforcement learning model from scratch is costly, and meta-learning saves the trouble by adapting a well-trained model from a similar task. Sometimes tasks are too complex and training from scratch does not provide a sufficiently predictive solution. However, it is mentioned in \cite{Wang2019a} and \cite{Clune2019} that meta-learning with generative evolution scheme provides good solutions to complex settings which cannot be solved through training from scratch. 

In \cite{Pan2010}, meta-learning and transfer learning are regarded as synonyms. Later developments in meta-learning and transfer learning diverge and they emerge as two different academic areas. 
Meta-learning is known as LTL (Learning-To-Learn) models which allow rapid and precise adaptation to unseen tasks \cite{Hutter2019}. 
Meta-learning is widely applied to few-shot datasets and adapts deep models for mining high-dimensional input from few-shot tasks.   
LTL gains attention under the popularity of continual learning research for strong AI. Transfer learning is concentrated upon adapting pre-trained deep model to solve similar tasks. In a sense, meta-learning seeks methodology with better adaptation to vastly different tasks by uncovering the deep similarity relation between tasks.
Recently there have been several research integrating domain adaptation into meta-learning to improve prediction accuracy in few-shot image classification task, as in \cite{Yu2018} and \cite{Guan2020}.
Similar to lifelong learning which accumulates knowledge and builds one model applicable to all, meta-learning aims to develop a general framework that can provide task-specific learners across vastly varying settings. 
Meta-learning mines fundamental rules and depends upon machine learning structures adaptable to a range of vastly different tasks. 
Meta-learning has two functionalities: achieving self-improvement along continual adaptations \cite{Schmidthuber} and providing  generalization framework for machine learning \cite{Finn2017}.

In recent developments, the most important application of meta-learning is few-shot learning \cite{Koch2015, Vinyals2016, Ravi2019a, Snell2017, Li2017, Sung2018, Gordon2018, Khodadadeh2018, Yu2018}. Actually most advances in few-shot learning lie in meta-learning, as mentioned in a survey paper of few-shot learning \cite{Wang2019b}. Most recently developed meta-learning methods in 2017-2019 focus upon few-shot image classification.
As we know, training of deep learning models demands a large amount of labelled data. For few-shot high-dimensional datasets, we resort to meta-learning which incorporates a generalization block based upon deep learners. By utilizing past model experiences, meta-learning provides  representation models of moderate complexity for few-shot tasks without demanding large amount of labelled data.
In image classification, few-shot learning is applicable to tasks where there are K (K$<10$) images in each image class known as N-class K-shot problems. It is observed that humans can learn new concepts from few demonstrations and utilize past knowledge to identify a new category efficiently.
The objective of strong AI is for machine learning models to approach the learning capabilities of human beings.
In few-shot learning, researchers seek human-like machines that can make fast and accurate classification based upon few images. Also for ancient languages where we only have few observations, and for dialects spoken by only a small group of people,
we resort to few-shot meta-learning for fast and accurate predictive modeling.
\cite{Wang2019b} provides an illustrative overview of meta-learning methods developed for few-shot learning.

Meta-learning models how human beings learn. Early research on meta-learning gains inspiration and motivation from cognitive science experiments explaining the process in which people learn new concept.
The primary component on learning how to solve different tasks is to model the similarity between tasks. 
In metric-based meta-learning, feature extraction and distance measure are jointly applied to represent similarity and are jointly tuned in the training process.
In layered meta-learning, base learner models task-specific characteristics and meta-learner models features shared by tasks. To represent similarity with appropriate complexity, hyperparameters implying the complexity of meta-learner can be tuned in training.
By modeling task similarity explicitly or implicitly, meta-learning introduces a flexible framework designed to be applicable to vastly different tasks. For example, MAML (Model-Agnostic Meta-Learning) \cite{Finn2017} is applicable to all learners that can be solved with SGD (Stochastic Gradient Descent). On the other hand,
based upon pre-trained deep models, meta-learning adapts task-specific learners to unseen tasks rapidly without precision loss \cite{Munkhdalai2018}. 
Both reinforcement learner \cite{Li2019} and neural network  \cite{Munkhdalai2017} have high representation capability and can be applied to search for an optimizer autonomously. In both cases, a general representation of an optimizer search space can be layed out explicitly.

In recent applications, meta-learning is focused upon integration with other machine learning frameworks and forms meta-reinforcement learning \cite{Schweighofer2003,  Jaafra2018, Gupta2018, Wang2018, Humplik2019, Rakelly2019, Dasgupta2019, Nagabandi2019} and meta-imitation learning
\cite{Duan2017, Yu2018} which are closely associated with robotics research.
The benefits of introducing meta-learning are twofold: to save computation by avoiding the need to re-train deep models from scratch; to increase reaction speed by adapting deep models to dynamic environment conditions.
Reinforcement learning estimates optimal actions based upon given policy and environment \cite{Finn2019b}. Imitation learning evaluates the reward function from observing behaviors of another agent in a similar environment \cite{Finn2017a}. Few-shot learning helps an agent make predictions based upon only few demonstrations from other agents \cite{Finn2017}. Recent application of meta-learning integrates reinforcement learning, imitation learning and few-shot meta-learning for robots to learn basic skills and react timely to rare situations \cite{Zhou2019a}.  
Meta-learning is feasible in tackling situations where out-of-distribution task predictive performance is required. Most deep learning models do not inherently embed any scheme like this and have to be integrated into a meta-learning framework to obtain this generalization capability to out-of-distribution situations.


A typical assumption behind meta-learning is that tasks share similarity so that they can be solved under the same meta-learning framework. 
To relax this assumption and improve model generalization, similarity model should be taken into consideration. 
Feature extraction model and distance measure can be jointly trained to represent similarity of any complexity. 
Meta-learner model specified as neural network also represents similarity shared by vastly different tasks. Similarity is an indisposable component in meta-learning whether it is implicitly or explicitly present.
In applications, statistical learning models such as convex linear classifiers and logistic regression are widely applied. 
Statistical models are less prone to over-fitting and can be designed to be robust to model misspecification. Deep learning models are intensively data-driven and highly integrated. 
Under the framework of meta-learning, we can combine statistical models and deep learning for fast and accurate adaptation. 
For few-shot datasets with high-dimensional input,
dimension reduction techniques in statistical learning can be specified as base learner and be integrated with meta-learner.
Base learner and meta-learner approach is proposed with the concept of meta-learning, both proposed in this paper \cite{Schmidthuber}.
In this survey paper, Base learner and meta-learner approach is called layered meta-learning, since it includes several layers from task-specific layer to task-generalization layer. 


The structure of our article is as follows. 
Section \ref{hist} presents history of meta-learning research. Section \ref{datasets} provides an outline of benchmark datasets for few-shot tasks and meta-learning formulation composed of meta-training, meta-validation and meta-testing.
Section \ref{meta-learning-models} summarizes meta-learning models into four categories. The distinction between methodologies is not exact. Different methods can be combined to solve application problems. 
Section \ref{Black-Box-Meta-Learning} presents black-box meta-learning methods that generalize deep learning models to unseen tasks.
Section \ref{Metric-Based-Meta-Learning} reviews metric-based meta-learning which utilizes feature extraction and distance measure jointly to represent similarity between tasks.
Section \ref{Layers-Meta-Learning} summarizes meta-learning models composed of task-specific base learner and meta-learners that explore similarity shared by vastly different tasks.
Section \ref{Bayesian-Meta-learning} surveys Bayesian meta-learning methods which integrate probabilistic methodology into meta-learning.
Section \ref{applications-meta-learning} brings attention to applications of meta-learning by integrating it with other machine learning frameworks.
Section \ref{meta-reinforcement} briefly reviews meta-reinforcement learning and section \ref{meta-imitation} surveys meta-imitation learning. Section \ref{online-meta-learning} briefly reviews online meta-learning and
section \ref{unsupervised-meta-learning} summarizes methods in unsupervised meta-learning.
Section \ref{Conclusion-and-Future-Research}  discusses future research directions based upon meta-learning.

\subsection{History}
\label{hist}


Jürgen Schmidhuber 
proposes meta-learning in his diploma thesis \cite{Schmidthuber} in 1987. \cite{Schmidthuber} achieves learning-to-learn with self-improvement using evolutionary algorithm and Turing machine. \cite{Schmidthuber} applies evolutionary algorithms to continually update learner so that it achieves higher-level of autonomy in AI. \cite{Schmidthuber} builds the first meta-learning framework upon inspirations from meta-cognition and introduces meta-hierarchy which constitutes base learner and many levels of meta-learners to represent task decomposition for a complex mission.
\cite{Schmidthuber} motivates many recent advances in meta-learning methodology. 
Later metareasoning proposed in \cite{Russell1991} lays out an integrated framework combining reinforcement learning, causal learning and meta-learning.
Metareasoning is the procedure for computers to autonomously make updated computation decision based upon causal reasoning under limited computation resources. Rather than evolutionary algorithm, self-improvement in \cite{Russell1991} relies upon causal reasoning which is goal-based and reward-driven.
\cite{Schmidhuber1994} lays out the framework for learning how to learn and proposes autonomous self-improvement in policy. The origin of meta-learning stems from policy self-improvement driven by interaction between agents and environment which dates back to \cite{Schmidthuber} and \cite{Schmidhuber1998}. 
Meta-learning does not necessarily involve the role of environment. For reinforcement learning tasks, reward-driven self-improvement in policy should be considered. For other complex missions, task decomposition is required to simplify the problem. 

Earlier developments in meta-learning research are concentrated upon hyper-parameter optimization, as in \cite{Schmidhuber1993, Caruana1995, Bengio2000, Jones2001, Stanley2002, Pellerin2004, Kordik2010, Bergstra2011,  Miranda2012, Bergstra2012, Maclaurin2015,AliEslami2016}. All machine learning algorithms can be framed as a base learner and all hyper-parameter optimization scheme can be seen as a meta-learner that guides, accelerates and contributes to the base learner optimization. Hyper-parameter optimization is a natural application of meta-learning.
In \cite{Caruana1995}, neural network is trained  using many distinct but related tasks simultaneously in order to improve model generalization capability. 
In \cite{Bengio2000}, gradients of a cross-validation loss function is used for joint optimization of several hyper-parameters. 
In \cite{Schweighofer2003}, meta-learning is integrated into reinforcement learning but only to tune hyper-parameters such as learning rate, exploration-exploitation tradeoff and discount factor of future reward. 

Later meta-learning can be applied to conduct autonomous model selection \cite{DeSouto2008, Lemke2010, Miranda2012, Leyva2015} such as neural architecture optimization \cite{Stanley2002, Kordik2010, Bosc2016, Jaafra2018, Baker2019}. 
Nowadays AutoML (Automated Machine Learning) is a hot topic where automatic search of a invincible deep neural network model is examined in various aspects. AutoML is expected to beat any human-designed deep neural model so that we (non-experts) do not need to go though the excruciating labor of hyper-parameter tuning ourselves. Recently AutoML also seeks integration with meta-learning in order to make the neural model search process more efficient, as in \cite{Real2020} and \cite{Liu2018c}.
In \cite{DeSouto2008}, meta-learning is applied to ranking and clustering, where algorithms are trained on meta-samples and the optimal model with highest prediction accuracy is selected. 
\cite{Lemke2010} considers using feedforward neural network, decision tree or support vector machine as learner model. Then it selects the class of models with the best performance on time series forecast.
In \cite{Miranda2012}, meta-learning is applied to select parameters in support vector machine (SVM), which demonstrates improved generalization capability in data modeling. Multi-objective particle swarm optimization is integrated with meta-learning to select parameters. 

One of the most common methodologies for meta-learning is learning how to learn by identifying the best optimizer. In deep learning applications, complex components which need high degree of representation power are modelled with deep neural networks. Searching through all neural network models can identify the optimal neural network for the complex components to bring out the best performance of deep learning models. For example, in deep reinforcement learning and deep imitation learning, a policy is modelled with deep neural network models,
since neural network model is known to be a good universal function approximator. \cite{Li2019} considers all first-order and second-order optimization methods under the framework of meta-imitation learning and minimizes the distance between predicted and target actions.
Policy update per iteration can be approximated using neural networks where weight parameters are estimated jointly with step direction and step size. 
By learning the optimizer autonomously, 
algorithms converge faster and outperform stochastic gradient descent. 


In recent developments, meta-learning is concentrated upon model adaptation between vastly different tasks which share certain similarity structure. Though in transfer learning, transferring between vastly different tasks may trigger negative transfer worsening model predictive performance.
For out-of-distribution tasks, we can extract the most similar experience from a large memory, and design predictive models based upon few-shot datasets collected in an unseen task. Similarity metric based end-to-end training approach is also mentioned in the early work of \cite{Schmidthuber}. \cite{Schmidthuber} indeed guides the development of meta-learning ever since and contains the essential philosophy behind meta-learning methodology.
Classification of recent meta-learning methodologies is not exact since meta-learning models exhibit the tendency to be more flexible and integrated in recent developments. But with classification, we can roughly outline recent research directions for later mixing. 
Recent meta-learning methodology can be categorized into four classes which are model-based, metric-based, optimization-based  and generative AI-GAs methods, as in \cite{Weng2018,Clune2019}. 
MANN (Memory-augmented neural networks) \cite{Santoro2016a} belongs to the model-based category. It stores all model training history in an external memory and loads the most relevant model parameters from external memory every time a new unseen task is present. 
Task similarity is reflected in the metric-based relevance of memory items extracted for model adaptation.
Second,
convolutional Siamese neural network \cite{Koch2015} is within the metric-based category. Metric refers to the similarity between tasks. Siamese network designs a metric that depends upon the similarity measure between convolutional features of different images. Matching networks \cite{Vinyals2016}, relation network \cite{Sung2018} and prototypical network \cite{Snell2017} are all metric-based methods.
Feature extraction mechanism and distance measure of embeddings jointly model task similarity.
In our work, we categorize memory-based and metric-based methods into metric-based meta-learning.

Optimization-based technique includes a learner for model estimation at task level and a meta-learner for model generalization across tasks. 
Meta-learners may consist of many layers of models for different communication patterns and generalization scopes according to task decomposition structure.
Many layers of meta-learners may collapse into one layer of meta-learner depending upon task complexity and the corresponding model complexity.
In our work, we summarize methods containing base learner and meta-learners to be layered meta-learning.
In meta-LSTM \cite{Ravi2019a}, a meta-learner updates learner parameters on different batches of training data and validation data. For learners optimized with stochastic gradient descent, a meta-learner can be specified to be a long short-term memory model \cite{Younger2001}.
MAML (Model-Agnostic Meta-Learning) proposed by Chelsea Finn in 2017 \cite{Finn2017} is famous and applicable to many real-life applications. MAML provides fast and accurate generalization of deep neural network models and does not impose any model assumption. MAML is applicable to any learner model optimized with stochastic gradient descent. 
First-order meta-learning algorithms FOMAML and Reptile \cite{Nichol2018} are also optimization-based, where iterative updates on parameters are designed to be the difference between previous estimates and new sample average estimate. 

From probabilistic perspective, meta-learning can be formulated under Bayesian inference framework. Bayesian inference provides an efficient estimation of the uncertainty in few-shot meta prediction and extends methodology to be more widely applicable. Bayesian approach is statistical and has a long history. For complex tasks, resorting to Bayesian procedure is worthwhile.
In \cite{Lake2015}, a Bayesian generative model is combined with deep Siamese convolutional network to make classification on hand-written characters. Bayesian generative model is very intriguing since it simulates samples that look real to augment training and validation data. The sample-generating scheme is estimated from training data itself. So no additional information is generated from generating more samples. However, augmenting data this way greatly improves prediction accuracy in few-shot image classification.
In \cite{Yoon2018a}, a Bayesian extension of MAML is proposed, where stochastic gradient descent (SGD) in MAML is replaced with Stein variational gradient descent (SVGD). SVGD offers an efficient combination of MCMC and variational inference, which are two main approaches in Bayesian approach.  Combination of MCMC and variational inference is ideal since the advantages and disadvantages of MCMC and variational inference complement each other.
In \cite{Gordon2018}, a Bayesian graphical model is embedded in the task-specific parameters and meta-parameters in meta-learning framework. An amortization network is used to map training data to weights in linear classifier. Amortization network is also used to map input data to task-specific stochastic parameter for further sampling. It utilizes an end-to-end stochastic training to compute approximate posterior distributions of task-specific parameters in meta-learner and unknown labels on new tasks.


One of the recent meta-learning applications is in robotics, where meta-imitation learning \cite{Finn2017a, Duan2017, Yu2018, Zhou2019a} and meta-reinforcement learning \cite{Duan2016, Wang2016, Levine2017, Gupta2018, Stadie2018, Nagabandi2019a, Rakelly2019, Lee2019a} are of primary interest. 
The objective of general AI is for machine learning to reach human intelligence so that machine learning can handle dangerous tasks for human. 
Children is capable of learning basic movements from only one or two demonstrations so that researchers hope robots can do the same through meta-imitation learning. 
Imitation of action, reward and policy is achieved by minimization of regret function which measures the distance between current state and imitation target. 
In \cite{Finn2017a}, MAML for one-shot imitation learning is proposed. Minimization of cloning loss leads to closely minimick target action that robots try to follow. It estimates a policy function that maps visual inputs to actions. Imitation learning is more efficient than reinforcement learning from scratch and is popular in robotics research. 
\cite{Yu2018} also integrates MAML into one-shot imitation learning. It collects one human demonstration video and one robot demonstration video for robots to imitate. The objective here is to minimize behavioral cloning loss with inner MAML parameter adaptation. It also considers domain adaptation with generalization to different objects or environment in the imitation task.

Meta-reinforcement learning (Meta-RL) is designed for RL tasks such as reward-driven situations with sparse reward, sequential decision and clear task definition \cite{Wang2018}. 
RL considers the interaction between agent and environment through policy and reward. By maximizing reward, robots select an optimal sequential decision.
In robotics, meta-RL is applied in cases where robots need rapid reaction to rare situations based upon previous experiences. 
\cite{Wang2016} provides an overview of meta-RL models in multi-bandit problems. Meta-learned RL models demonstrate better performance than RL models from scratch. 
\cite{Rakelly2019} constructs a highly integrated meta-RL method PEARL which combines variational inference and latent context embedding in off-policy meta-RL. 
In addition, reward-driven neuro-activities in animals can be explained with meta-RL. In \cite{Wang2018}, phasic dopamine (DA) release is viewed as reward and meta-RL model explains well the DA regulations in guiding animal behaviors with respect to the changing environment in animal experiment. 

In addition to meta-RL and meta-imitation learning, meta-learning can be flexibly combined with  machine learning models for applications in real-life problems. For example, unsupervised meta-learning conducts rapid model adaptation using unlabelled data. Online meta-learning analyzes streaming data and performs real-time model adaptation. The feature of meta-learning application is small sample size and high dimensional input. In meta-RL, number of trajectories for interaction between agent and environment is small. In meta imitation learning, number of demonstrations from human or agent for a similar task is only one or two, thus small sample size. 
First, unsupervised meta-learning \cite{Bengio2011, Gupta2018, Khodadadeh2018, Garg2018, Hsu2019, Metz2019} is for modelling unlabelled data. Unsupervised clustering methods such as adversarially constrained autoencoder interpolation (ACAI) \cite{Berthelot2019}, bidirectional GAN (BiGAN) \cite{Berthelot2019}, DeepCluster \cite{Caron2018} and InfoGAN \cite{Chen2016} are applied to cluster data and estimate data labels \cite{Hsu2019}. Afterwards meta-learning methods are used on unlabelled data and predicted labels obtained through unsupervised clustering. It is mentioned in \cite{Hsu2019} that unsupervised meta-learning may perform better than supervised meta-learning.
Another combination of unsupervised learning and meta-learning is in \cite{Metz2019}. It replaces supervised parameter update in inner loop with unsupervised update using unlabelled data. Meta-learner in the outer loop applies supervised learning using labeled data to update the unsupervised weight update rule.
It demonstrates that this unique combination performs better in model generalization.

Second, online meta-learning analyzes streaming data so that the model should respond to changing conditions rapidly using a small batch of data in each model adaptation
\cite{Harrison2018, Nagabandi2019a, Finn2019}. Robots are supposed to react real-time to dynamic environment so that robots should learn to update deep model each time with real-time obtained data, which is of small sample size. 
\cite{Harrison2018} proposes a Bayesian online learning model ALPaCA where kernel-based Gaussian process (GP) regression is performed on the last layer of neural network for fast adaptation. It trains an offline model to estimate GP regression parameters which stay fixed through all online model adaptation.  
\cite{Nagabandi2019a} applies MAML to continually update the task-specific parameter in prior distribution so that the Bayesian online model adapts rapidly to streaming data. 
\cite{Finn2019} integrates MAML into an online algorithm follow the leader (FTL) and creates an online meta-learning method follow the meta-leader (FTML). MAML updates meta-parameters which are inputs into FTL and this integrated online algorithm generalizes better than previously developed methods. 

Meta-learning algorithms are hybrid, flexible and can be combined with machine learning models such as Bayesian deep learning, RL, imitation learning, online algorithms, unsupervised learning and graph models. In these combinations, meta-learning adds a model generalization module to existing machine learning methods. Many real-life applications require deep neural network models and integrating meta-learning methods such as MAML, Reptile and Prototypical Nets into existing deep models can bring additional benefits.
Integration of MAML into deep models is convenient since all deep neural network models use SGD to update weight parameters. 
For complex tasks, we can decompose complex tasks into simpler sub-tasks which can be solved with base learners of less complexity.
We may utilize complex meta-learner to represent complex similarity between vastly different tasks. 
AI-GAs \cite{Clune2019} contain coevolution between agents and environment and reach feasible solutions for very complex tasks which are otherwise unsolvable by training from scratch. I like the integration of meta-learning methodology and high-dimensional research in statistics. Recent developments in meta-learning are focused upon small sample with high dimensional input. High-dimensional research in statistics handles data where number of features is greater than sample size, ie small sample with high dimensional input. They approach the same kind of application problems from different perspectives and more integration is expected in this intersection area.

\subsection{Datasets and Formulation}
\label{datasets}

Few-shot datasets used as benchmarks for performance comparison in meta-learning literature are reviewed in \cite{Triantafillou2019a}. Many meta-datasets are available at \url{https://github.com/google-research/meta-dataset}. Though most datasets are large and annotated, the number of few-shot meta-learning datasets is constantly growing. Still, meta-learning datasets are not as mainstream as large annotated high-dimensional dataset, where deeper and more complex neural network model is expected in pursuit of higher predictive accuracy. As for application, recent advances in meta-learning are concentrated upon few-shot high-dimensional data. I feel that meta-learning datasets are abundant in real life and there will be many more challenging baseline datasets available in the future. For general AI application, most datasets are either for vision or for language. Currently commonly used meta-learning datasets are briefly listed as follows.

\begin{itemize}
	
	\item  \textbf{Omniglot} \cite{Lake2011a} is available at \url{https://github.com/brendenlake/omniglot}.
	Omniglot is a large dataset of hand-written characters with 1623 characters and 20 examples for each character. These characters are collected based upon 50 alphabets from different countries. It contains both images and strokes data. Stroke data are coordinates with time in miliseconds. 
	
	\item  \textbf{ImageNet} \cite{Russakovsky2015}  is available at \url{http://www.image-net.org/}.
	ImageNet contains 14 million images and 22 thousand classes for these images. Large scale visual recognition challenge 2012 (ILSVRC2012) dataset is a subset of ImageNet. It contains 1,281,167 images and labels in training data, 50,000 images and labels in validation data, and 100,000 images in testing data. 
	
	\item 
	\textbf{\textit{mini}ImageNet} \cite{Vinyals2016, Lee2019} is a subset of ILSVRC2012. It contains 60,000 images which are of size 84$\times$84. There are 100 classes and 600 images within each class. \cite{Ren2018} splits 64 classes as training data, 16 as validation data, and 20 as testing data.
	
	\item \textbf{\textit{tiered}ImageNet} \cite{Ren2018} is also a subset of ILSVRC2012 with 34 classes and 10-30 sub-classes within each. It splits 20 classes as training data, 6 as validation data and 8 as testing data.
	
	\item  \textbf{CIFAR-10/CIFAR-100} \cite{Krizhevsky2010, Oreshkin2018}  is available at \url{https://www.cs.toronto.edu/~kriz/cifar.html}. CIFAR-10 contains 60,000 colored images which are of size 32$\times$32. There are 10 classes, each contains 6,000 images. CIFAR-100 contains 100 classes, each includes 600 images. \textbf{CIAFR-FS} \cite{Bertinetto2019}
	is randomly sampled from CIFAR-100 for few-shot learning in the same mechanism as miniImageNet. 
	\textbf{FC100} \cite{Oreshkin2018} is also a few-shot subset of CIFAR-100. It splits 12 superclasses as training data, 5 superclasses as validation data and 5 superclasses as testing data.
	
	\item  \textbf{Penn Treebank (PTB)} \cite{Santorini1993}  is available at \url{https://catalog.ldc.upenn.edu/LDC99T42}.
	PTB is a large dataset of over 4.5 million American English words, which contain part-of-speech (POS) annotations. Over half of all words have been given syntactic tags. It is used for sentiment analysis and classification of words, sentences and documents.
	
	\item \textbf{CUB-200} \cite{Triantafillou2019a, Welinder2010}  is available at \url{http://www.vision.caltech.edu/visipedia/CUB-200.html}. CUB-200 is an annotated image dataset that contains 200 bird species, a rough image segmentation and image attributes.

	\item \textbf{CelebA} (CelebFaces Attributes Dataset) is available at
	\url{http://mmlab.ie.cuhk.edu.hk/projects/CelebA.html}.
	CelebA is an open-source facial image dataset that contains 200,000 images, each with 40 attributes including identities, locations and facial expressions. 
	
	\item \textbf{YouTube Faces} database is available at
	\url{https://www.cs.tau.ac.il/~wolf/ytfaces/}.
	YouTube Faces contains 3,425 face videos from 1,595 different individuals. Number of frames in each video clip varies from 48 to 6070.
	
\end{itemize}

Among these datasets,  
\textbf{\textit{mini}ImageNet} \cite{Vinyals2016, Lee2019} and
\textbf{\textit{tiered}ImageNet} \cite{Ren2018} are the most useful few-shot image classification datasets. They are used to compare predictive classification accuracy of many meta-learning few-shot methods. The application of meta-learning is not limited to few-shot learning. These meta-learning methodologies may serve as useful ingredients for constructing highly integrated models to suit the need of real-life complex missions. For few-shot image classification, datasets listed here are widely applied in literature as comparison benchmarks.

\begin{figure}[htpb]
	\centering
	\includegraphics[width=3in]{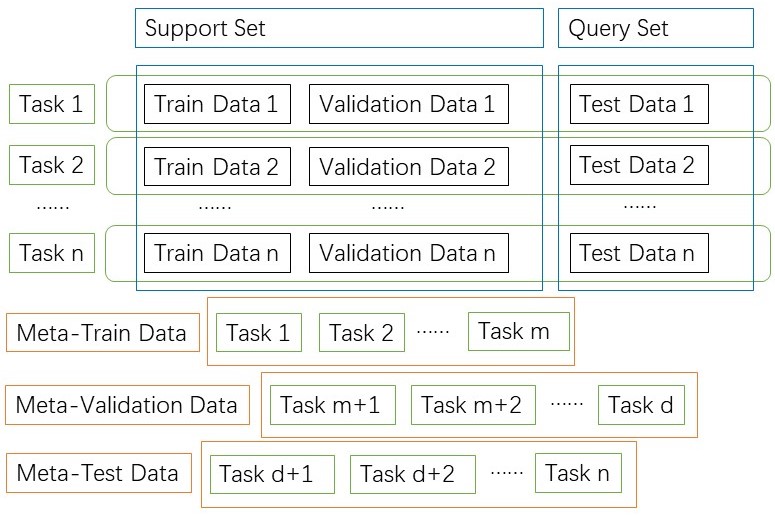}
	\caption{Upper part shows data in each task: training data, validation data and testing data. Lower part shows meta-training data, meta-validation data and meta-testing data that consists of tasks. Support set is the set of all labelled data. Query set is the set of all unlabelled data. Due to limited square spaces, I wrote 'train data' instead of 'training data' and 'test data' instead of 'testing data'.}
	\label{fig1}
\end{figure}

Training and testing framework for few-shot image classification used in meta-learning are outlined in figure \ref{fig1}. This meta-learning framework is after episodic training used in recent few-shot meta-learning research papers. Task data are randomly sampled from large annotated dataset. For few-shot image classification, image classes are sampled first and images within each class are sampled afterwards. On each task, task-specific parameters are updated. After training one or several tasks, meta-parameters are updated. Meta-parameters depict the common feature shared across several tasks.
Within each task, there are training data $\mathcal{D}^{tr}$, validation data $\mathcal{D}^{val}$ and testing data $\mathcal{D}^{test}$. 
Support set $\mathcal{S}$ is the set of all labelled data. Training data and validation data are randomly sampled from support set. Training data are used to update task-specific parameters by minimizing loss function on training data. Validation data are used to update meta-parameters by minimizing loss function on validation data. Loss function on validation data measures generalization capability of deep model. Minimization of loss function on validation data maximizes generalization capability of deep models.
Query set $\mathcal{Q}$ is the set of all unlabelled data and test data are randomly sampled from query set. Test data are unlabelled and we can view the predictive results on test data to see whether the results match intuition or not.
Meta-learning datasets include non-overlapping meta-training data $\mathcal{D}_{meta-train}$, meta-validation data $\mathcal{D}_{meta-val}$ and meta-testing data $\mathcal{D}_{meta-test}$ which consists of tasks. As in the case of few-shot image classification, the image classes in training data and validation data are non-overlapping, so that validation data can be viewed as an unseen task compared with training data. Loss function on validation data measures model capability to solve unseen tasks. Within each task, there is a training dataset, a validation dataset and a testing dataset. For few-shot tasks, the sample size in training dataset is small.

In supervised meta-learning, input is labelled data $(\pmb{x},y)$, where $\pmb{x}$ is an image or a feature embedding vector and $y$ is a label.  
Data model is $y=h_{\theta}(\pmb{x})$ parameterized by meta-parameter $\theta$.
As in \cite{Finn2019b}, a task is defined as 
\[
\mathcal{T}=\{p(\pmb{x}),p(y|\pmb{x}),\mathcal{L}\},
\] 
where $\mathcal{L}$ is a loss function, $p(\pmb{x})$ and $p(y|\pmb{x})$ are data-generating distributions of inputs and labels respectively. 
Task follows a task distribution $\mathcal{T}\sim p(\mathcal{T})$.
K-shot N-class learning is a typical problem setting in few-shot image classification, where there are N image classes each with K examples and K$<10$.

\section{Meta-learning}
\label{meta-learning-models}

For fast and accurate adaptation to unseen tasks with meta-learning, we need to balance exploration and exploitation. 
In exploration, we define a complete model search space which covers all algorithms for the task. In exploitation, we optimize over the search space, identify the optimal learner and estimate learner parameter. 
For example, learning-optimizers methods proposed in \cite{Li2019, Wichrowska2017} define an extensive search space of optimizers for model exploration.
In \cite{Triantafillou2017}, mean average precision defined as the precision in predicted similarity is a proposed loss function used for model exploitation. 
More exploitation is necessary when data distribution is multi-modal distribution, fat-tailed distribution, highly skewed distribution or others. Efficient exploration strategies are required when the computational cost is high. We may apply meta-learners to predict performance, provide guidance for efficient exploration and to exclude areas where superior performance is unlikely, as in \cite{Domhan2015} and \cite{Baker2018}. 

On the other hand, meta-learning models can combine offline deep learning and online algorithms. 
In offline deep model, we aggregate past experiences by training a deep model on large historical datasets. 
In online algorithms, we continually adapt a deep offline model to conduct predictive analysis on few-shot datasets from novel tasks.
For instance, memory-based meta-learning model in \cite{Kaiser2019} stores offline training results in memory so that they can be retrieved efficiently in online model adaptation. 
Online Bayesian regression in \cite{Harrison2018}
uses offline training results to initiate task-specific parameters in prior distributions and update these parameters continually for rapid adaptation to online streaming data.
Based upon pre-trained deep learning models, meta-learning methods adapt to new unseen tasks efficiently. 

A typical assumption behind meta-learning is that tasks share similarity structure, and model generalization between tasks can be performed efficiently. Degree of similarity between tasks depends upon the similarity metric function which is an important component in the meta-learning objective function. 
Similarity is explicitly or implicitly incorporated in meta-learning models and similarity of any complexity should be allowed to achieve model generalization between different tasks.  
Reliable adaptation between different tasks relies upon identifying the similarity structure between them. In meta-learning research, primary interest lies in relaxing the requirements upon degree of similarity between tasks and improving model adaptivity. Similarity is difficult to measure. First, we should know which part of one task is related to which part of another task and align the tasks. Second, we should know how the parts of different tasks are related to each other and estimate a representational model between parts.

In this section, we briefly summarize meta-learning frameworks that emerge in recent literature into the following categories: black-box meta-learning \cite{Finn2019b}, metric-based meta-learning, layered meta-learning and Bayesian meta-learning. 
This classification of meta-learning frameworks is not exact and the boundaries are vague between different classes of methodologies. This paper briefly summarizes research directions in meta-learning methods. Every method is a combination of several meta-learning routes mentioned in this section. There is no clear boundary between these methods which classifies them into different methodology lines. Components of one method belong to several different research lines. We review these methods to summarize how these meta-learning routes are integrated into the real meta-learning methods.

Black-box meta-learning utilizes neural network as the main focus to improve model generalization capability:
\begin{itemize}
	\item Apply neural network to model policy and policy self-improvement. Usually policy and policy self-improvement are modelled with neural network. Neural network represents all kinds of policy update strategies including stochastic gradient descent (SGD). An optimal policy update strategy from extensive search should outperform SGD. Similarity between tasks is implicitly considered in black-box neural network.
	\item Activation to parameter \cite{Qiao2018} generalizes to unseen tasks by only updating the mapping between activation to weight parameters in output layers. This mapping captures the relation between feature to classifier output. Parameters in the feature extraction module remain fixed. Predictive accuracy on trained tasks does not decrease after model adaptation to unseen tasks.
	\item AdaResNet and AdaCNN \cite{Munkhdalai2018}
	re-design neurons in ResNet and CNN to include task-specific parameters in activation or weights. 
	Meta-parameters are for representation of the similarity shared by tasks and task-specific parameters are for representation of task-specific information.
	In generalization to unseen tasks, only task-specific parameters are updated to reflect upon task data. Memory module stores task features and parameter estimates to accelerate the adaptation of task-specific parameters.
	\item Meta-learning is widely applicable in the area of AutoML to accelerate the search for an optimal neural network model. For different hyper-parameter combinations, meta-learner can predict model performance based upon previous training experiences. Meta-learner points base learner (AutoML) to search through only promising regions in hyper-parameter surface. Besides, for an unseen task, meta-learner can provide a good initialization for model selection and parameter specification. 
\end{itemize}

Metric-based meta-learning relies upon the similarity measure between unseen tasks and trained tasks to find the most relevant past model training experiences to refer to:
\begin{itemize}
	\item SNAIL \cite{Mishra2018} includes an attention block which identifies the most relevant item from memory. The attention block measures the similarity shared by tasks and contains only meta-parameters. Attention blocks typically consist of two parts: feature extractor neural network and deep metric neural network measuring task similarity. Meta-parameters are in feature extractor blocks and supervised deep metric learning is utilized to train metric parameters.
	\item Relation Network \cite{Sung2018}
	contains meta-parameters in feature extraction neural network and meta-parameters in distance metrics. Feature extraction model and distance measure are jointly tuned to represent similarity between tasks. These metric-based meta-learning methods are very similar in structure but include different degrees of flexibility in training. 
	\item Prototypical Network \cite{Snell2017} measures distance between extracted feature embeddings and these centroids from different classes. There are many methods built upon Prototypical Network which show superior performance in few-shot image classification, as in \cite{Guan2020} and \cite{Li2020}. \cite{Guan2020} integrates domain adaptation into Prototypical Network and shifts and scales class centroids in adaptation to unseen tasks. \cite{Li2020} utilizes adaptive margin loss in deep metric learning of Prototypical Network to improve classification power.
	\item Dynamic few-shot \cite{Gidaris2018}
	integrates a ConvNet-based classifier, a few-shot classification weight generator and a classifier based upon cosine distance metric.
	Classification weight remains fixed across all seen classes and can be generalized to unseen tasks. For unseen classes, simply add extra neurons to the output layer and use the few-shot classification weight generator to estimate additional parameters. Feature extraction neural network and classification weight generator for testing data jointly represent similarity shared by different tasks. Complexity of this framework may be tuned to match the complexity of real similarity between tasks.
	\item MAP \cite{Triantafillou2017} (Mean Average Precision)
	designs a similarity-ranking based measure as the meta-objective function to estimate meta-parameters. It measures the difference between predicted similarity and real similarity between tasks. For each individual sample in a batch of data, the other samples in the same batch are ranked by predicted similarity with this individual sample. When processing each individual sample, relevant model experiences from relevant samples are referred to. In mAP, memory module scheme is used much more often than other methods, since memory module is only consulted on the unit of tasks not on the unit of sample points.
\end{itemize}

Layered meta-learning relies upon task decomposition to construct generalization tree and consists of base learner and meta-learners for different generalization scopes:
\begin{itemize}
	\item MAML \cite{Finn2017} is the most popular methods within GBML (Gradient-based meta-learning) methods. Meta-parameter is the initial value of task-specific parameter on each task. Meta-parameter is updated in meta-learner by minimizing loss function on validation data. Task-specific parameter is updated in base learner by minimizing loss function on training data. MAML and GBML are compatible with all learners optimized using stochastic gradient descent. Most deep neural networks can be optimized through gradient back-propagation and can use MAML or GBML to improve model generalization capability. Since MAML includes SGD update in both base learner and meta-learner, MAML introduces second-order derivative of loss function which costs computational time. Later Reptile \cite{Nichol2018a} and FOMAML \cite{Nichol2018} are proposed as first-order approximation of MAML. Reptile and FOMAML are faster but less accurate than MAML.
	\item Meta-LSTM \cite{Ravi2019a} uses LSTM as meta-learner and is also compatible with any learner optimized by SGD. LSTM is one of the most important forms of RNN. LSTM contains memory cells, self-connected recursive neurons, multiplicative gates and forget gates. LSTM ensures that gradients do not explode or vanish in back-propagation and can account for long-range dependence in sequential data. Memory cells contain critical past model training experiences and forget gates ensure that memory do not explode. LSTM can model task similarity of any complexity and self-connected recursive neurons allow self-improvement along the meta-training process. There is correspondence between LSTM cell calculations and stochastic gradient descent. Specifying gates in LSTM properly, calculations in LSTM cell are exactly stochastic gradient descent.
	\item R2-D2 and LR-D2 \cite{Bertinetto2019}
	considers using widely applicable base learners such as ridge regression and logistic regression. These base learners are of less complexity than deep models, but they are also widely applied in real-life applications. 
	Meta-learner is specified to be neural network model that has high representation power for task similarity of any complexity. Base learner of lower complexity and Meta-learner of high complexity is an efficient way to specify layered meta-learning model. Base learner is fast to train on each few-shot task and meta-learner slowly uncovers shared features of all tasks to update meta-parameter.
	\item TPN \cite{Liu2019}
	applies end-to-end transductive learning between meta-training data and meta-testing data. Feature extraction model and label propagation model based upon graph are jointly trained to reflect the similarity between unseen tasks and trained tasks.
	Transductive label propagation model based upon graph is regarded as the meta-learner. Data variance is taken into account when modelling similarity metric between samples. Meta-parameters are parameters in feature extraction module and parameters in data variance model. Label propagation for each task predicts labels jointly for all input data using an explicit solution formula, which constitutes a very efficient base learner.
	\item LEO \cite{Rusu2019}
	utilizes a dimension reduction encoder network, a relation network and a decoder function. Encoder network performs feature extraction, relation network measures similarity between samples with data variance accounted, decoder network projects features to classifier. For generalization to unseen tasks, parameters in all these modules are updated with MAML. 
	Meta-learner provides initial values for these parameters in base learner during generalization to unseen tasks.  
	In the inner loop, base learner considers task training data and updates these parameters from initial values supplied from meta-learner. 
\end{itemize}

Bayesian meta-learning recasts meta-learning to be in the probabilistic framework and integrates Bayesian great minds thoughts into meta-learning:
\begin{itemize}
	\item Bayesian program learning \cite{Lake2015} generates hand-written characters indistinguishable from real hand writings using visual checks. Joint distribution of character types, image-specific parameters and images are utilized to simulate data indistinguishable from real data. Simulated data is augmentation to the original few-shot dataset and contributes to improvement in predictive accuracy in few-shot learning. Joint posterior distribution of image-specific parameters and character types is updated during model generalization to unseen tasks. 
	\item AI-GAs \cite{Clune2019} consist of meta-learning architecture, learner algorithms and generation of more complex environment. AI-GAs consider coevolution between agents and environment. AI-GAs provide solutions for complex environments which are otherwise unsolvable by training from scratch. AI-GAs contains important interaction between agent and environment. Within evolution, as environment becomes more complex, so does the solver agent. Each small change in environment corresponds to each small improvement in solver agent. Evolution of environments guides solver agent to grow and to tackle complex tasks which cannot be solved by tranining from scratch. AI-GAs are optimization techniques which use meta-learner to guide base learner to find optimum closer to global optimum. 
	\item Neural Statistician \cite{Edwards2016}
	applies variational autoencoder (VAE) to approximate the posterior distribution of task-specific context. Given posterior distribution, context corresponding to highest posterior probability is the predicted label of input data. From posterior distribution, we can also obtain confidence interval of unknow label on input data. VAE is an efficient Bayesian inference tool which not only provides estimation on unknown quantity but also the uncertainty in predicted value.
	\item LLAMA \cite{Grant2018}
	recasts MAML as probabilistic inference in a hierarchical Bayesian model. 
	It uses first-order and second-order Laplace approximation to Gaussian distribution in log likelihood function. For other data distributions such as multi-modal distribution, highly skewed distribution or fat-tailed distribution, Laplace approximation does not work well. Laplace approximation shows best results for tightly distributed, symmetric distributions such as Gaussian distrbution. All distributions in Bayesian framework are taken to be Gaussian so that log likelihood function can be well approximated by Laplace approximation, which uses point estimation at the mode to approximate integral of Gaussian distribution.
	\item BMAML \cite{Yoon2018a}
	replaces SGD (Stochastic gradient descent) in MAML with SVGD (Stein variational gradient descent), where SVGD is an efficient Bayesian inference method combining MCMC and variational inference. SVGD does not only work for all learners optimized with SGD. It uses chaser loss based upon SVGD to estimate meta-parameter and Bayesian fast adaptation to update task-specific parameters. 
	\item VERSA \cite{Gordon2018}
	relies upon distributional Bayesian decision theory for amortized variational inference. Posterior distribution of true label is approximated and updated during model generalization to unseen tasks. 
	\item Under Bayesian framework, not only task-specific parameters and meta-parameters are regarded as random variables, task can also be viewed as a random variable. Task can be considered as countably infinitely many and Chinese restaurant process is used as the prior distribution for task. Posterior distribution of task is continually updated and maximum posterior estimate of task is provided in the model generalization process. All task-specific parameters are taken to be functions of task. Estimation and uncertainty measure for task-specific parameter are derived from posterior distribution of task. Meta-parameters should be updated using aggregation of all task posterior distributions.
\end{itemize}




\subsection{Black-Box Meta-Learning}
\label{Black-Box-Meta-Learning}


Hyperparameter optimization can be achieved through random grid search or manual search
\cite{Bergstra2012}. Model search space is usually indexed by hyperparameters \cite{Li2019}. In adaptation to novel tasks, hyperparameters are re-optimized using data from the novel task. Optimizers can be approximated with neural networks or reinforcement learners \cite{Wichrowska2017}. 
Neural networks can approximate any function with good convergence results. By using neural networks, the optimizer search space represents a wide range of functions that guarantee better potential optima.
For neural network with varying architectures, hyperparameters include number of layers, number of block modules, number of neurons etc. These hyperparameters are discrete and optimization techniques are different from continuous hyperparameters such as learning rate, momentum etc. Bayesian optimization, reinforcement learning optimization, genetic programming optimization may be employed to optimize both discrete and continuous hyperparameters in deep neural network model.

In \cite{Li2019}, optimization is through guided policy search and neural network is used to model policy update strategy which is usually stochastic gradient descent. This algorithm searches for the optimal policy update strategy represented by neural network instead of using SGD directly.
Policy update is formulated as
\begin{eqnarray}
&\Delta x\leftarrow\pi(f,\{x_0,\cdots,x_{i-1}\}),
\nonumber
\end{eqnarray}
where $f$ is a neural network to model policy update strategy $\pi$ and $\Delta x$, $\gamma$ is the step size. In SGD, policy update is through gradients $\Delta x=-\gamma \sum_{j=0}^{i-1}\alpha^{j}\bigtriangledown_{x_{j}} f(x_{j})$, where $\alpha$ is a discount factor.
In this case, policy update is approximated with a neural network and is continually adapted using task data. 

\begin{figure}[htpb]
	\centering
	\includegraphics[width=3in]{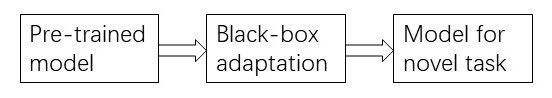}
	\caption{Black-box adaptation. Pre-trained model is an offline deep learning model on a large historical dataset. Black-box adaptation is applied to adapt the pre-trained deep model to a lightweight novel task.}
	\label{blackbox}
\end{figure}

Another approach is the adaptation of a pre-trained neural network from offline deep model to unseen tasks, as in figure \ref{blackbox}. It is commonly used in transfer learning. From this perspective, meta-learning and transfer learning are the same concept.
For few-shot unseen tasks, training data is limited so that black-box adaptation is only for a small portion of all parameters in neural network model.
For similar few-shot image classification tasks, feature extraction modules remain intact and only classifier parameters are adapted for novel tasks.
For unseen tasks with sufficiently large annotated dataset, black-box adaptation is for almost all parameters, hyperparameters, and even neural network architecture depending on the amount of labelled new data available.
For similar tasks, solution models are also similar, therefore black-box adaptation does not need to be for all parameters in the deep model. Only updating task-specific parameters is sufficient for adaptation to few-shot unseen tasks. For vastly different tasks, solution models are also vastly different. Black-box adaptation should cover all of parameters, hyperparameters and network architecture. Unseen task should include sufficiently large labelled dataset to achieve accurate adaptation to vastly different tasks. 

In deep neural network, weights and activations are highly correlated causing severe parameter redundancy in the model, as a result we can use a few parameters to predict the others. In unseen tasks, we estimate a few parameters rapidly and use the pre-trained predictive mapping to estimate the output directly. The pre-requisite for this approach is that pre-trained predictive mapping stays fixed across all unseen few-shot tasks and represents shared features of all tasks.
\textbf{Activation to parameter} \cite{Qiao2018} proposes using a feedforward pass that maps activations to parameters in the last layer of a pre-trained deep neural network. It applies to few-shot learning where the number of categories is large and the sample size per category is small.

In offline pre-trained deep model,
a deep neural network is trained on a large dataset $\mathcal{D}_{large}$ with categories $\mathcal{C}_{large}$. Few-shot adaptation model is trained
on a small dataset $\mathcal{D}_{few}$ with categories $\mathcal{C}_{few}$.
Denote the activations before fully connected layer as $a(\pmb{x})$.
Denote the set of activations for label $y$ as $\mathcal{A}_y=\{a(\pmb{x})|\pmb{x}\in\mathcal{D}_{large}\cap\mathcal{D}_{few},Y(\pmb{x})=y\}$.
Denote the mean of activations in $\mathcal{A}_y$ as $\bar{a}_y$.
Denote the parameter for category $y$ in fully connected layer as $w_y$ and $w_y$ is used to compute classifier output directly.
The pre-trained mapping from activations to parameters in deep neural network is
\begin{eqnarray}
\phi:\bar{a}_y\to w_y.
\nonumber
\end{eqnarray}
Activation $s_y$ is sampled from $\mathcal{A}_y\cup \bar{a}_y$ with probability $p$ to be $\bar{a}_y$ and $1-p$ to be uniform in
$\mathcal{A}_y$. Pre-trained mapping from activation to classifier parameter can be re-written as $\phi:s_y\to w_y$.
Define $S_{large}=\{s_1,\cdots,s_{|C_{large}|}\}$.
The pre-trained mapping
$\phi$ is estimated by minimizing cross-entropy loss function:
\begin{eqnarray}
\mathcal{L}(\phi)=\sum_{(\pmb{x},y)\in\mathcal{D}_{large}} \mathbb{E}_{S_{large}} [ -\phi(s_y)a(\pmb{x})+\log \sum_{k \in\mathcal{C}_{large}}\exp\left\{\phi(s_{k})a(\pmb{x})\right\} ],
\nonumber
\end{eqnarray}
maxmizing probability of belonging to the right class and minimizing probability of being in the wrong classes.
In few-shot adaptation of pre-trained mapping $\phi$,
define $\mathcal{C}=\mathcal{C}_{large}\cup \mathcal{C}_{few}$ and $\mathcal{D}=\mathcal{D}_{large}\cup \mathcal{D}_{few}$.
This is a typical setting in adaptation of pre-trained deep model, a large annotated dataset $\mathcal{D}_{large}$ is used to pre-train a deep model and to estimate these generalizable fixed mappings from parameter to classifier output.
Define $S=\{s_1,\cdots,s_{|\mathcal{C}|}\}$.
Category prediction in few-shot data is represented as the probability of $\pmb{x}$ in category $y$:
\begin{eqnarray}
P(y|\pmb{x})=\exp\left\{\mathbb{E}_S [\phi(s_y)a(\pmb{x})]\right\} /\sum_{k\in \mathcal{C}}\exp \left[\mathbb{E}_S\left(\phi(s_{k})a(\pmb{x})\right)\right].
\nonumber
\end{eqnarray}
Pre-training of a deep neural network consumes time, memory and electricity. In order for the pre-trained deep model to be accurate, a large labelled dataset is required. The degree of similarity between tasks should be reasonably high for precise adaptation of the pre-trained mapping. 
In image classification tasks, prediction accuracy is high implying that pre-training the offline deep learner is worthwhile. Nowadays there are many well-known large labelled datasets to pre-train deep model for typical AI missions such as computer vision or language processing. The pre-trained model is available online for free use so that they can be adapted easily to generalize to similar smaller tasks.

In order to adapt pre-trained deep neural network, we can parameterize it with meta-parameter and task-specific parameters so that the network itself allows for rapid adaptation. Task-specific parameters are updated by minimizing loss function on task training data. Meta-parameter is updated by minimizing generalization error function on multiple tasks. Given only few-shot data on unseen task, only a small proportion of parameters can be updated for model adaptation.
Network model proposed in \cite{Munkhdalai2018}
specifies neural network neurons to be conditionally shifted neurons (CSN), which contain activations with task-specific parameters obtained from linear combinations of training experiences in an external memory module. 
Activation of CSN is defined to be
\begin{eqnarray}
h_l=\begin{cases}
\sigma(a_l)+\sigma(\phi_l),~~~~~l\neq M,\\
\text{softmax}(a_l+\phi_l),~~~l=M,
\end{cases}
\nonumber
\end{eqnarray}
where $M$ is the index of output layer in neural network, $a_l$ is the pre-activation and $\phi_l$ is the layer-wise task-specific conditional shift parameter. Pre-activation refers to the linear combination of previous layer neuron states right before activation operation of this layer. 
For ResNet block, we can use CSN as activations and stack these blocks to construct a deep adaptive ResNet model (\textbf{AdaResNet}).
Similarly, we can also stack LSTM layers with CSN activations to build a deep adaptive LSTM model. We may re-design CNN with CSN activations to be a deep adaptive CNN model (\textbf{AdaCNN}). Indeed all famous deep neural network models can be modified to include CSNs for efficient adaptation to few-shot datasets.

For task $\mathcal{T}$, training data is denoted as $\mathcal{D}^{tr}_{\mathcal{T}}=\{(x_i,y_i\}_{i=1}^n$ and validation data is denoted as
$\mathcal{D}^{val}_{\mathcal{T}}=\{(x^*_i,y^*_i\}_{i=1}^m$.
Base learner is a mapping from input data $(x_i,y_i)$ to output label $\hat{y}_i$.
Information in neural network can be specified as a layer-wise amortized error gradient or a direct feedback (DF) which measures layer-wise contribution to the current difference between predicted output label and true label. Higher information implies higher individual influence of the network component upon output label. In black-box adaptation, parameters with higher influence in most relevant tasks are retrieved from external memory module.
Meta-learner
$g_\theta$ is an MLP with $i$th neuron in $l$th layer conditional upon information $I_{l,i}$. 
For validation input data $x^*_j$, task-specific parameter $\phi_l$ in CSN activation represents the task-specific conditional shift:
\begin{eqnarray}
&V_{l,i}=g_\theta(I_{l,i}),~
\alpha_i=\text{softmax}(\cos(f(x^*_j),f(x_i))),\nonumber\\
&\phi_l=[\alpha_1,\cdots,\alpha_n]^T[V_{l,1}, \cdots,V_{l,n}],
\nonumber
\end{eqnarray}
where $I_{l,i}$ is the conditioning information for each neuron stored in external memory module, $f$ is an MLP with a linear output layer for memory query, and $\cos$ is the cosine similarity between extracted features for memory query. $\alpha_i$ is the individual attention mechanism for efficient information retrieval. Experiences in more similar tasks receive more weight in retrieved linear combination.
Neural network is trained by minimizing the cross-entropy loss for task $\mathcal{T}$:
$\mathcal{L}_{\mathcal{T}}=\sum_j \mathcal{L}(\hat{y}^*_j,y^*_j)$, minimizing distance between predicted label and true label on validation data.
Ada+network model is computationally efficient with high prediction accuracy in few-shot image classification. Ada+network combines conditionally shifted neurons, base learner+meta-learner, external memory module. No meta-learning method reflects only one meta-learning approach mentioned in this paper. We have no intention to classify these methods accurately. We want to use these meta-learning methods as examples to illustrate how these meta-learning approaches are used to solve few-shot meta-learning missions.

\begin{table}[htpb]
	\centering
	\caption{Testing accuracy of black-box adaptation meta-learning methods on 5-way 5-shot miniImageNet classification.}
	\label{tabblackbox}
	\begin{tabular}{|m{10cm}|m{2.3cm}|}
		\hline
		Method & Accuracy \\ \hline
		\cite{Qiao2018} Activation to parameter with neural network  & $67.87\pm 0.20\%$  \\ \hline
		\cite{Qiao2018} with wide residual network WRN  & $73.74\pm 0.19\%$  \\ \hline
		\cite{Munkhdalai2018} AdaCNN with DF & $62.00\pm 0.55\%$  \\ \hline
		\cite{Munkhdalai2018} AdaResNet with DF & $71.94\pm 0.57\%$  \\ \hline
	\end{tabular}
\end{table}

Black-box adaptation applies deep learners at meta-level aggregating all task-specific information and provides a direction for model adaptation to out-of-distribution tasks. There are many approaches for black-box adaptation such as stochastic gradient descent, Bayesian optimization, reinforcement learners, genetic algorithms etc.
Table \ref{tabblackbox} presents testing accuracy of these methods on 5-way 5-shot miniImageNet. In miniImageNet classification, wide deep residual network performs better than neural network and CNN. We can see that performance highly depends upon deep model specifications and the fine tuning of pre-trained deep learner. 
Performance of black-box adaptation is comparable to similarity-based meta-learning models in the next section.

AutoML (Automated machine learning) seeks to identify an optimal neural network model which outperforms human-designed network. AutoML is about searching through all hyperparameter combinations to find a global optimum neural network model. At each hyperparameter combination, neural network model is trained to convergence to see whether it outperforms current best target. Hyperparameter includes both continuous hyperparameters such as learning rate, momentum etc and discrete hyperparameters regarding neural model architecture.
Searching efficiency and training efficiency are two main factors to work on for improving global optimum found by AutoML. Meta-learning can be integrated into AutoML to improve either searching efficiency or training efficiency.

\cite{Domhan2015} uses a meta-learner to aggregate learning curves in neural network training process. A parametric model or nonparametric model is applied to model various learning curves and to predict final prediction accuracy at training convergence from early-stage learning curve. At each hyperparameter combination, after only some training iterations, meta-learner provides an estimation of prediction accuracy at convergence, and whether this combo beats current best performance. It improves training efficiency at each hyperparameter combination. Accurate prediction model of learning curve is vital in this algorithm and both parametric and nonparametric models have been developed to improve prediction accuracy. 

\cite{Baker2018} uses a meta-learner to aggregate the relation between hyperparameter combination and neural network model prediction accuracy. Meta-learner learns to predict network prediction performance for new hyperparameter combination. Meta-learner supplies hyperparameter combinations which are likely to outperform current best results for the base learner to explore. Base learner is to train neural network model at each hyperparameter combination, which is viewed as one task. This meta-learning system increases searching efficiency of global optimum hyperparameter setting in AutoML. Prediction accuracy in meta-learner is vital for improving searching efficiency.

\cite{Feurer2015} uses a meta-learner to aggregate datasets and their corresponding optimal machine learning model. Meta-learner extract features from datasets and use these features to determine the best machine learning model for it. In aggregation of previous training experience, meta-learner learns the set of dataset features which are closely related to optimal model selection and parameter initialization. For an unseen task, meta-learner computes the set of dataset features and provides an optimal model recommendation with initial parameters specified. Meta-learner determines the most appropriate model and parameter initialization after extracting relevant features from dataset. This algorithm improves searching efficiency of AutoML by providing a good initial model.

\cite{Liu2018c} develops a progressive neural architecture search algorithm which adds units to neural network model through genetic programming and decision making. This algorithm starts from a simple neural network, adds neurons or weights to network, measures network prediction accuracy after adding each unit, then chooses the action with the best performance gain. Optimal network mutates to several other network models of the same complexity, called children. Then children continue the progressive process of further adding more units. Meta-learner aggregates network model and action choice at each step. Meta-learner learns to predict action performance based upon current network state. Meta-learner recommends actions which are likely to bring the best performance outcome to base learner. This algorithm improves search efficiency at each step by predicting performance gain of each action choice.
\cite{Real2020} is also a progressive neural architecture search algorithm like \cite{Liu2018c}. It starts from a null model, uses genetic programming and the neural network grows to be more complex after adding more units to it. Progressive AutoML algorithms never look back or delete previous actions, never evaluate several previous steps jointly to see whether they are still optimal. Still, progressive search improves searching efficiency in AutoML although it may not reach the global optimal neural network model. Isn't it interesting, forgetting and regret are time-consuming but are the keys to global optimum.

\subsection{Metric-Based Meta-Learning}
\label{Metric-Based-Meta-Learning}

\begin{figure}[htpb]
	\centering
	\includegraphics[width=2in]{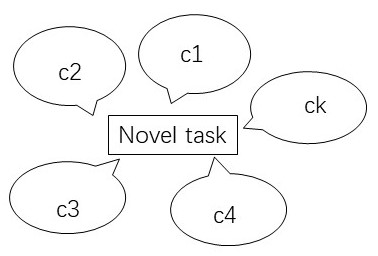}
	\caption{Metric-based meta-learning. $c_i$ is the centroid of class $i$. Distances between a novel task and centroids are compared. A new sample joins the closest class.}
	\label{similarity}
\end{figure}

In this section, we review a class of meta-learning models which depend upon the similarity measure between unseen tasks and previously learned models. From \cite{Schmidthuber}, end-to-end training by directly referring to experiences of most similar tasks is also an efficient scheme for meta-learning.
As in figure \ref{similarity},
distances between an unseen task and the centroids of all trained models are compared and the most similar experiences are applied directly to unseen task.
The pre-requisite is that unseen tasks share similarity structure with previous tasks. Otherwise unseen tasks are not identifiable under the framework of similarity-based meta-learning methods.

A subclass of methods resort to an additional external memory module with an efficient critical information retrieval mechanism. LSTM itself is a complete meta-learning system with an internal memory cell, a self-recursive neuron for self adaptation to unseen tasks, and gates to control information flow. We can also add an external memory module to any deep learning model so that they can efficiently use previous experiences to accelerate task solving. An external memory module has several features: information storage and indexing, information retrieval and query, information deleting and forgetting etc. 
\textbf{MANN} (Memory-augmented neural networks) in \cite{Santoro2016a} adds an external memory module to store most important training experience for query later in unseen tasks.
\textbf{Memory Mod} proposed in \cite{Kaiser2019} adds an external memory module $\mathcal{M}$ to all kinds of classification network so that training experience can be used later to accelerate model convergence. Memory module in 'Memory Mod' stores model training results and has two properties: an efficient nearest neighbor search mechanism and a memory update rule.
Memory module $\mathcal{M}$ consists of three components:
a matrix of memory keys $K_{|\mathcal{M}|\times key-size}$,
a vector of memory values $V_{|\mathcal{M}|}$ and a vector of memory item ages $A_{|\mathcal{M}|}$. Memory keys are the query index through which we can quickly find the most relevant task training experience. Memory values are task training experiences that are closely related to model output, inference of important parameters etc.
The nearest neighbor of a query $q$ from an unseen task in memory module $\mathcal{M}$ is
\begin{eqnarray}
\text{NN}(q,\mathcal{M})=\text{argmax}_i
\cos(q, K[i]),
\nonumber
\end{eqnarray}
where $\cos$ is the cosine similarity measure and $K[i]$ is the key value at location $i$. The
$k$ nearest neighbor of query $q$ in  $\mathcal{M}$
is defined as 
$\text{NN}_k(q,\mathcal{M})=(n_1,n_2,\cdots,n_k)$, where $K[n_i]$ is the key with $i$th highest similarity to query $q$. We may also use supervised deep metric model to estimate an optimal metric which assigns highest similarity to truly relevant tasks.
Memory update finds items with longest ages and writes to these item locations randomly. Items which have been recently used may also be replaced with new experience, since they may not be referred to for a long time. Ideally we expect to store all training experiences in memory module. However, memory space is limited, we must learn to forget to avoid memory explosion. Besides, searching for the most relevant experience may be of low efficiency if there is too much memory to search through.
Memory module can be added to any neural network model such as CNN, LSTM and ResNet.
Memory module is simple, efficient, useful and is compatible with all machine learning models.
For any machine learning model, we only need to specify the most important training experience to be stored in memory and how to query them using similarity metrics.
By adding a memory module with memory update and efficient nearest neighbor search, it enables all classification networks to rapidly and accurately adapt to novel tasks. For out-of-distribution tasks, we accumulate experience in external memory module for a long time so that there will be sufficient experience to use for solving these rarely seen tasks.

Another subclass of methodology is to apply an embedding mapping $f$ from input $x$ to feature $z$, which can then be used to compute a similarity measure between tasks.
Embedding mapping $f$ performs dimension reduction for high-dimensional input $x$.
Similarity measure can be output from an attention block, distance between query input and prototype of each class, or complex similarity score. Choices of embedding mapping function and similarity measure may affect performance of these methods. For each dataset, more options of feature extractor and distance metric function should be explored to find model specifications producing higher prediction accuracy.

\textbf{SNAIL} (Simple Neural Attentive Meta-Learner) \cite{Mishra2018} applies temporal convolutions in a dense block to extract features from images.
An attention block using soft attention function is applied to identify critical features from past experiences and make classification. SNAIL processes sequential data and current neuron state only depends upon past states not future ones. Attention mechanism is embedded within neural network to identify the most relevant previous states to compute current states. For high-dimensional input, attention mechanism is very efficient in dimension reduction and improve data flow efficiency in neural network. Attention mechanism can be used before linear combination of previous states and before activation of linear combination.
In sequence-to-sequence tasks such as language translation,
inputs for task $\mathcal{T}_i$ are denoted as $\{x_s\}_{s=1}^{H_i}$.
Loss function is  $\mathcal{L}_i(x_s,a_s)$, where
input feature is $x_s\sim P_i(x_s|x_{s-1},a_{s-1})$ and
output from attention block is $a_s\sim\pi(a_s|x_1,\cdots,x_s;\theta)$.
Meta-parameter in attention block $\theta$ is updated by minimizing loss function:
\begin{eqnarray}
\min_{\theta}\mathbb{E}_{\mathcal{T}_i\sim p(\mathcal{T})}[\sum_{s=1}^{H_i} \mathcal{L}_i\left\{x_s,a_s(x_1,\cdots,x_s;\theta)\right\}].
\nonumber
\end{eqnarray}
In 5-way 5-shot miniImageNet classification task, SNAIL shows testing accuracy of $68\%$ \cite{Mishra2018} comparable to other meta-learning methods reviewed in black-box adaptation. Obviously SNAIL is not only for few-shot tasks and can be used for large-sample high-dimensional input as well. SNAIL consists of two parts for dimension reduction: temporal convolution layers and causal attention blocks. Convolution layers reduce dimension through averaging data to extact invariant features. Attention layers extract most relevant previous states to compute current state. SNAIL only has dimension reduction layers and final output classifier layer. 
SNAIL can be viewed as modifications to deep convolutional network by embedding attention mechanism between layers and enforcing current state to depend only on previous states to process sequential data.

In addition, RN (\textbf{Relation Network})  \cite{Sung2018} applies supervised deep metric learning to train a similarity metric between extracted features of high-dimensional input.
Denote an embedding function (feature extractor) as $f_{\phi}$ parameterized by $\phi$
and a relation function (similarity metric) as $g_{\theta}$ parameterized by $\theta$.
Relation score $r_{i,j}$ measures the similarity metric between query $\pmb{x}^*_j$ and training data $\pmb{x}_i$:
\begin{eqnarray}
r_{i,j}=g_{\theta}[f_{\phi}(\pmb{x}_i),f_{\phi}(\pmb{x}^*_j) ],
\nonumber
\end{eqnarray}
where feature extractor $f_{\phi}$ is first applied to input data and then similarity metric $g_{\theta}$ is computed.
Higher relation score indicates greater similarity between query and training data.
Parameters in the embedding function and relation function are jointly estimated through the minimization of classification error function:
\begin{eqnarray}
\min_{\phi,\theta} \sum_{i=1}^n\sum_{j=1}^m \{r_{i,j}-\mathbb{I}(y_i=y^*_j)\}^2,
\nonumber
\end{eqnarray}
where $\mathbb{I}(y_i=y^*_j)$ is $1$ if $y_i=y^*_j$ and $0$ otherwise. Minimization of classification error implies that relation score is highest (closest to 1) when two samples are in the same class and lowest (closest to 0) when two samples are in different classes. Relation score itself is a similarity measure between different samples. Trainable parameters in relation score include parameters in feature extractor and parameters in similarity metric function. Usually metric-based meta-learning methods only include trainable parameters in feature extractor when designing similarity measures.
On 5-way 5-shot miniImageNet, RN shows prediction accuracy of $65\%$ comparable to  SNAIL but lower than black-box adaptation in \cite{Qiao2018} and \cite{Munkhdalai2018}. Relation network model can be embedded within larger, more complex meta-learning framework. 

Similarly, \textbf{prototypical network} proposed in \cite{Snell2017} is also based upon deep metric learning. 
Prototype in class $k$ is the centroid $c_k$, average of all data features in this class:
\begin{eqnarray}
c_k=\frac{1}{|S_k|}\sum_{(\pmb{x}_i,y_i)\in S_k}
f_{\phi}(\pmb{x}_i),
\nonumber
\end{eqnarray}
where $S_k$ is the set of all data in class $k$.
Distance function $g$ measures the distance between data and centroid of each class. 
Softmax mapping from distance metric $g$ to class probability is
\begin{eqnarray}
p_{\phi}(y=k|\pmb{x})=
\frac{\exp(-g[f_{\phi}(\pmb{x}),c_k])}{\sum_{k^\prime}\exp(-g[f_{\phi}(\pmb{x}),c_{k^\prime}])}.
\nonumber
\end{eqnarray}
Different from relation network, distance metric $g$ contains no trainable parameter. All parameters are located in feature extraction model.
Finally, we minimize the negative log-probability loss function $J(\phi)=-\log p_{\phi}(y=k|\pmb{x})$ via SGD. Minimization of this loss function maximizes the probability for the input data to be assigned to its true class. The setup in prototypical network is simple, efficient and can be integrated into various more complex meta-learning framework.

AM3 (Adaptive Modality Mixture Mechanism) \cite{Xing2019} combines multi-modal learned image and text data to perform few-shot learning. AM3 uses semantic text to help with few-shot image classification. 
\textbf{TRAML} (Task-Relevant Adaptive Margin Loss) \cite{Li2020} is based upon prototypical network with adaptive margin loss. 
AM3 data used with TRAML algorithm achieves better few-shot image classification than algorithms purely based upon images. The gain comes from adaptive margin loss and multi-modal image data in few-shot tasks.
TRAML achieves $67.10\pm 0.52\%$ on 5-way 1-shot miniImageNet and reaches $79.54\pm 0.60\%$ on 5-way 5-shot miniImageNet. By combining multi-modal learning (image+text) and metric-based meta-learning (Prototypical Network), AM3+TRAML provides better prediction accuracy. 

\textbf{DAPNA} (domain adaptation prototypical network with attention) \cite{Guan2020} integrates domain adaptation into Prototypical Network and shifts and scales class centroids in adaptation to unseen tasks. DAPNA makes more components in prototypical network to be task-dependent in order to gain better prediction accuracy in few-shot tasks. In DAPNA, for each unseen task, centroids in prototypical network is now task-dependent. In addition, DAPNA utilizes margin disparity discrepancy to train deep distance metric which also contributes to higher prediction accuracy. DAPNA compares the performance of few-shot meta-learning methods and lists out the feature extraction models they use. DAPNA uses wide resnet to be feature extraction model. Compared to deep resnet and deep convolutional network, generally wide resnet \cite{Zagoruyko2016} has better performance than resnet, than deep convolutional network. Better feature extraction model also improves few-shot image classification accuracy.

\cite{Das2020} also combines similarity-based metric learning and meta-learning. The setup of \cite{Das2020} is similar to prototypical network \cite{Snell2017}, where  cluster centroids are benchmarks in classification. \cite{Das2020} is the first paper which applies Mahalanobis distance to make few-shot classification within meta-learning framework. Mahalanobis distance is popular in deep metric learning. It contains a parameter matrix which includes many trainable parameters.
Here in \cite{Das2020} Mahalanobis distance between extracted features $\pmb{x}_i$ and $\pmb{x}_j$ is defined as
\begin{eqnarray}
g(\pmb{x}_i,\pmb{x}_j)=\frac{\Vert \pmb{x}_i - \pmb{x}_j \Vert_2^2}{\sigma^2},\nonumber
\end{eqnarray}
where cluster-wise data variance $\sigma^2$ is estimated and updated in adaptation to unseen tasks. In Mahalanobis distance, parameter matrix is set to be an inversion of cluster-wise variance. Considering cluster-wise data variation leads to more accurate classification prediction especially when data variation is vastly different in different clusters.
In addition, in feature extraction model, \cite{Das2020} proposes a relative-feature extractor in addition to absolute deep-feature extractor $f_\phi$. Relative-feature extractor is a modified version of usual feature extraction model. It ensures that the number of extracted features is less than sample size for few-shot tasks. Few-shot tasks use few-shot high-dimensional input where the number of input features is greater than sample size. Over-fitting occurrs when the number of extracted features is greater than sample size. To avoid over-fitting, dimension of relative-feature extractor does not exceed sample size in few-shot novel tasks.
However, generally higher dimension of extracted features will improve classification prediction accuracy in supervised deep metric learning. Usually in  loss function, a regularization constraint is imposed upon feature extraction model to limit its complexity so to avoid over-fitting. Usually there is no strict constraint that the number of extracted features is no greater than sample size.


Lastly, \textbf{TADAM} \cite{Oreshkin2018} is an extension of prototypical network and is also based upon supervised deep metric learning. On mimiImageNet 5-shot 5-way, TADAM shows testing accuracy of $76\%$, greater than SNAIL, RN and prototypical network.
%
%
Similar to prototypical network, a softmax function is applied to compute the probability of belonging to class $k$:
\begin{eqnarray}
p_{\phi,\alpha}(y=k|\pmb{x})=
\text{Softmax}(-\alpha g[f_{\phi}(\pmb{x}),c_k]),
\nonumber
\end{eqnarray}
where $g$ is a distance metric with no trainable parameter, $\alpha$ is the temperature parameter.
Parameters $\phi$ and $\alpha$ are jointly estimated by minimizing the
class-wise cross-entropy loss function
$J_k(\phi,\alpha)$:
\begin{eqnarray}
\sum_{\pmb{x}_i}
\left\{\alpha g[f_{\phi}(\pmb{x}_i),c_k]+\log\sum_{j} e^{-\alpha g[f_{\phi}(\pmb{x}_i),c_j]}\right\}.
\nonumber
\end{eqnarray}
By introducing an additional parameter $\alpha$ in softmax function and through joint optimization of embedding parameter $\phi$ and metric hyperparameter $\alpha$, prediction accuracy of TADAM on 5-way 5-shot miniImageNet is $8\%$ greater than prototypical network.
TADAM applies task-dependent centroids in prototypical network. Given centroids, TADAM uses fully-connected residual neural network and task-dependent centroids to model scaling and shifting parameter of features extracted from previous layer. 
All layers in the feature extractor of TADAM are task-dependent, therefore centroids are task-dependent as well. By allowing more components in prototypical network to be task-dependent, TADAM achieves better adaptation to unseen task and shows higher prediction accuracy per task.

\cite{Gidaris2018} proposes 
 \textbf{Dynamic Few-Shot} algorithm which
consists of 
a ConvNet-based classification model,
a few-shot classification weight generator
and a cosine-similarity based classifier.
Compared with former similarity metric based methods, it contains an additional classification weight generator which adapts ConvNet parameters to novel tasks. It combines black-box adaptation with similarity-based approach and an external memory module. 'Dynamic Few-Shot' preserves model performance upon all trained tasks and seeks higher prediction accuracy on unseen few-shot tasks. ConvNet trained parameters and their corresponding class-wise feature centroids are all stored in the external memory. For an unseen task with unseen image classes, attention mechanism is applied to form a linear combination of ConvNet parameters from most similar tasks and classes. In final layer classifier, extra neurons corresponding to unseen image classes are added, and link weights come from the linear combination of ConvNet parameters created by attention mechanism.

Training data contains $N_{tr}$ base categories each with $K$ examples.
Training data is denoted as $\mathcal{D}^{tr}=\cup_{b=1}^{N_{tr}}\{x_{b,i}\}_{i=1}^K$.
Testing data contains $N_{ts}$ novel categories each with $K$ examples.
Testing data is denoted as $\mathcal{D}^{test}=\cup_{n=1}^{N_{ts}}\{x^*_{n,i}\}_{i=1}^K$.
Embedding feature
$z^*_n=\{z^*_{n,i}\}_{i=1}^K$, where
$z^*_{n,i}=f_\phi(x^*_{n,i})$.
Denote $W_{tr}$ as the classification weight of base categories.
Generate a classification weight vector for novel categories
$W^*_n=G(z^*_n,W_{tr}|\theta)$ using testing data features and base categories weights.
Denote $W_{ts}^*=\{W_n^*\}_{n=1}^{N_{ts}}$ as a classification weight vector of novel categories.
Let $W^*=W_{tr}\cup W_{ts}^*$.

Classification weight is
$W^*=(w_1^*,w_2^*,\cdots,w_N^*)$.
Raw classification score of query $z$ is based upon cosine similarity:
$g_k=\tau \cos(z,w_k^*)$, where $k=1,\cdots,N$ and $\tau>0$.
Probability of belonging to class $k$ is
$p_k=\text{Softmax}(g_k)$. 
Embedding parameter $\phi$ and parameter in classification weight generator $\theta$ are jointly estimated through the
minimization of loss function:
\begin{eqnarray}
\min_{\phi,\theta}\frac{1}{N_{tr}}\sum_{b=1}^{N_{tr}}
\frac{1}{K}\sum_{i=1}^K \mathcal{L}_{\phi,\theta}(x_{b,i},b),
\nonumber
\end{eqnarray}
where $x_{b,i}$ is the $i$th training example in category $b$.
Minimization of loss function maximizes probability of $x_{b,i}$ belonging to its true class $b$ and minimizes probability of wrong classes.
Feature extractor options and classification weight generators may be explored to achieve higher adaptation performance. 'Dynamic few-shot' designs a unique way of adapting pre-trained deep model to unseen tasks: by adding extra neurons for unseen classes to final layer classifer, by generating weights for corresponding extra links using attention mechanism to extract parameters from external memory module.
It modifies both network architecture and weight parameters to adapt pre-trained neural network. For adaptation to vastly different tasks, a greater portion of parameters and hyperparameters should be allowed to change during adaptation to unseen tasks.

In addition, distance metric measure may be designed to estimate similarity structure between tasks, as in \cite{Triantafillou2017}. It proposes a similarity ranking based measure
\textbf{mAP} (mean Average Precision). 'mAP' identifies most similar samples for each sample, and use model experiences from these most similar samples to accelerate training on each sample. Rather than searching for relevant experience to refer to when training each task, 'mAP' performs this process much more frequently by doing so for each sample within task. 
Support set is denoted as 
$S=\{(\pmb{x}_1,y_1),\cdots,(\pmb{x}_N,y_N)\}$.
Define input data $\mathcal{B}=(\pmb{x}_1,\cdots,\pmb{x}_N)$.
Let $R^{\pmb{x}_i}=\{\pmb{x}_j\in\mathcal{B}:y_j=y_i\}$ be the set
of all samples in class $y_i$.
Let $O^{\pmb{x}_i}$ be the ranking of predicted similarity between $\pmb{x}_i$ and the other points in $\mathcal{B}$, where
$O^{\pmb{x}_i}_j$ is the $j$th element in $O^{\pmb{x}_i}$ with the $j$th highest similarity to $\pmb{x}_i$.
Let $P(j,\pmb{x}_i)$ be the proportion of points that are truly relevant to $\pmb{x}_i$ in the first $j$ items in $O^{\pmb{x}_i}$.

Average Precision (AP) of this ranking is defined as
\begin{eqnarray}
AP^{\pmb{x}_i}=|R^{\pmb{x}_i}|^{-1}\sum_{j:O^{\pmb{x}_i}_j \in R^{\pmb{x}_i}}
P(j,\pmb{x}_i),\nonumber\\
\text{where}~P(j,\pmb{x}_i)=
j^{-1}|\{k\leq j: O^{\pmb{x}_i}_k\in R^{\pmb{x}_i}\}|.
\nonumber
\end{eqnarray}
Mean Average Precision (mAP) is the mean of AP for all points:
$
mAP=|\mathcal{B}|^{-1}\sum_{i\in\mathcal{B}}
AP^{\pmb{x}_i}.
$
'mAP' measures quality of ranking samples based upon predicted similarity, since samples from same class should exhibit highest similarity and samples from different classes should show lowest similarity. Higher 'mAP' implies that predicted similarity is more reliable. We may use supervised deep metric learning to find an optimal similarity measure which maximizes 'mAP'.
Let $\varphi_\phi(\pmb{x}_i,\pmb{x}_j)$ be the cosine similarity between $\pmb{x}_i$ and $\pmb{x}_j$: $\varphi_\phi(\pmb{x}_i,\pmb{x}_j)=\cos[f_\phi(\pmb{x}_i),f_\phi(\pmb{x}_j)]$.
Denote $\mathcal{P}^{\pmb{x}_i}$ as the set of all points with predicted similarity to $\pmb{x}_i$. 
Denote $\mathcal{N}^{\pmb{x}_i}$ as the set of all points without predicted similarity to $\pmb{x}_i$. 
Indicator $y_{kj}^i$ is $1$ if $\varphi_\phi(\pmb{x}_i,\pmb{x}_k)>\varphi_\phi(\pmb{x}_i,\pmb{x}_j)$ and $0$ otherwise.
For each query $\pmb{x}_i$, 
score function is defined as
\begin{eqnarray}
&F_\phi(\mathcal{B},y)=\sum_{\pmb{x}_i\in\mathcal{B}}F_\phi^{\pmb{x}_i}(\mathcal{B},y),~\text{where}\nonumber\\
&F_\phi^{\pmb{x}_i}(\mathcal{B},y)=\frac{1}{|\mathcal{P}^{\pmb{x}_i}||\mathcal{N}^{\pmb{x}_i}|}
\sum_{k\in \mathcal{P}^{\pmb{x}_i}\backslash i}
~~\sum_{j\in \mathcal{N}^{\pmb{x}_i}} 
y_{kj}^i \left\{\varphi_\phi(\pmb{x}_i,\pmb{x}_k)-\varphi_\phi(\pmb{x}_i,\pmb{x}_j)\right\}.
\nonumber
\end{eqnarray}
Maximization of score function estimates the optimal ranking most compatible with predicted similarity ranking for each sample.
Indicator $p^i_g$ is $1$ if $\pmb{x}_g$ is truly relevant to query $\pmb{x}_i$ and $-1$ otherwise. Denote $p^i$ to be a vector of all $p^i_g$ and denote $\hat{p}^i$ to be a vector of all predicted indicator $\hat{p}^i_g$.
AP loss for query $\pmb{x}_i$ is formulated as
\begin{eqnarray}
\mathcal{L}^{\pmb{x}_i}(p^i,\hat{p}^i)=1-\frac{1}{|\mathcal{P}^{\pmb{x}_i}|} \sum_{j:\hat{p}^i_j=1}
P(j,\pmb{x}_i).
\nonumber
\end{eqnarray}
mAP loss is the mean of AP loss over all query points. Minimization of AP loss maximizes accuracy of predicted similarity, ie samples from the same class have highest similarity and samples from different classes have lowest similarity.
Loss-augmented label prediction for query $\pmb{x}_i$ is a linear combination of score function and AP loss:
\begin{eqnarray}
\max_{\hat{y}_i}
\left\{F_\phi^{\pmb{x}_i}(\mathcal{B},y)-\epsilon \mathcal{L}^{\pmb{x}_i}(p^i,\hat{p}^i)\right\},
\nonumber
\end{eqnarray}
where $\epsilon>0$. Simultaneous maximization of score function $F_\phi^{\pmb{x}_i}(\mathcal{B},y)$ and minimization of AP loss $\mathcal{L}^{\pmb{x}_i}(p^i,\hat{p}^i)$ estimates predicted similarity such that  samples from the same class have highest similarity and samples from different classes have lowest similarity.

\begin{table}[htpb]
	\centering
	\caption{Testing accuracy of metric-based meta-learning methods on 5-way 5-shot miniImageNet classification. }
	\label{tabsimilarity}
	\begin{tabular}{|m{10cm}|m{2.3cm}|}
		\hline
		Method & Accuracy \\ \hline
		\cite{Vinyals2016} Matching Net & $60.0\%$  \\ \hline
		\cite{Mishra2018} SNAIL  & $68.88\pm 0.92\%$  \\ \hline
		\cite{Sung2018} Relation Net  & $65.32\pm 0.70\%$  \\ \hline
		\cite{Snell2017} Prototypical Net  & $68.20\pm 0.66\%$  \\ \hline
		\cite{Das2020} & $70.91\pm 0.85\%$  \\ \hline
		\cite{Li2020} Prototypical Net+TRAML  & $77.94\pm 0.57\%$  \\ \hline
		\cite{Li2020} AM3+TRAML  & $79.54\pm 0.60\%$  \\ \hline
		\cite{Guan2020} DAPNA  & $84.07 \pm 0.16\%$  \\ \hline
		\cite{Oreshkin2018} TADAM with $\alpha$, AT and TC  & $76.7\pm 0.3\%$ \\ \hline
		\cite{Oreshkin2018} TADAM without tuning  & $74.2\pm 0.2\%$  \\ \hline
		\cite{Gidaris2018} Dynamic few-shot with C128F feature extractor  & $73.00\pm 0.64\%$ \\ \hline
		\cite{Gidaris2018} with ResNet feature extractor  & $70.13\pm 0.68\%$  \\ \hline
		\cite{Gidaris2018} with cosine classifier and attention based weight generator  & $74.92\pm 0.36\%$  \\ \hline
		\cite{Gidaris2018} with cosine classifier and no weight generator  & $72.83\pm 0.35\%$  \\ \hline
		\cite{Triantafillou2017} mAP-SSVM  & $63.94\pm 0.72\%$  \\ \hline
		\cite{Triantafillou2017} mAP-DLM  & $63.70\pm 0.70\%$  \\ \hline
	\end{tabular}
\end{table}

From table \ref{tabsimilarity} which presents testing accuracy of similarity-based methods on 5-way 5-shot miniImageNet, we can see that TADAM and DAPNA show highest accuracy. Both TADAM and DAPNA are modifications from prototypical network by making more components in it to be task-dependent so that the algorithm fits each unseen task better. On average, performance of similarity-based methods is slightly better than black-box adaptation which requires heavy pre-training computation.
Similarity-based methods are more flexible and allow choosing from various options of feature extractor, similarity measure, loss function and hyperparameters. Similarity-based methods require fine tuning on these choices for best adaptation performance. Better feature extractor, better similarity measure, more discriminative loss function and more variable hyperparameters generally improve prediction accuracy on few-shot tasks.

\subsection{Layered Meta-Learning}
\label{Layers-Meta-Learning}

\begin{figure}[htpb]
	\centering
	\includegraphics[width=2in]{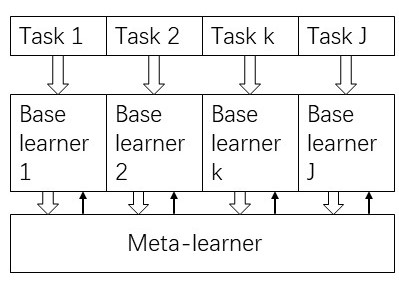}
	\caption{Learner and meta-learner framework of meta-learning. Base learner is trained on each task. Meta-learner updates task-specific components in base learner for adaptation across different tasks.}
	\label{metalearner}
\end{figure}

\cite{Schmidthuber} proposes meta-learning, base layer (base learner), meta layer (meta-learner), meta-meta layer etc. This line of methodology is referred to as layered meta-learning here in this paper. It may have other names elsewhere.
\cite{Bock1988} provides an overview of two-level few-shot meta-learning models which consist of a base model that learns rapidly from few-shot data, and a meta-model that optimizes the base learner across few-shot tasks. As in figure \ref{metalearner}, base learner is designed for solving each task. 
Base learner contains two parts of parameters: task-specific parameters and meta-parameters.
Base learner updates task-specific parameters in adaptation to different tasks.
Meta-learner accumulates experiences from multiple tasks, mines their shared features, minimizes loss function on validation data to minimize generalization error and updates meta-parameters in base learner. 
Base learner learns task-specific features by updating task-specific parameters in adaptation to each unseen task.
Meta-learner models features shared by all tasks, guides base learner to better generalize to unseen tasks, maximizes generalization capability of base learner.
Base learner supplies task-specific information and generalization error on task validation data to meta-learner. Meta-learner aggregates task-specific information to learn features shared by all tasks and minimizes generalization error of base learner to update meta-parameters. Meta-learner supplies updated meta-parameters to base learner. Communication between base learner and meta-learner can be designed to be quite complicated to allow closer cooperation between base layer and meta layer.
Meta layer (Meta-learner) is used to guide training of base layer (base learner). Meta meta layer (meta-meta learner) may be applied to guide training of meta layer (meta-learner).

Usually there is tradeoff between model accuracy on training data and model accuracy on validation data. Model accuracy on training data evaluates quality of model estimation. Model accuracy on validation data measures generalization capability of deep model. That is, usually there is tradeoff between model estimation quality and model generalization capability. 
Overfitting refers to the case where model generalization capability is very low and model estimation quality is very high.
Base learner concentrates upon efficient and accurate model estimation on each unseen task. Meta-learner focuses upon minimizing generalization error of base learner. By introducing base learner and meta-learner,
layered meta-learning approach separates model generalization from model estimation and allows the combination of different variations in adaptation specifications and estimation procedures. Optimization on base learner maximizes quality of model estimation and optimization on meta-learner maximizes model generalization capability. Through optimization on layered meta-learning, both model estimation quality and model generalization capability are maximized. In layered meta-learning, there is no tradeoff between model estimation and model generalization capability. 
Meta-LSTM \cite{Younger2001}, Meta Network (MetaNet) \cite{Munkhdalai2017}, MetaOptNet \cite{Lee2019}, Meta-SGD \cite{Li2017} and MAML \cite{Finn2017} are all methods under this meta-learning framework. 
Base learner is designed for each task and can be specified as a machine learning algorithm or a statistical model. 
The goal of few-shot meta-learning is to find an efficient and accurate solution on each task. As a result, base learner should be efficient and accurate on each task. Meta-learner can be slow to aggregate model training experiences on multiple tasks.
In MetaOptNet \cite{Lee2019}, base learner is specified to be a convex linear classifier and meta-learner is posited as a neural network. MetaOptNet shows superior prediction performance in miniImageNet classification.




Communication between base learner and meta-learner should be efficient, supporting effective adaptation of base learner to unseen tasks. More complex communication leads to increased complexity in communication but may help reduce training time of base learner. \cite{Li2019a} considers the setup of meta-learning under federated learning where data privacy is of primary concern. Data privacy limits the degree of communication between base learner of individual tasks and meta-learner across all tasks. \cite{Li2020a} uses meta-learner to adapt feature transformation matrix module and pixel-wise prediction module in a single model for image cropping under pre-specified aspect ratio.


\textbf{GBML} (Gradient-based meta-learning) is a group of meta-learning methods that update learner parameters using stochastic gradient descent to provide efficient model generalization. 
In GBML, base learner is specified to include both task-specific parameters and meta-parameters. Here meta-parameters are the initial values of task-specific parameters. Meta-learner minimizes the sum of loss function on validation data across all tasks to compute meta-parameters. Meta-learner provides meta-parameters to base learner as good initial values for task-specific parameters. On each few-shot task, after only few rounds of SGD iterations, task-specific parameters are updated to achieve sufficiently high prediction accuracy on the task. Meta-learner provides good initial task-specific parameters to base learner so that base learner meets high prediction accuracy efficiently after only few rounds of SGD updates on task-specific parameters.
GBML only applies to similar tasks with similar data structure. For example, in few-shot image classification tasks, unseen tasks have the same data structure as trained tasks with the same number of classes and same number of images per class. Unseen tasks should also be few-shot image classification tasks and images should have the same number of pixels as trained tasks. In transfer learning, generalization to vastly different tasks is difficult and may cause negative transfer leading to inferior performance. In meta-learning, researchers have been trying to develop meta-learning framework that generalizes better and solves vastly different tasks.

Typical GBML methods include MAML \cite{Finn2017}, Reptile \cite{Nichol2018a}, FOMAML \cite{Nichol2018}, HF-MAML \cite{Fallah2019}, TMAML \cite{Liu2019a}, ES-MAML \cite{Song2019} and PROMP \cite{Rothfuss2019}. GBML is applicable to any learner optimized with stochastic gradient descent. 
Reptile \cite{Nichol2018a}, FOMAML \cite{Nichol2018} and HF-MAML \cite{Fallah2019} are first-order approximations of MAML. They are more efficient than MAML in training but are less accurate than MAML in prediction accuracy.
TMAML \cite{Liu2019a} and PROMP \cite{Rothfuss2019} propose low-variance and high-quality estimators of gradient and Hessian, which also improve training process of meta-learning framework. 
ES-MAML \cite{Song2019} applies evolution strategies (ES) in the searching process for a globally optimum solution and extends MAML to nonsmooth adaptation operations. 
GBML is widely applied for generalization of neural network to complex tasks.  

\textbf{MAML} (Model-Agnostic Meta-Learning) \cite{Finn2017}
is applicable to any learner that can be optimized with SGD (Stochastic Gradient Descent). It is the most famous meta-learning method in recent literature. 
MFR (Meta Face Recognition) \cite{Guo2020} applies MAML to generalize face recognition model across different human races. \cite{Guo2020} regards different human races as different domains. Generalization between domains can be easily conducted through direct application of well-trained deep model with domain adaptation embedded. GBML can be integrated within any deep model to maintain model estimation quality and gain additionally model generalization capability. After integration with GBML, deep model maintains high prediction accuracy on in-distribution tasks and has higher prediction accuracy on out-of-distribution tasks.
\cite{Choi2020} integrates MAML in deep neural network for scene-adaptive video frame interpolation. There are many deep learning applications where integration with GBML improves deep model performance and reaches new state-of-the-art accuracy in that application area.
We will review Bayesian extensions of MAML in next section about Bayesian meta-learning framework and integrated learning frameworks based upon MAML in later section about applications of meta-learning methodology.
Base learner is denoted as $h_{\theta}$ parameterized by meta-parameter $\theta$.
Task-specific parameter is denoted as $\phi$.
Task $\mathcal{T}_i$ follows task distribution $p(\mathcal{T})$.
In the inner loop of MAML, base learner is trained where
task-specific parameter $\phi$ is updated using
\begin{eqnarray}
\phi_i=\theta-\alpha \bigtriangledown_\theta 
\mathcal{L}_{\mathcal{T}_i} (h_\theta),
\nonumber
\end{eqnarray}
where $\alpha$ is the trainable step size parameter or learning rate parameter, $\theta$ is the meta-parameter which is the initial values of task-specific parameter $\phi$ in base learner. $\mathcal{L}_{\mathcal{T}_i} (h_\theta)$ is loss function on task training data. $\bigtriangledown_\theta 
\mathcal{L}_{\mathcal{T}_i} (h_\theta)$ is training data loss gradient evaluated at meta-parameter $\theta$.
In the outer loop, meta-learner is trained where
meta-parameter $\theta$ is updated using (meta-update)
\begin{eqnarray}
\theta\leftarrow \theta-\beta\bigtriangledown_\theta 
\sum_{\mathcal{T}_i\sim p(\mathcal{T})}
\mathcal{L}_{\mathcal{T}_i} (h_{\phi_i}),
\nonumber
\end{eqnarray}
where $\beta$ is the step size or learning rate. $\mathcal{L}_{\mathcal{T}_i} (h_{\phi_i})$ is loss function on task validation data evaluated at trained task-specific parameter $\phi_i$ for task $\mathcal{T}_i$. Trained task-specific parameter $\phi_i$ is a function of its initial value (meta-parameter) $\theta$.
In this meta-update, by chain rule, derivative with respect to $\theta$ includes loss gradient and derivative $\partial \phi_i/ \partial \theta$, which leads to higher-order derivative of loss.
MAML is applicable to RL (reinforcement learning) tasks as well. After defining loss function, and specifying task-specific parameters and meta-parameters, we can easily write out MAML framework for few-shot RL tasks.
Base learner updates task-specific parameter $\phi_i$ using task training data $\mathcal{T}_i$ and meta-learner updates meta-parameter $\theta$ based upon all task validation data. In MAML, both base learner and meta-learner are SGD optimizers which ensure efficient adaptation to novel tasks. \cite{Antoniou2019} provides valuable empirical experiences for training MAML framework, since in MAML higher-order gradient of loss function is included in parameter update which may cause instability in training.

\textbf{Meta-SGD} \cite{Li2017} also belongs to GBML and has the same structure as MAML. But meta-SGD tunes the base learner SGD step size $\alpha$ jointly with meta-parameter $\theta$. In the meta-learner, step size $\alpha$ and meta-parameter $\theta$ are jointly updated using (meta-update)
\begin{eqnarray}
(\theta,\alpha)\leftarrow (\theta,\alpha)-\beta\bigtriangledown_{(\theta,\alpha)}
\sum_{\mathcal{T}_i\sim p(\mathcal{T})}
\mathcal{L}_{\mathcal{T}_i} (h_{\phi_i}),
\nonumber
\end{eqnarray}
where $\beta$ is step size of meta-learner SGD iteration and $\alpha$ is step size of base learner SGD iteration. Trained task-specific parameter $\phi_i$ is a function of both its initial values (meta-parameter) $\theta$ and SGD step size $\alpha$. Similar to MAML, Meta-SGD also includes higher-order gradient of loss function in meta-update. Training art of meta-SGD should be similar to MAML and experiences listed out in \cite{Antoniou2019} should be helpful.
On 5-way 5-shot miniImageNet, testing accuracy of meta-SGD is $64\%$ which is $0.9\%$ greater than MAML.  

In addition, \textbf{Reptile} \cite{Nichol2018a} belongs to GBML and also has the same structure as MAML. But the meta-update on meta-parameter $\theta$ is based upon first-order approximation to loss gradient so to avoid computation of higher-order derivative on loss function. 
MetaPix \cite{Lee2019c} applies Reptile combined with Pix2PixHD \cite{Wang2018a} in adapting human pose skeletons from one person to another person standing in slightly different backgrounds. 
Meta-parameter $\theta$ is estimated through the minimization of expected loss function
$
\min_{\theta} \mathbb{E}_{\mathcal{T}}\left[\bigtriangledown_{\theta}\mathcal{L}_{\mathcal{T}} (h_{\theta})\right].
$
Episodic training is applied and
sampled tasks are denoted as $\mathcal{T}_i$, $i=1,2,\cdots,J$.
Task-specific parameters $\phi_i$ are neural network weights and they are updated using SGD:
\begin{eqnarray}
\phi_i=\text{SGD}(\mathcal{L}_{\mathcal{T}_i},\theta,k),
\nonumber
\end{eqnarray}
where $k$ is the number of gradient iterations taken to update neural network weights $\phi_i$ from its initial values (meta-parameter) $\theta$.
In the meta-learner, meta-paramater $\theta$ is updated using (meta-update)
\begin{eqnarray}
\theta \leftarrow \theta+\epsilon\frac{1}{J}\sum_{i=1}^J (\phi_i-\theta),
\nonumber
\end{eqnarray}
where $\epsilon>0$ is step size or learning rate parameter and $\frac{1}{J}\sum_{i=1}^J (\phi_i-\theta)$ is an approximation to first-order loss gradient. For Euclidean loss which uses $L_2$ norm to measure the distance between predicted value and true value, first-order loss gradient looks like this formula. It does not compute the derivative of $\partial \phi_i/\partial \theta$ which introduces computation of higher-order loss gradient.  
MAML, Reptile and Meta-SGD are all gradient-based meta-learning methods (GBML). They share similar structure and can be easily integrated with any deep learning application model for better performance. 

Different from introduced GBML methods, \textbf{Meta-LSTM} \cite{Ravi2019a} uses a deep neural network classifier as base learner and an LSTM as meta-learner.
Though LSTM itself is a complete meta-learning system \cite{Younger2001} with memory cell and self-recursive neuron, LSTM is applied merely as a meta-learner in meta-LSTM. \cite{Schmidhuber2015} points out using RNN as meta-learner or base learner to achieve higher level of autonomy in robots. Communication between base learner and meta-learner ensures that mutual information between them is high. Efficiency in either base learner or meta-learner should improve efficiency of whole meta-learning sytem overall.
Meta-learner supplies initial values (meta-parameter) of task-specific parameters that contain shared features among tasks. 
Base learner takes this parameter as input and converges faster on novel few-shot tasks. 
Main gradient-based optimization techniques in LSTM include
momentum \cite{NESTEROV1983},
Adagrad \cite{Duchi2010},
Adadelta \cite{Zeiler2012} and
ADAM \cite{Kingma2015}.
There is correspondence between gradient-based optimization techniques and computations in LSTM self-recursive cell. 
In LSTM, by setting proper values for multiplicative input gate, multiplicative output gate, forget gate, candidate cell state and cell state, computations in recursive cell are equivalent to iterations in gradient-based optimization techniques.
In meta-LSTM, LSTM is a meta-learner and meta-parameter update is conducted using computations in recursive cells.
In convex optimization, convergence speed is guaranteed.
In non-convex optimization such as genetic programming and simulated annealing, convergence holds under assumptions as well.
In base learner, task-specific parameters in neural network are updated with MAML. For task $\mathcal{T}_i$, $t$th SGD iteration to update task-specific parameter is
\[
\phi_t=\phi_{t-1}-\alpha_t\bigtriangledown_{\phi_{t-1}}\mathcal{L}_{t}(h_{\phi_{t-1}}).
\]
Neural network weights $\phi_t$ for task $\mathcal{T}_i$ in base learner are updated with loss gradients corresponding to cell state update in LSTM:
\begin{eqnarray}
&c_t=f_t\odot c_{t-1}+i_t\odot \tilde{c}_t,
\nonumber\\
&c_t=\phi_t,\nonumber\\
&\tilde{c}_t=\bigtriangledown_{\phi_{t-1}} 
\mathcal{L}_t,\nonumber\\
&i_t=\sigma(W_I[\bigtriangledown_{\phi_{t-1}}\mathcal{L}_t,\mathcal{L}_t,\phi_{t-1},i_{t-1}]+b_I),
\nonumber\\
&f_t=\sigma(W_F[\bigtriangledown_{\phi_{t-1}}\mathcal{L}_t,\mathcal{L}_t,\phi_{t-1},f_{t-1}]+b_F).
\nonumber
\end{eqnarray}
These formulas show that cell update computations in LSTM recursive neuron are equivalent to SGD parameter update iterations.
Cell state of LSTM is the parameter to be updated and cell state update corresponds to SGD update iteration for the parameter.
Due to the correspondence between SGD iteration and LSTM cell state update, LSTM can be applied to provide task-specific parameter update in base model adaptation.
Both meta-learner and base learner in MAML contains SGD parameter update procedure. Both meta-learner and base learner in MAML can be specified to be LSTM using formulas above.
In meta-LSTM, meta-learner is specified to be LSTM which uses this correspondence to conduct meta-parameter update (meta-update) in LSTM recursive cell. Then for task $\mathcal{T}_t$, meta-parameter $\theta_t$ is updated using SGD iterations in LSTM meta-learner. LSTM is very stable in training even when LSTM depth is large. 

In this few-shot meta-learning framework, there are
nested loops, where the
inner loop is a base learner from each individual task and the
outer loop is a meta-learner from all available tasks. Usually, between-task adaptation depends upon the quality of similarity measure.
\cite{Schmidthuber} proposes meta-learning, base layer (base learner), meta layer (meta-learner) etc.
Later \cite{SchmidhuberIdsia1997} proposes that neural network models with low Kolmogorov complexity exhibit high generalization capability. Efficient lightweight base learner demonstrates improvement in generalization.
\cite{Bertinetto2019} uses this layered meta-learning framework \cite{Schmidthuber}, in which base learner concentrates upon model fitting and meta-learner focuses upon model adaptation. According to \cite{SchmidhuberIdsia1997}, \cite{Bertinetto2019} specifies base learner as an efficient and differentiable learner preferably with an explicit closed-form solution.
Later \cite{Lee2019} follows this line of research and specifies base learner to be convex optimizers which can be solved efficiently.
\cite{Liu2019} applies transductive inference in base learner which has closed-form solution for joint prediction of the whole validation dataset.
   Efficient base learner for fast and accurate adaptation to unseen tasks
+  Relatively slower meta-learner to aggregate training experiences and 
     minimize generalization error across many tasks 
=  An efficient layered meta-learning system.
Under this specification, over-fitting is reduced in base learner and generalization to unseen tasks is more efficient.
In \textbf{R2D2} (Ridge Regression Differentiable Discriminator) \cite{Bertinetto2019}, base learner is specified to be ridge regression which has closed-form solution. 
In \textbf{LR-D2} (Logistic Regression Differentiable Discriminator),
base learner is specified as iteratively reweighted least squares (IRLS)
derived from logistic regression.

\textbf{MetaOptNet} \cite{Lee2019} is developed under layered meta-learning framework \cite{Bertinetto2019}, where base learner is formulated as a regularized linear classifier solvable with convex optimization. This idea of base learner specification is similar to that in \cite{Bertinetto2019}, where an efficient and differentiable statistical base learner is preferred.
MetaOptNet is an extension of \cite{Bertinetto2019} in the sense that it explores more options of base learner specification under similar framework as in \cite{Bertinetto2019}.
Base learner options for convex optimization include linear classifiers such as kNN classifier, SVM classifier with nonlinear kernels and ridge regression with hinge loss etc.
Quadratic programming (QP) can be applied for convex optimization where Karush-Kuhn-Tucker (KKT) condition is derived and then back-propagation using implicit function theorem is performed. 
Parameter $\theta$ is estimated using
\begin{eqnarray}
\min_\theta [\mathcal{L}^{base}
(\mathcal{D}^{tr};\theta,\phi)+\mathcal{R}(\theta)],
\nonumber
\end{eqnarray}
where 
$\mathcal{L}^{base}$ is a loss function of base learner and
$\mathcal{R}(\theta)$ is an $L_2$ regularization on parameter $\theta$ so to avoid overfitting. This optimizer in base learner can be solved efficiently with convex optimization.
Meta-learning objective is to learn an embedding model $f_\phi$ such that base learner generalizes well across tasks. The meta-objective is to minimize generalization error on testing / validation dataset:
\begin{eqnarray}
\min_{\phi}\mathbb{E}_{\mathcal{T}\sim p(\mathcal{T})}[\mathcal{L}^{meta}(\mathcal{D}^{test};\theta,\phi)],
\nonumber
\end{eqnarray}
where $\mathcal{L}^{meta}$ is a loss function of meta-learner. Meta-learner should be more complex to represent complex similarity features shared by many tasks.
For example, meta-loss function can be specified as 
\begin{eqnarray}
\mathcal{L}^{meta}(\mathcal{D}^{test};\theta,\phi,\gamma)=\sum_{(\pmb{x},y)\in \mathcal{D}^{test}}[-\gamma \theta_y f_{\phi}(\pmb{x})
+\log\sum_k \exp(\gamma \theta_k f_{\phi}(\pmb{x}))].
\nonumber
\end{eqnarray}
Minimization of this cross entropy meta-loss function maximizes probability of belonging to the correct class and minimizes probability of belonging to wrong classes.
Note that $\phi$ is meta-parameter and $\theta$ is task-dependent in MetaOptNet, contrary to notations in MAML.
We may choose statistical models as base learners since they are less prone to over-fitting. For example, in MetaOptNet-SVM, base learner can be specified as an SVM classifier. 
\textbf{Auto-Meta} \cite{Kim2018} also follows this line of layered meta-learning framework and specifies meta-learner as a deep CNN model. Auto-Meta trains an optimal CNN structure to be an optimal meta-learner model. By optimizing meta-learner deep model through AutoML, Auto-Meta reduces the number of parameters in meta-learner and increases prediction accuracy in few-shot image classification.

\textbf{TPN} (Transductive Propagation Network)
\cite{Liu2019}
learns adaptation to novel classification tasks using transductive inference in few-shot learning.
Transductive inference 
predicts on whole test data simultaneously, which is different from inductive inference which predicts test data one-by-one. TPN considers all data from training set and validation set to construct a graph model, the link weight of which is defined to be a trained similarity metric with data variance included. 
TPN uses labelled data from training set and a joint graph model of both training and validation data to perform transductive inference for unknown labels of validation data. Generalization of TPN from training to validation data is different from previously introduced methods, since TPN uses both training and validation data to build a graph model rather than relying on only training data to build a predictive model.
TPN maps directly from meta-training data to meta-testing data, constructs a joint graph network model on input data from both support set $\mathcal{S}$ and query set $\mathcal{Q}$ to propagate labels between them.
Transductive meta-learning framework has two components:
a deep CNN feature embedding function $f_{\phi}$ and a learned manifold structure for the space of unseen classes from $\mathcal{S}\cup\mathcal{Q}$.
The cross-entropy loss is computed using feature embedding and graph parameter with
end-to-end updates on all parameters through back-propagation.

Support set is denoted as 
$\mathcal{S}=\{(\pmb{x}_1,y_1),\cdots,(\pmb{x}_{n},y_{n})\}$.
Query set is denoted as 
$\mathcal{Q}=\{(\pmb{x}^*_1,y^*_1),\cdots,(\pmb{x}^*_{m},y^*_{m})\}$.
The graph edge weight is defined as Gaussian similarity measure:
\begin{eqnarray}
W_{ij}=\exp\left\{-\frac{g[f_\phi(\pmb{x}_i),f_\phi(\pmb{x}_j)]}{2\sigma^2}\right\},
\nonumber
\end{eqnarray}
where $\sigma^2$ is data variance modelled as $\sigma=g_\varphi (f_\phi(\pmb{x}))$ parameterized by $\varphi$, $g$ is similarity function between extracted feature embeddings $f_\phi(\pmb{x}_i)$ and $f_\phi(\pmb{x}_j)$.
Only the $k$ largest edge weights in each row are retained to create a $k$-nearest neighbor graph.
$W$ is a row-normalized matrix of all weights in graph.
Indicator $Y_{ij}$ is $1$ if $\pmb{x}_i$ is in the task training set and has label $y_i=j$. Indicator $Y_{ij}$ is $0$ if $\pmb{x}_i$ is in the task validation / testing set or does not have label $y_i=j$. $Y$ is a vector of indicator $Y_{ij}$.
Label propagation in the graph network learned from support set
$\mathcal{S}$ and validation set $\mathcal{Q}$ is specified to be
\begin{eqnarray}
F_{t+1}=\theta W F_t +(1-\theta) Y,
\nonumber
\end{eqnarray}
where
$F_t$ is the predicted labels at timestamp $t$ and the sequence of $\{F_t\}$ converges to 
\[
F^*=(I-\theta W)^{-1}Y.
\]
$F^*$ is an explicit closed-form solution to predicted labels of task validation dataset. Computation of $F^*$ is in base learner and label prediction in base learner is very efficient.
In meta-learner, meta-objective function is defined as
\begin{eqnarray}
&J(\phi,\theta,\sigma)=\sum_{i=1}^m \sum_{j=1}^N 
-Y_{ij}\log P(\hat{y}_i=j|\pmb{x}_i),
\nonumber\\
&\text{where}~P(\hat{y}_i=j|\pmb{x}_i)=\exp(F^*_{ij})/\sum_{j=1}^N \exp(F^*_{ij}),\nonumber
\end{eqnarray}
We minimize meta-loss function $J(\phi,\theta,\sigma)$ to estimate meta-parameters $\phi$, $\theta$ and $\sigma$, where $\phi$ is feature extractor / embedding parameter, $\theta$ is hyperparameter of base learner and $\sigma$ is data variance model parameter.
$y_i$ and $\hat{y}_i$ are the true and predicted label of $\pmb{x}_i$
respectively. Minimization of meta-loss function maximizes the probability for data to be assigned to its true class.
All parameters in TPN are regarded as meta-parameters and are updated in meta-update. Base learner plugs in meta-parameters and all task data to directly calculate predicted labels on validation data. Base learner does not include any parameter update and is very efficient.

\textbf{LEO} (latent embedding optimization) \cite{Rusu2019} is also for
K-shot N-class few-shot learning, where K is small and N is large.
LEO consists of an encoding process and a decoding process, where
encoding process is a feed-forward mapping followed by a relation network, and decoding process is a softmax classifier.
Then parameters in encoder-decoder framework are updated using base learner and meta-learner procedure. To train this meta-learning framework including an encoder-decoder model, we may explore the following options.
(1) Encoder-decoder model is updated efficiently in base learner during adaptation to unseen tasks. Hyperparameters in the encoder-decoder model are meta-parameters that are updated in meta-learner. Hyperparameters in encoder-decoder model reflect shared features of many tasks and parameters in encoder-decoder model reflect task-specific features of each unseen task. (2) To make base learner as efficient as possible, all parameters in encoder-decoder model are regarded as meta-parameters. In base learner, for each task, fitted encoder-decoder model is utilized to directly calculate predicted labels by plugging in task validation data and meta-parameters from meta-learner. In meta-learner, across many task validation data, generalization error is minimized in meta-objective function to estimate all parameters in encoder-decoder model. (3) LEO applies MAML training framework here. Meta-parameters are the initial values of all parameters in encoder-decoder model. In base learner, all parameters in encoder-decoder model are updated for only few SGD iterations from good initial values. In meta-learner, loss on validation data is minimized to update meta-parameters.

Dimension reduction encoder network is denoted as
$f_{\phi_e}:\mathcal{R}^{n_x}\to \mathcal{R}^{n_h}$ parameterized by $\phi_e$, where $n_x>n_h$.
Training data for class $n$ is denoted as
$\mathcal{D}_n^{tr}=\{x_n^k,y_n^k\}_{k=1}^K$.
Relation network is denoted as
$g_{\phi_r}$ parameterized by $\phi_r$.
Denote $z_n$ as a dimension reduced feature embedding from original input $\mathcal{D}_n^{tr}$.
Distribution of $z_n$ is a Gaussian distribution:
\begin{eqnarray}
&z_n\sim q(z_n|\mathcal{D}_n^{tr})
=\mathcal{N}(\mu_n^e,\text{diag}(\sigma_n^{e2})),
\nonumber\\
&\text{where}~\mu_n^e,\sigma_n^e=\frac{1}{NK^2}\sum_{k_n=1}^K\sum_{m=1}^N
\sum_{k_m=1}^K 
g_{\phi_r} \left[f_{\phi_e}(x_n^{k_n}),
f_{\phi_e}(x_m^{k_m})\right].\nonumber
\end{eqnarray}
Encoder module is a mapping from original high-dimensional input data to feature embedding.
Decoder function is denoted as
$f_{\phi_d}:\mathcal{Z}\to\Theta$ parameterized by $\phi_d$
where $\mathcal{Z}$ is of lower dimension than $\Theta$. 
Denote $w_n$ as a sample from the decoded distribution, which is also a Gaussian distribution:
\begin{eqnarray}
w_n\sim p(w|z_n)=\mathcal{N}(\mu_n^d,\text{diag}(\sigma_n^{d2})),
\text{where}~\mu_n^d,\sigma_n^d=f_{\phi_d}(z_n).\nonumber
\end{eqnarray}
Decoder module is a mapping from feature embedding to a vector of probabilities for belonging to each class.
In the base learner within inner loop, loss function of LEO for task $\mathcal{T}_i$ is 
\begin{eqnarray}
\mathcal{L}_{\mathcal{T}_i}^{tr}(\phi_i)=\sum_{(\pmb{x},y)\in\mathcal{D}^{tr}}
\left[-w_y\cdot \pmb{x}+\log \left(\sum_{j=1}^N e^{w_j\cdot \pmb{x}}\right)\right],
\nonumber
\end{eqnarray}
where $\phi_i=(\phi_{ei},\phi_{di},\phi_{ri})$.
Apply MAML in inner loop to update feature embedding $z_n$
\begin{eqnarray}
z_n^\prime =z_n-\alpha \bigtriangledown_{z_n}\mathcal{L}_{\mathcal{T}_i}^{tr}.
\nonumber
\end{eqnarray}

In the meta-learner within outer loop, use meta-training data to update the parameters in encoder, relation net and decoder:
\begin{eqnarray}
\min_{\phi_e,\phi_r,\phi_d}\sum_{\mathcal{T}_i\sim p(\mathcal{T})}[\mathcal{L}^{val}_{\mathcal{T}_i}(\phi_i)+\beta D_{KL}(q(z_n|\mathcal{D}_n^{tr})\Vert p(z_n))
+\gamma\Vert \text{stopgrad}(z_n^\prime)-z_n\Vert_2^2]\nonumber\\
+\lambda_1\left(\Vert\phi_e\Vert_2^2+\Vert\phi_r\Vert_2^2+\Vert\phi_d\Vert_2^2\right)+\lambda_2\Vert\mathcal{C}_d-\mathcal{I}\Vert_2,
\nonumber
\end{eqnarray}
where meta-parameters $\phi_e,\phi_r,\phi_d$ are the initial values of task-specific parameters $\phi_i=(\phi_{ei},\phi_{di},\phi_{ri})$. Besides minimization of model generalization error, meta-objective also puts regularization constraints upon meta-parameters to avoid overfitting. $\mathcal{C}_d$ is the correlation matrix of rows in decoder parameter matrix $\phi_d$.
LEO is a highly integrated meta-learning method based upon base learner and meta-learner. From LEO, we know that base learner model can be of any complexity. Meta-learner makes the whole system much more efficient by training meta-parameters and accelerating adaptation of base learner to unseen tasks. The setup of base learner model may be complex but the complex parameter updates are all in the meta-learner.

\begin{table}[htpb]
	\centering
	\caption{Testing accuracy of layered meta-learning methods on 5-way 5-shot miniImageNet classification.}
	\label{tabmetalearner}
	\begin{tabular}{|m{10cm}|m{2.3cm}|}
		\hline
		Method & Accuracy \\ \hline
		\cite{Finn2017} MAML  & $63.11\pm 0.92\%$ \\ \hline
		\cite{Li2017} Meta-SGD  & $64.03 \pm 0.94\%$  \\ \hline
		\cite{Nichol2018a} Reptile without transduction  & $61.98 \pm 0.69\%$ \\ \hline
		\cite{Nichol2018a} Reptile with transduction  & $66.00 \pm 0.62\%$  \\ \hline
		\cite{Ravi2019a} Meta-LSTM  & $60.60 \pm 0.71\%$  \\ \hline
		\cite{Kim2018} Auto-Meta (without Transduction F=10)   & $65.09 \pm 0.24\%$  \\ \hline
		\cite{Kim2018} Auto-Meta (with Transduction F=12)   & $74.65 \pm 0.19\%$  \\ \hline
		\cite{Bertinetto2019} R2-D2  & $68.4\pm 0.2\%$  \\ \hline
		\cite{Bertinetto2019} LR-D2 with 5 iterations  & $68.7\pm 0.2\%$  \\ \hline
		\cite{Bertinetto2019} LR-D2 with 1 iteration  & $65.6\pm 0.2\%$  \\ \hline
		\cite{Lee2019} MetaOptNet-RidgeReg  & $77.88 \pm 0.46\%$  \\ \hline
		\cite{Lee2019} MetaOptNet-SVM  & $78.63 \pm 0.46\%$  \\ \hline
		\cite{Lee2019} MetaOptNet-SVM-trainval  & $80.00 \pm 0.45\%$  \\ \hline
		\cite{Liu2019} TPN  & $69.86\pm 0.65\%$  \\ \hline
		\cite{Rusu2019} LEO  & $77.59 \pm 0.12\%$  \\ \hline
	\end{tabular}
\end{table}

Table \ref{tabmetalearner} summarizes testing accuracy of layered meta-learning methods on 5-way 5-shot miniImageNet classification.
We can see that MetaOptNet shows highest accuracy. With better feature extraction module applied, better trained similarity metrics, more discriminative classification loss function, all methods get better predictive performance. Comparison of few-shot image classification accuracy here is not strictly based upon the same baseline feature extraction module / trained similarity metric / discriminative classification loss function etc. Comparison listed here is rough and not exact. Higher prediction accuracy does not necessarily indicate that one line of meta-learning methodology is superior to another. In real meta-learning applications, all these lines of meta-learning methodologies are integrated.
Prediction accuracy of highly integrated LEO is next to MetaOptNet. Combination of statistical models and machine learning methods improves model generalization capability, since statistical models can adjust for multiplicity in highly correlated features, adjust for noise in features and avoid overfitting. More integrated and sophisticatedly designed meta-learning frameworks demonstrate better prediction accuracy in few-shot image classification.

\subsection{Bayesian Meta-Learning}
\label{Bayesian-Meta-learning}


From frequentist perspective, meta-parameter $\theta$ and task-specific parameter $\phi$ are regarded as unknown fixed values of interest to be predicted. Under the Bayesian framework, both $\theta$ and $\phi$ are treated as random variables. Random variables have distributions and distributions may be represented by samples. Updating meta-parameters and task-specific parameters equals updating their distributions or updating samples from their distributions. Distributions or samples present much more information about unknown parameters than just an estimated value. From distributions or samples, we can view that parameter values lie in which region with highest probability, and how much uncertainty the parameter contains.
In Bayesian meta-learning, we compute the posterior distributions of model parameters and provide inference upon the predictions on out-of-distribution tasks. In few-shot tasks, task-specific parameters are updated using few training data and might be subject to large uncertainty. As a result, introducing Bayesian thinking into few-shot meta-learning framework is necessary to provide uncertainty estimation for predicted labels and parameters of interest. At least three Bayesian variations of MAML are presented in this section, since MAML is popular and there have been many literature surrounding extensions of MAML.

Human learns new concepts through creative thinking and meta cognition by linking data from several sources to create understanding about a new concept.
In few-shot tasks where training data is limited, human mind can perform data augmentation by brainstorm and free imagnination to link limited data to previous experiences and create psuedo simulated augmented data. 
Generative model under Bayesian framework simulates innovative imagination in human mind to learn new concepts. This subclass of methodology focuses upon integration of a generative model into an existing meta-learning framework.
In \cite{Lake2015}, a generative model \textbf{BPL} (Bayesian program learning) is proposed to classify hand-written characters using images and strokes data. 
A Bayesian generative model is integrated into deep Siamese convolutional network.  
Generated hand-written characters are indistinguishable from real hand writings using visual checks.


Character type is denoted as $\theta$.
A set of $M$ parameters from these types is denoted as
$\Phi=(\phi_1,\cdots,\phi_M)$
corresponding to binary images $\mathcal{I}=(I_1,\cdots,I_M)$. The joint distribution of character types, image-specific parameters and images is
\begin{eqnarray}
P(\theta,\Phi,\mathcal{I})
=P(\theta)\prod_{m=1}^M P(I_m|\phi_m) P(\phi_m|\theta).
\nonumber
\end{eqnarray}
Based upon the joint distribution, we compute a discrete approximation to the joint posterior distribution $P(\theta,\phi_m|I_m)$ which is updated for adaptation to different tasks.
Under Bayesian framework, we work from the joint distribution of data and parameters. Under proper independence and conditional independence assumptions between data and parameters, joint distribution can be decomposed into important conditional distributions. We aggregate all task training data to get accurate empirical estimates of all these important conditional distributions. Based upon estimated conditional distributions, we simulate samples that look like real data and add them into few-shot tasks for data augmentation. Though the process of data simulation does not create any additional information, data augmentation does contribute to improvement in prediction accuracy in few-shot image classification tasks.

\textbf{Neural Statistician} method \cite{Edwards2016} also contains a generative model but this generative model is used to approximate posterior distributions of key parameters. Bayesian thinking integrated within meta-learning algorithm can provide efficient uncertainty estimation for important parameters and unknown labels. Bayesian variational inference is a key methodology for efficient inference. Bayesian variational inference provides an efficient approximation to posterior distribution and uses the approximated posterior to estimate parameter uncertainty.
For task $\mathcal{T}_i$, a task-specific generative model $\hat{p}_i$ is estimated. 
The generative distribution is defined as
$p_i=p(\cdot|c_i)$, where 
$c_i$ is the task-specific context.
Then the
approximate posterior of task-specific context $q(c|\mathcal{D})$ is computed using variational autoencoder (VAE).

In VAE,
decoder with meta-parameter $\theta$ is denoted as
$p(\mathcal{D}|c;\theta)$ and encoder with task-specific parameter $\phi$ is denoted as
$q(c|\mathcal{D};\phi)$. Decoder $p(\mathcal{D}|c;\theta)$ is also the probability distribution of data which is equal to the value of likelihood function. Encoder $q(c|\mathcal{D};\phi)$ is also the posterior distribution of context $c$. Context $c$ can be viewed as data label or data class.  For adaptation to different tasks, posterior distribution of $c$ modelled as an inference network $q(c|\mathcal{D};\phi)$ is updated.
The standard variational lower bound of log likelihood function is
\begin{eqnarray}
\mathcal{L}=\mathbb{E}_{q(c|\mathcal{D};\phi)}
[\log p(\mathcal{D}|c;\theta)]+\alpha D_{KL}(q(c|\mathcal{D};\phi)\Vert p(c)).
\nonumber
\end{eqnarray}
In maximum likelihood estimation, we maximize log likelihood function to estimate key parameters. In VAE, we maximize standard variational lower bound of log likelihood function to estimate $\theta$ and $\phi$. It is assumed that log likelihood function is close to global optimum when its variational lower bound is maximized. We maximize variational lower bound instead of log likelihood, as a result, estimated posterior distribution is only an approximation to true posterior.
The instance encoder $f$ is a feedforward neural network for feature extraction such that
$z_i=f(x_i)$.
In exchangeable instance pooling layer, the mapping is from encoded features to a pooled vector:
$[z_1,\cdots,z_k]\to v$.
In post-pooling network, the mapping is from a pooled vector to a decoded distribution:
$v\to$ a diagonal Gaussian (decoder).
After estimating an approximated posterior distribution $q(c|\mathcal{D};\phi)$,
predicted label is $\text{argmax}_c q(c|\mathcal{D};\phi)$ where posterior is at its maximum.


The following three papers extend MAML to Bayesian framework.
First, \cite{Grant2018} proposes 
\textbf{LLAMA} (Lightweight Laplace Approximation for Meta-Adaptation) which formulates MAML as probabilistic inference in a hierarchical Bayesian model.
It considers first-order and second-order Laplace approximations of log likelihood function to construct inference with quadratic curvature estimation.
In a hierarchical Bayesian model,  all marginal distributions and conditional distributions are specified to be Gaussian. Log likelihood function is represented as an integral of Gaussian density kernel. Laplace approximation uses point estimation to approximate the integral in log likelihood function. Gaussian distribution is tight and symmetric, and its integral can be well approximated by a point estimate in Laplace approximation. With Laplace approximation, log likelihood can be computed more efficiently, consequently computation of maximum log likelihood is also more efficient.

In task $\mathcal{T}_j$, training data is denoted as $\pmb{x}_j=(x_{j1},\cdots,x_{jn})$, and
validation data is denoted as 
$\pmb{x}_j^*=(x^*_{j1},\cdots,x^*_{jm})$.
For probabilistic inference,
marginal likelihood of observed data from $J$ tasks
$\pmb{x}=(\pmb{x}_1,\cdots,\pmb{x}_J)$ can be written as
\begin{eqnarray}
&p(\pmb{x}|\theta)=\prod_{j=1}^J \left\{\int p(\pmb{x}_j|\phi_j)
p(\phi_j|\theta)d\phi_j\right\},
\nonumber
\end{eqnarray}
where $\phi_j$ is task-specific parameter and $\theta$ is meta-parameter.
In meta-learner, meta-parameter $\theta$ is 
estimated using maximum likelihood
$\max_\theta p(\pmb{x}|\theta)$.
In base learner, task-specific parameter $\phi_j$
for task $\mathcal{T}_j$ can be estimated with MAML where task-specific parameter $\phi_j$ is estimated using one SGD iteration from meta-parameter $\theta$:
\begin{eqnarray} 
&\hat{\phi}_j=\theta+\alpha\bigtriangledown_\theta\log p(\pmb{x}_j|\theta).
\nonumber
\end{eqnarray}
In \cite{Grant2018}, it is shown that update of task-specific parameter $\phi_j$ is equivalent to estimating $\phi_j$ using a maximum a posteriori (MAP) estimator.
MAP estimator is to derive the posterior distribution of $\phi_j$ and MAP is where the posterior is at its global maximum.
From Bayesian perspective, posterior distribution of $\phi_j$ is
$p(\phi_j|\pmb{x}_{j},\theta)
\propto p(\pmb{x}_{j}|\phi_j)
p(\phi_j|\theta)$,
where $p(\phi_j|\theta)$ is the prior distribution of task-specific parameter and $p(\pmb{x}_{j}|\phi_j)$ is the distribution of task training data $\pmb{x}_{j}$. Maximum a posteriori estimate of $\phi_j$ 
is the global mode of $p(\phi_j|\pmb{x}_{j},\theta)$:
\[
\hat{\phi}_j=\text{argmax}_{\phi_j} p(\phi_j|\pmb{x}_{j},\theta).
\]
Based upon first-order Laplace approximation, log likelihood function of validation data $\pmb{x}^*_j$ is approximated with
\begin{eqnarray}
&\log p(\pmb{x}|\theta)\approx \sum_{j=1}^J
\left\{\log p\left(\pmb{x}^*_j \vert \hat{\phi}_j\right)\right\},
\nonumber
\end{eqnarray}
where approximate log likelihood for task $\mathcal{T}_j$ is given by
$
\mathbb{E}_{\pmb{x}^*_j}[\log p(\pmb{x}^*_j|\phi_j)]\approx
m^{-1}\sum_{d=1}^m \log p(x^*_{jd}|\hat{\phi}_j).
$

Second-order Laplace approximation to likelihood function of task $\mathcal{T}_j$ data is 
\begin{eqnarray}
p(\pmb{x}_j|\theta)=\int p(\pmb{x}_j|\phi_j)p(\phi_j|\theta)d\phi_j
\approx p(\pmb{x}_j|\hat{\phi}_j)p(\hat{\phi}_j|\theta)
\text{det}(H_j/2\pi)^{-1/2},
\nonumber
\end{eqnarray}
where $H_j$ is the
Hessian of log likelihood function:
\begin{eqnarray}
H_j=\bigtriangledown_{\phi_j}^2[\log p(\pmb{x}_j|\phi_j)]+\bigtriangledown_{\phi_j}^2
[\log p(\phi_j|\theta)].
\nonumber
\end{eqnarray}
From second-order Laplace approximation,
first-order approximation $\log p(\pmb{x}_j|\phi_j)$ is replaced with second-order approximation
$\log p(\pmb{x}_j|\phi_j)-\eta\log\text{det}(\hat{H}_j)$. Instead of just a point estimate in first-order approximation, second-order approximation considers an additional piece ie Hessian matrix. 
Both first-order and second-order Laplace approximation of log likelihood function can be applied to update task-specific parameters in fast model adaptation to unseen tasks. Hessian matrix takes second-derivative of log likelihood function and requires much more computation than first-order Laplace approximation. We may assume Hessian matrix to be diagonal, block-wise diagonal, sparse etc to accelerate its estimation. Laplace only works for tightly distributed data and parameters. For multi-modal distribution, highly skewed distribution, fat-tailed distributions etc, their integrals cannot be estimated using a point estimator and Laplace approximation does not work well for them.

Another Bayesian extension of MAML is \textbf{BMAML}
(Bayesian Model-Agnostic Meta-Learning) proposed
in \cite{Yoon2018a}, where stochastic gradient descent (SGD) in base learner is replaced with Stein variational gradient descent (SVGD) method \cite{Liu2016,Liu2017}. SVGD is an efficient sampling method which is a combination of MCMC and variational inference. Both MCMC and Bayesian variational inference are popular inference methods under Bayesian framework. MCMC relies upon obtaining samples from target distribution to provide uncertainty estimate of key parameters. Bayesian variational inference maximizes variational lower bound of log likelihood to estimate an approximate posterior distribution of key parameters. Uncertainty estimate and confidence interval are derived from the approximate posterior distribution. Bayesian variational inference is more efficient than MCMC since it does not require sampling or convergence. MCMC is based upon sampling and works for a wider class of data and parameter distributions, such as multi-modal distribution, highly skewed distribution, fat-tailed distributions etc. SVGD inherits advantages and avoids disadvantages of both MCMC and Bayesian variational inference.
We first draw a sample  $\Theta_0=\{\theta^0_i\}_{i=1}^n$ from target distribution $p(\theta)$ as the initial particles.
In the $l$th iteration, particles $\Theta_l=\{\theta^l_i\}_{i=1}^n$ are updated using
\begin{eqnarray}
g(\theta)=n^{-1}\sum_{j=1}^n\left\{k(\theta_j^l,\theta)\bigtriangledown_{\theta_j^l}\log p(\theta_j^l)
+\bigtriangledown_{\theta_j^l}k(\theta_j^l,\theta)\right\},\nonumber\\
\theta_i^{l+1}\leftarrow \theta_i^l+\epsilon_l g(\theta_i^l),~\text{for}~
i=1,\cdots,n,
\nonumber
\end{eqnarray}
where 
$\epsilon_l$ is the step size,
$k(\cdot,\cdot)$ is a positive kernel function and
$\bigtriangledown_{\theta_j^l}k(\theta_j^l,\theta)$ is a repulsive force so that $\theta_i^{l+1}\neq \theta_i^l$. In order for the samples to reach as much region as possible in target distribution $p(\theta)$, samples are expected to be farther away from each other and do not collapse to be around the same spot so that samples better represent properties of $p(\theta)$.
During meta-training process, we
update these particles to update posterior distribution of task-specific parameters.
In BMAML, we use this SVGD procedure to update samples rather than use SGD to update parameter values. SVGD requires that target distribution $p(\theta)$ is differentiable. SGD requires that loss function is differentiable. BMAML extends MAML in the sense that MAML is applicable to cases optimized with SGD and BMAML is applicable to cases optimized with SVGD.

In base learner, Bayesian Fast Adaptation (BFA) is performed, where we update
task-specific parameter particles ensemble $\Phi_i=\{\phi_i^m\}_{m=1}^M$ for task $\mathcal{T}_i$ using SVGD based likelihood approximation:
\[
\Phi_i\leftarrow \text{SVGD}_n(\Theta_0,\mathcal{D}_i^{tr},\alpha),
\]
where after $n$ SVGD iterations, particles ensemble representing distribution of task-specific parameter is updated to be $\Phi_i$. Hyperparameters controlling each SVGD iteration is denoted as $\alpha$.
The contribution of task $\mathcal{T}_i$ to meta-objective is to maximize its generalization capability
\begin{eqnarray}
&p(\mathcal{D}_i^{val}|\Theta_0,\mathcal{D}_i^{tr})=\int p(\mathcal{D}_i^{val}|\Theta_0,\Phi_i)p(\Phi_i|\Theta_0,\mathcal{D}_i^{tr})d\Phi_i  \approx M^{-1}\sum_{m=1}^M p(\mathcal{D}_i^{val}|\Theta_0,\phi^m_i),
\nonumber
\end{eqnarray}
where $\phi^m_i\sim p(\Phi_i|\Theta_0,\mathcal{D}_i^{tr})$ and $\Theta_0$ is the initial particle ensemble to generate $\Phi_i$ using SVGD.
After a batch of tasks $\mathcal{T}_1,\cdots,\mathcal{T}_J$ are trained, meta-parameter $p(\theta)$ in meta-learner is updated. In one way, $\theta$ can be updated through minimization of generalization error. In another, initial particles ensemble $\Theta_0$ may be updated using SVGD.
Meta-objective in meta-learner is to maximize
\begin{eqnarray}
&\log p(\mathcal{D}^{val}|\Theta_0,\mathcal{D}^{tr})
\approx \sum_{i=1}^J \mathcal{L}_{BFA}(\Phi_i;\mathcal{D}^{val}_i),\nonumber\\
&\text{where}~\mathcal{L}_{BFA}(\Phi_i;\mathcal{D}^{val}_i)\approx
\log \left\{M^{-1}\sum_{m=1}^M p(\mathcal{D}^{val}_i|\Theta_0,\phi^m_i)\right\}.
\nonumber
\end{eqnarray}

In base learner, for task $\mathcal{T}_i$, task-specific particle ensemble $\Phi_i$ is updated using
$\Phi_i\leftarrow \text{SVGD}(\Theta_0;\mathcal{D}^{tr}_i,\alpha)$,
where $\alpha$ is the hyperparameters in SVGD iteration such as step size or learning rate etc.
In meta-learner, meta-parameter particles ensemble initialization $\Theta_0$ is evaluated using
\begin{eqnarray}
&\Theta_0\leftarrow\Theta_0+\beta\bigtriangledown_{\Theta_0}\left[\sum_{i=1}^J \mathcal{L}_{BFA}(\Phi_i;\mathcal{D}^{val}_i)\right],
\nonumber
\end{eqnarray}
where initial particles ensemble is updated using SVGD. It may be better represented as
\[
\Theta_0\leftarrow \text{SVGD}(\Theta_0;\mathcal{D}_i^{val},\beta).
\]
Meta-learner described above is similar to meta-learner of MAML.
BMAML presents another way of defining meta-objective function: a chaser loss meta-objective.
Meta-objective function is specified as the dissimilarity between approximate and true posterior distributions of task-specific parameter $\Phi_i$.
Posterior of task-specific parameter after $n$ SVGDs from initial particles ensemble $\Theta_0$ is denoted as
$p_i^n\equiv p_n(\Phi_i|\mathcal{D}_i^{tr};\Theta_0)$,
from which we randomly sample $\Phi^n_i$.
True task-specific parameter posterior is
$p_i^\infty\equiv p(\Phi_i|\mathcal{D}_i^{tr}\cup \mathcal{D}_i^{val})$,
from which we randomly sample $\Phi_i^\infty$.
BMAML meta-learner seeks the minimization of the dissimilarity between estimated approximate posterior and true posterior:
\begin{eqnarray}
&\min_{\Theta_0} \sum_{i=1}^J d_p (p_i^n\Vert p_i^\infty)\approx \min_{\Theta_0} \sum_{i=1}^J d_s(\Phi^n_i\Vert \Phi_i^\infty),
\nonumber
\end{eqnarray}
where 
$d_p$ and $d_s$ are the distance measures between two distributions and two particles ensembles respectively.
For definition of chaser loss, true posterior
$\Phi_i^\infty$ is approximated with a posterior estimate after $s$ more SVGD iterations $\Phi_i^{n+s}$, where $s>0$.
Chaser is defined as
$\Phi_i^n=\text{SVGD}_n(\Theta_0;\mathcal{D}_i^{tr},\alpha)$.
Leader is defined as
$\Phi_i^{n+s}=\text{SVGD}_s(\Phi_i^n;\mathcal{D}_i^{tr}\cup \mathcal{D}_i^{val},\alpha)$.
Meta-objective in BMAML based upon chaser loss is 
\begin{eqnarray}
&\mathcal{L}_{BMAML}(\Theta_0)=
\sum_{i=1}^J d_s(\Phi_i^n\Vert\Phi_i^{n+s}).\nonumber
\end{eqnarray}
Finally, meta-parameter $\Theta_0$ is updated using
$\Theta_0\leftarrow \Theta_0-\beta\bigtriangledown_{\Theta_0} \left[\mathcal{L}_{BMAML}(\Theta_0)\right]$, where $\bigtriangledown_{\Theta_0}$ is also SVGD iteration rather than SGD iteration.
Prediction accuracy of BMAML is slightly better than LLAMA in 5-way 1-shot miniImageNet classification. However, LLAMA is more lightweight than BMAML.

Third, \textbf{PLATIPUS}
(Probabilistic LATent model for Incorporating Priors and Uncertainty in few-Shot learning
based upon MAML) \cite{Finn2018} is also an extension to MAML under Bayesian framework. PLATIPUS constructs a Bayesian network for data and parameters and utilizes Bayesian variational inference to estimate uncertainty in key parameters. Through maximizing variational lower bound of log likelihood function, PLATIPUS computes approximate posterior distributions of task-specific parameters and meta-parameters. MAP estimates at the global maximum of approximate posterior distributions are used to update task-specific parameters and meta-parameter.
Training data is
$\{(\pmb{x}_1,y_1),\cdots,(\pmb{x}_n,y_n)\}$.
Validation data is
$\{(\pmb{x}^*_1,y^*_1),\cdots,(\pmb{x}^*_m,y^*_m)\}$.
For task $\mathcal{T}_i$, inference network is
$q_i(\theta,\phi_i)=q_i(\theta)q_i(\phi_i|\theta)$.
Conditional distribution is specified as 
$q_i(\phi_i|\theta)=q_{\psi}(\phi_i|\theta,\pmb{x}_i,y_i,\pmb{x}^*_i,y^*_i)$ and prior distribution of $\theta$ is 
$q_i(\theta)=q_{\psi}(\theta|\pmb{x}_i,y_i,\pmb{x}^*_i,y^*_i)$. Both components in inference network are parameterized by $\psi$. These two inference networks are used to model approximate posterior distributions of key parameters.  All marginal and conditional distributions are specified to be Gaussian. 
Variational lower bound of log likelihood is
\begin{eqnarray}
&\log p(y_i^*|\pmb{x}^*_i,\pmb{x}_i,y_i)\geq
\mathbb{E}_{\theta,\phi_i\sim q_\psi}
[\log p(y_i|\pmb{x}_i,\phi_i)\nonumber
\\
&+\log p(y^*_i|\pmb{x}^*_i,\phi_i)
+\log p(\phi_i|\theta)
+\log p(\theta)]\nonumber\\
&+\mathcal{H}(q_{\psi}(\phi_i|\theta,\pmb{x}_i,y_i,\pmb{x}^*_i,y^*_i))+\mathcal{H}(q_{\psi}(\theta|\pmb{x}_i,y_i,\pmb{x}^*_i,y^*_i)),\nonumber
\nonumber
\end{eqnarray}
where $\mathcal{H}$ is regularization upon the approximate distributional model, such that its complexity is moderate and it's reasonably close to its true distribution etc.
These inference networks are posited to be Gaussian distributions
\begin{eqnarray}
q_{\psi}(\theta|\pmb{x}_i,y_i,\pmb{x}^*_i,y^*_i)
=\mathcal{N}(\mu_\theta+\gamma_p\bigtriangledown_{\mu_\theta}\log p(y_i|\pmb{x}_i,\mu_\theta)
\nonumber\\+\gamma_q\bigtriangledown_{\mu_\theta}\log p(y^*_i|\pmb{x}^*_i,\mu_\theta),v_q).\nonumber
\end{eqnarray}

In PLATIPUS, we first
initialize all meta-parameters $\Theta=\{\mu_\theta,\sigma^2_\theta,v_q,\gamma_p,\gamma_q\}$ in inference networks.
%
%
Based upon these inference networks, we
sample meta-parameters
$\theta\sim q_\psi=\mathcal{N}(\mu_\theta-\gamma_q\bigtriangledown_{\mu_\theta}\mathcal{L}(\mu_\theta,\mathcal{D}^{test}),v_q)$.
%
%
Based upon MAML, task-specific parameter in the inner loop is
updated using
$\phi_i=\theta-\alpha\bigtriangledown_\theta \mathcal{L}(\theta,\mathcal{D}^{tr})$.
Gaussian inference network is specified as $p(\theta|\mathcal{D}^{tr})=\mathcal{N}(\mu_\theta-\gamma_p\bigtriangledown_{\mu_\theta}\mathcal{L}(\mu_\theta,\mathcal{D}^{tr}),\sigma^2_\theta)$.
Finally, meta-parameters $\Theta$ are updated using ADAM
\begin{eqnarray}
&\bigtriangledown_\Theta \left\{\sum_i \mathcal{L}(\phi_i,\mathcal{D}^{test})+D_{KL}(q(\theta|\mathcal{D}^{test})\Vert p(\theta|\mathcal{D}^{tr}))\right\}.\nonumber
\end{eqnarray}
Prediction accuracy of PLATIPUS in 5-way 1-shot miniImageNet classification is comparable to LLAMA and BMAML which are also Bayesian extensions of MAML.

\textbf{VERSA} (Versatile and Efficient Amortization of Few-Shot Learning) \cite{Gordon2018} uses Bayesian decision theory (BDT) in integration with Bayesian variational inference to estimate approximate posterior distribution of unknown label. Predicted label is where approximate posterior distribution of unknown label is at its global maximum. We can estimate the interval where unknown label has the highest probability to be. We can also estimate the uncertainty in unknown label using approximate posterior distribution.
Training data is 
$\{(\pmb{x}_k,y_k)\}_{k=1}^n$.
Validation data is 
$\{(\pmb{x}^*_k,y^*_k)\}_{k=1}^m$.
Task data is the union of training data and validation data denoted by
$\{(\pmb{x}_k,y_k)\}_{k=1}^N$.
Task training data of task $\mathcal{T}_i$ is $(\pmb{x}_i,y_i)$.
Task validation data of task $\mathcal{T}_i$ is $(\pmb{x}^*_i,y^*_i)$.
Meta-parameter is $\theta$ and
task-specific parameter for task $\mathcal{T}_i$ is $\phi_i$.
Joint distribution of all data and all parameters is 
\begin{eqnarray}
p\left(\{y_i,\phi_i\}_{i=1}^N |\{\pmb{x}_i\}_{i=1}^N, \theta\right)=\prod_{i=1}^N \left.p(\phi_i|\theta)\prod_{j=1}^n
p(y_j|\pmb{x}_j,\phi_j,\theta)\prod_{k=1}^m p(y^*_k|\pmb{x}_k^*,\phi_k,\theta)\right..
\nonumber
\end{eqnarray}
Joint distribution of all data and all parameters is decomposed into a multiplication of three parts: conditional distribution of parameters, distribution of task training data, distribution of task validation data.
For neural network base learners, task-specific parameters $\phi_i=\{W_i,b_i\}$ are composed of weights and biases.
In task testing / validation data,
predicted label is $\hat{y}$ and
unknown true label is $\tilde{y}$.
Unknown label is predicted with
\begin{eqnarray}
\min_{\hat{y}}
\int \mathcal{L}(\tilde{y},\hat{y}) p(\tilde{y}|\mathcal{D}_i) d\tilde{y},
\nonumber
\end{eqnarray}
where $\mathcal{D}_i$ is all data in task $\mathcal{T}_i$, and $\mathcal{L}(\tilde{y},\hat{y})$ is the distance between predicted and true label.
Posterior distribution of unknown true label is
\begin{eqnarray}
&p(\tilde{y}|\mathcal{D}_i)
=\int p(\tilde{y}|\phi_i)p(\phi_i|\mathcal{D}_i)
d\phi_i,
\nonumber
\end{eqnarray}
where  $p(\phi_i|\mathcal{D}_i)$ is the posterior distribution of task-specific parameter.
In distributional BDT, true label $\tilde{y}$ follows distribution $q(\tilde{y})$ and the objective is to estimate a predictive distribution of unknown label:
\begin{eqnarray}
\min_{q\in\mathcal{Q}}\int 
\mathcal{L}(\tilde{y},q(\tilde{y}))
p(\tilde{y}|\mathcal{D}_i)  d\tilde{y},
\nonumber
\end{eqnarray}
where predictive label distribution $q$ lies within a pre-specified distribution family $\mathcal{Q}$.
Amortized variational training is applied to make estimation based upon distributional BDT.
Distribution of unknown label is specified as
$q(\tilde{y})=q_\phi (\tilde{y}|\mathcal{D})$ where $\phi$ is the unknown parameter of interest.
Parameter estimator in predictive label distribution is
\begin{eqnarray}
\min_\phi 
\mathbb{E}_{p(\mathcal{D},\tilde{y})}\mathcal{L}(\tilde{y},q_\phi (\tilde{y}|\mathcal{D})).
\nonumber
\end{eqnarray}
Loss function $\mathcal{L}$ is specified as negative log likelihood:
\begin{eqnarray}
\mathcal{L}(q_\phi)=\mathbb{E}_{p(\mathcal{D},\tilde{y})} [-\log q_\phi (\tilde{y}|\mathcal{D})].
\nonumber
\end{eqnarray}
Minimization of loss function $\mathcal{L}$ maximizes the probability of data being in its true category.
Variational lower bound of log likelihood function used to estimate unknown label is
\begin{eqnarray}
\mathbb{E}_{p(\mathcal{D},\tilde{y})}
[D_{KL}\{p(\tilde{y}|\mathcal{D})\Vert q_\phi (\tilde{y}|\mathcal{D})\}]+\mathcal{H}[q_\phi (\tilde{y}|\mathcal{D})].
\nonumber
\end{eqnarray}
Prediction based upon distributional Bayesian decision theory and amortized variational inference
is precise. 


\begin{table}[htpb]
	\centering
	\caption{Testing accuracy of meta-learning methods on 5-way 1-shot miniImageNet classification.}
	\label{tabbayes}
	\begin{tabular}{|m{10cm}|m{2.3cm}|}
		\hline
		Method & Accuracy \\ \hline
		\cite{Grant2018} LLAMA  & $49.40\pm 1.83\%$  \\ \hline 
		\cite{Yoon2018a} BMAML  & $53.8\pm 1.46\%$  \\ \hline 
		\cite{Finn2018} PLATIPUS  & $50.13\pm 1.86\%$ \\ \hline
		\cite{Gordon2018} VERSA  & $53.40\pm 1.82\%$  \\ \hline\hline
		\cite{Qiao2018} Activation to parameter with neural network  & $54.53 \pm
		0.40\%$  \\ \hline
		\cite{Qiao2018} wide residual network WRN  & $59.60 \pm 0.41\%$  \\ \hline
		\cite{Munkhdalai2018} AdaCNN with DF & $48.34 \pm 0.68\%$  \\ \hline
		\cite{Munkhdalai2018} AdaResNet with DF & $56.88 \pm 0.62\%$  \\ \hline\hline
		\cite{Vinyals2016} Matching Net & $46.6\%$  \\ \hline
		\cite{Snell2017} Prototypical Net  & $46.61\pm 0.78\%$ \\ \hline
		\cite{Das2020} PVRT & $52.68\pm 0.51\%$ \\ \hline
		\cite{Li2020} Prototypical Network+TRAML & $60.31\pm 0.48\%$ \\ \hline
		\cite{Sung2018} Relation Net & $50.44\pm 0.82\%$ \\ \hline 
		\cite{Mishra2018} SNAIL & $45.1\%$  \\ \hline
		\cite{Oreshkin2018} TADAM with $\alpha$, AT and TC  & $58.5 \pm 0.3\%$ \\ \hline
		\cite{Oreshkin2018} TADAM without tuning  & $56.5 \pm 0.4\%$  \\ \hline
		\cite{Gidaris2018} Dynamic few-shot with C128F feature extractor  & $55.95 \pm 0.84\%$ \\ \hline
		\cite{Gidaris2018} with ResNet feature extractor  & $55.45 \pm 0.89\%$  \\ \hline
		\cite{Gidaris2018} with cosine classifier and attention based weight generator  & $58.55 \pm 0.50\%$  \\ \hline
		\cite{Gidaris2018} with cosine classifier and no weight generator  & $54.55 \pm 0.44\%$  \\ \hline
		\cite{Triantafillou2017} mAP-SSVM & $50.32\pm 0.80\%$  \\ \hline
		\cite{Triantafillou2017} mAP-DLM & $50.28\pm 0.80\%$ \\ \hline\hline
		\cite{Finn2017} MAML & $48.7\pm 1.84\%$ \\ \hline
		\cite{Munkhdalai2017} MetaNet & $49.21 \pm 0.96\%$ \\ \hline
		\cite{Nichol2018} Reptile & $49.97\pm 0.32\%$ \\ \hline
		\cite{Li2017} Meta-SGD & $50.47\pm 1.87\%$ \\ \hline
		\cite{Ravi2019a} Meta-LSTM  & $43.44\pm 0.77\%$ \\ \hline
		\cite{Kim2018} Auto-Meta (without Transduction F=10) & $49.58\pm 0.20\%$ \\ \hline
		\cite{Kim2018} Auto-Meta (with Transduction F=12) & $57.58\pm 0.20\%$ \\ \hline
		\cite{Bertinetto2019} R2-D2  & $51.8\pm 0.2\%$  \\ \hline
		\cite{Bertinetto2019} LR-D2 with 5 iterations  & $51.9\pm 0.2\%$  \\ \hline
		\cite{Bertinetto2019} LR-D2 with 1 iteration  & $51.0\pm 0.2\%$  \\ \hline
		\cite{Lee2019} MetaOptNet-RidgeReg  & $61.41 \pm 0.61\%$  \\ \hline
		\cite{Lee2019} MetaOptNet-SVM  & $62.64 \pm 0.61\%$  \\ \hline
		\cite{Lee2019} MetaOptNet-SVM-trainval  & $64.09 \pm 0.62\%$  \\ \hline
		\cite{Liu2019} TPN  & $55.51\pm 0.86\%$  \\ \hline
		\cite{Rusu2019} LEO  & $61.76 \pm 0.08\%$  \\ \hline
	\end{tabular}
\end{table}

Table \ref{tabbayes} presents testing accuracy of meta-learning methods on 5-way 1-shot miniImageNet classification. We can see that performance of Bayesian meta-learning methods is comparable to black-box adaptation, similarity-based models and layered meta-learning approach. 
Among Bayesian meta-learning methods, BMAML and VERSA show best prediction performance.
Among layered meta-learning models reviewed, MetaOptNet shows highest prediction accuracy. Comparison of these meta-learning methods on few-shot image classification tasks is not based upon the same baseline model such as feature extraction model, discriminative loss function etc. Comparison listed here is rough and not exact. We can see that prediction accuracy is far from $100\%$ and there is room for improvement in few-shot image classification.

\section{Applications of Meta-learning}
\label{applications-meta-learning}

Meta-learning serves as a complementary generalization block for deep-learning models, which have been widely applied to tackle challenging problems and to achieve autonomous AI. Though deep-learning models provide accurate in-sample prediction, it does not guarantee accurate out-of-sample prediction results. Meta-learning, however, concentrates upon out-of-sample generalization of deep-learning models. 
With MAML embedded, base learner is deep model, and meta-learner provides initial values of task-specific parameters in base learner. Meta-learner guides base learner to search for an optimal model training procedure, in which performances upon in-distribution tasks are preserved and performances upon out-of-distribution tasks are improved. Meta-learning is extremely helpful in cases where deep model is required and data amount is little, since deep model contains lots of parameters that cannot be estimated accurately with few-shot data.

First, meta-learning is widely applied in robotics research where robots are expected to reach higher level of autonomy in general AI. Robots acquire basic skills from one or two human or video demonstrations using meta-imitation learning. 
Robots perform decision making and make optimal response to different environments through meta-RL. 
RL model is hard to train and it takes a long time for robots to learn an optimal action strategy from training RL model from scratch. 
Meta-RL model does not train RL model from scratch and it needs only one or two interaction trajectories between agent and slightly different environment to adapt a pre-trained deep RL model to solve the current task.
Imitation learning makes robots learn an optimal policy efficiently by imitating many human demonstrations for an exactly same task. Generally imitation learning achieves the same objective but is more efficient than RL model. Meta imitation learning makes robots learn an optimal policy by imitating only one demonstration under a slightly different task, such as different background color, different light condition, different target shape etc.  

Similarly, for trading robot, with imitation learning, robots can imitate human trading and learn complex trading decision making scheme.
With meta-learning, trading robots can autonomously adapt decision making mechanism to fast changing market conditions continually. Under the same environmental market condition, trading robots learn one behavioral skill from human demonstration through imitation learning. Under a slightly different market condition, trading robots learn from one human demonstration which is obtained under the previous market condition. To be more precise and to perform better, robots should capture complex interaction mechanisms between human behaviors and market conditions. 
Human interacts with market by trading on two alternative assumptions: (1) human trading behaviors affect market conditions, (2) human trading behaviors do not affect market conditions. Market conditions determine the scheme through which human obtains reward, and further determine optimal trading behaviors taken by human.
Under meta imitation scheme, robots imitate human trading strategies under a slightly different background. 
In addition, meta-learning can handle unusual situations in financial market such as sudden market crashes. Proper trading strategies to minimize expected loss can be developed under meta-learning framework. This scheme can be used as a fail-safe and be integrated into the current trading machines, preserving prediction accuracy on in-distribution market conditions and improving prediction accuracy on out-of-distribution market conditions.

Second application of meta-learning is in drug discovery for handling high-dimensional data with small sample size. Sample size in medical studies is usually small especially for rare disease or gene SNPs. Feature dimension in such datasets, on the other hand, is very large. A task is regarded as finding an effective treatment for a virus which might mutate frequently or identifying a set of features that are truly associated with response of interest. Computation for drug candidate is expensive thus adapting previously trained models to new tasks should be considered to save time and expense in drug discovery. Under such situations, deep models may be applied or not applied. But for processing image data, deep models have to be considered in the whole decision system. In such situations of high-dimensional data with small sample size, meta-learning system is helpful in adapting pre-trained deep models to solving unseen tasks.

Third application is in translation of rarely used languages. Rare language may be in these two cases: (1) rarely used words or characters in commonly used languages, (2) rarely used languages. 
Frequently used words constitute only a small proportion of all words.
Most words appear less frequently and their translation depends upon a relatively smaller sample. 
Recent meta-learning models are applicable to few-shot datasets where translation can be learned using adaptation of pre-trained deep models. Though sample size is small for ancient or rare languages, a few-shot meta-learning model can be trained to translate these scripts. Over a sufficiently long time of experience accumulation, meta-learner has stored relevant experiences for rare language translation tasks.

For applications where an explicit definition of task and label can be clearly specified, meta-learning provides a feasible plan for problem-solving. Meta-learning framework is flexible and can be conveniently integrated with most machine learning algorithms to provide feasible solutions.
For tasks which generally require heavy computation, meta-learning presents the option of aggregating or adapting previous results to save computation. For application of meta-learning framework in deep learning models, all components in meta-learning framework including objective functions, task-specific parameters, meta-parameters, task training data, task validation data etc should be explicitly defined under deep learning framework.

\subsection{Meta-Reinforcement Learning}
\label{meta-reinforcement}

Reinforcement learning (RL) is based upon
markov decision process (MDP)
$(\mathcal{S},\mathcal{A},p,r,\gamma,\rho_0,H)$
where $\mathcal{S}$ is the set of all states,
$\mathcal{A}$ is the set of all actions,
$p(s^\prime|s,a)$ is the transition process,
$r(s,a)$ is a reward function,
$\gamma$ is the discount factor on future reward,
$\rho_0(s)$ is the initial state distribution,
and $H$ is the time horizon \cite{Nagabandi2019}.
A trajectory between timestamp $i$ and timestamp $j$ is denoted as
$\tau(i,j)\equiv (s_i,a_i,\cdots,s_j,a_j,s_{j+1})$.
Policy is defined as $\pi:\mathcal{S}\to\mathcal{A}$.
Objective is to identify an optimal policy that maximizes the summation of all rewards from a trajectory.
An RL task \cite{Finn2017} is defined as
\begin{eqnarray}
\mathcal{T}_i\equiv (S_i,A(s),p_i(s),p_i(s^\prime|s,a),r_i(s,a)),
\nonumber
\end{eqnarray}
where $S_i$ is the state space, $A(s)$ is the action space, $p_i(s)$ is the initial state distribution, $p_i(s^\prime|s,a)$ is the transition process and $r_i(s,a)$ is the reward process.
In an RL task, we observe states, actions and find an optimal policy to maximize the estimated sparse reward function. 

\begin{figure}[htpb]
	\centering
	\includegraphics[width=0.8in]{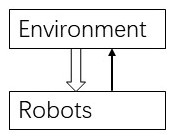}
	\caption{Meta-reinforcement learning. Robots interact with environment to get state and reward. Environment is different after interaction with robots. Environmental conditions vary in different tasks. In few-shot RL tasks, reinforcement learning is integrated into meta-learning framework to train deep RL policy using few-shot data.}
	\label{meta-reinforce}
\end{figure}

Early research on meta-learning \cite{Schmidthuber} is focused upon reinforcement learning settings. Meta-learning is applied to achieve continual learning or lifelong learning for robots in RL settings. Recently meta-learning is applied to train deep RL model when there are not sufficiently large number of observed trajectories in each RL task. 
Meta-reinforcement learning (meta-RL) considers the interaction between agent and changing environment, as in figure \ref{meta-reinforce}. 
Main objective of meta-RL is training robots to handle unusual situations in reward-driven tasks with only one or two observed trajectories. 
Meta-RL concerns data-efficient fast adaptation to out-of-distribution RL tasks after only one or two observed trajectories
\cite{Humplik2019}.
All meta-learning models reviewed in section \ref{meta-learning-models} can be extended to RL tasks. 

\subsubsection{Lifelong RL with Self-Modifications}

\textbf{SSA} (Success-Story Algorithm) \cite{Schmidhuber1998} with SMP (Self-Modifying Policy) is for lifelong RL that continually improves policy.
Action decision points along the whole lifetime are denoted by
$c_1,\cdots,c_H$, which constitute a solution path for sequential decision making.
At each decision point, SMP-modification candidates are generated using GP (Genetic Programming) and evaluated to see whether they accelerate reward production. GP evolves slowly and is not aggressive for identifying useful SMP-modifications, making it stable and ideal for lifelong RL.
SSC (Success-Story Criterion) is
\begin{eqnarray}
\frac{R(t)}{t}<\frac{R(t)-R(c_1)}{t-c_1}<\frac{R(t)-R(c_2)}{t-c_2}<\cdots<\frac{R(t)-R(c_{|V_t|})}{t-c_{|V_t|}},\nonumber
\end{eqnarray} 
where $V_t$ is the set of all checkpoints up to time $t$. Ratio of reward and time is a measure of RL algorithm efficiency.
Useful SMP-modifications meet SSC and are kept in RL algorithm. 
As in \cite{Schmidhuber2003}, G$\ddot{\text{o}}$del machines based upon useful incremental self-improvement theoretically achieve global optimum and is never trapped in any local optimum. 
SSA with SMP is a computationally stable G$\ddot{\text{o}}$del machine that is suitable for lifelong RL. 
\cite{Yao1993} points out that evolutionary algorithms can be faster than backpropagation in estimating neural network weights.

\subsubsection{MetaRL with GBML}

Gradient-based meta-learning models such as MAML and Reptile can be extended to meta-RL framework for efficient adaptation of policy over changing environments.
Actually GBML can be easily integrated into almost all deep learning settings. For modification of GBML to be in RL settings, the following routes may be pursued: (1) re-define loss function to be maximization of reward function in RL setting, (2) re-define task-specific parameters to be in deep policy model and meta-parameters are the initial values of task-specific parameters in GBML, (3) re-design algorithms to update task-specific parameters in base learner and algorithms to update meta-parameters in meta-learner etc.
An extension of MAML to RL tasks is presented in \cite{Finn2017}, where the loss function of task $\mathcal{T}_i$ is written as
\begin{eqnarray}
&\mathcal{L}_{
	\mathcal{T}_i}(f_\phi)=-\mathbb{E}_{
	\substack{S_i,A_i}
}
\left\{\sum_{t=1}^H \gamma^{t}r_i(s_t,a_t)\right\},
\nonumber
\end{eqnarray}
where $f_\phi$ is a deep model specified for policy, and $\phi$ is the task-specific policy parameter of interest. In base learner, task training data is plugged into this loss function and minimization of loss function leads to update on task-specific parameters. In meta-learner, task validation data is plugged into this loss function and minimization of loss function leads to update on meta-parameters.
This is the direct application of MAML to few-shot RL tasks.
For RL tasks, we minimize the loss function $\mathcal{L}_{\mathcal{T}_i}(f_\phi)$ to estimate an optimal policy $f_\phi$ and seek fast adaptation to novel tasks with changing environment. Here GBML is applied to solve few-shot RL tasks, as a result, unseen few-shot RL tasks should be similar to trained tasks and have the same data structure as trained tasks. 

Another application of MAML in meta-RL is E-MAML \cite{Stadie2018a} which adds an exploratory term to MAML. E-MAML considers the influence of sampling $\pi_\theta$ upon future rewards $R(\tau)$, where $\tau\sim\pi_{U(\theta)}$.  
MAML applies stochastic gradient of 
$\int R(\tau)\pi_{U(\theta)}(\tau)d\tau$.
E-MAML uses stochastic gradient of 
$\int\int R(\tau)\pi_{U(\theta)}(\tau)\pi_\theta(\bar{\tau})d\bar{\tau}d\tau$, where $\bar{\tau}\sim\pi_\theta$.

\subsubsection{MetaRL with Q-Learning}

Meta-RL in CNN architecture optimization is surveyed in \cite{Jaafra2018}. 
Reinforcement learning can be applied to solve complex optimization problem which includes optimization of both discrete and continuous hyperparameters. In such situations, reward function is taken to be the prediction accuracy of candidate models along optimization process. Action option is taken to be the value of hyperparameter to be explored. Through deep RL model, an optimal policy is estimated such that model has the highest prediction accuracy with hyperparameter values chosen by the optimal policy. Deep RL models are not as efficient to train so that meta-RL models are applied to accelerate the training of deep RL models under slightly different settings using only few-shot RL task data. 
Evolutionary algorithms provide candidates for hyperparameter search. We may
utilize a goal-based Q-learning model to update parameters in adaptation.
Reward function is denoted as
\begin{eqnarray}
V^\pi (s)=\mathbb{E}(r|s,\pi).
\nonumber
\end{eqnarray}
Bellman update on Q function is written as
\begin{eqnarray}
Q(s_t,a_t)\leftarrow Q(s_t,a_t)+\alpha [r_{t+1}+\gamma \max_a Q(s_{t+1},a)-Q(s_t,a_t)].
\nonumber
\end{eqnarray}
The optimal action that maximizes Q-function is the best CNN architecture design. Auto-Meta \cite{Kim2018} can use neural network architecture search algorithms in \cite{Jaafra2018} to optimize  meta-learner models specified as CNN or RNN.

\textbf{MQL} (Meta-Q-Learning) \cite{Fakoor2020} considers meta-RL using Q-learning embedded in context. Meta-parameter is estimated by maximizing average reward from training data and minimizing temporal-difference error from Bellman equation. Adaptation is achieved by propensity-score weighted sampling from previous training replay buffer. In openAI-gym \cite{Brockman2016} environments, MQL is efficient and demonstrates performance comparable to PEARL. 

\subsubsection{MetaRL with Actor-Critic}

\textbf{PEARL} (Probabilistic Embeddings of Actor-critic RL) 
\cite{Rakelly2019} defines loss function with actor-critic RL and applies MAML for few-shot update on parameters in adaptation. Actor-critic RL contains policy estimation (actor) and value evaluation (critic).
First, variational inference is applied to infer hidden state $z$.
Inference network $q_\phi(z|c)$ is specified as
\begin{eqnarray}
&q_\phi(z|c_1,\cdots,c_N)\propto \prod_{n=1}^N
\Psi_\phi (z|c_n),
\nonumber
\end{eqnarray}
where context $c_n=(s_n,a_n,s_n^\prime,r_n)$ consists of state-action pairs, $\Psi_\phi (z|c_n)=\mathcal{N}(f^{\mu_n}_\phi(c_n),f^{\sigma_n}_\phi(c_n))$ is a Gaussian distribution and $f_\phi$ is a neural network that predicts $\mu_n$ and $\sigma_n$ for each context $c_n$.
Variational lower bound optimized to estimate $z$ is
\begin{eqnarray}
\mathbb{E}_{\mathcal{T}}\mathbb{E}_{z} [
R(\mathcal{T},z)+\beta D_{KL}\{q_\phi (z|c)\Vert p(z)\}],
\nonumber
\end{eqnarray}
where $\mathcal{T}\sim p(\mathcal{T})$, $z\sim q_\phi(z|c)$, $p(z)$ is a prior distribution over $z$ and $R(\mathcal{T},z)$ is a task-specific objective.
This meta-RL algorithm is based upon soft actor-critic (SAC) method where we consider the 
joint optimization of inference network $q_\phi(z|c)$, actor $\pi_\theta(a|s,z)$ and critic
$Q_\xi(s,a,z)$.
Critic loss function is defined to be
\begin{eqnarray}
\mathcal{L}_{critic}=\mathbb{E}_{	\substack{
		z\sim q_\phi(z|c)\\
		(s,a,r,s^\prime)\sim\mathcal{B} }
} [Q_\xi (s,a,z)-(r+\bar{V}(s^\prime,\bar{z}))]^2,
\nonumber
\end{eqnarray}
where $Q_\xi (s,a,z)$ is the value of current state-action pair, $\bar{z}$ is the average condition, $r$ is current reward and $\bar{V}(s^\prime,\bar{z})$ is the optimal target value derived from all possible future states. Critic loss minimizes the value difference between current state-action and optimal target value.
Actor loss function \cite{Haarnoja2018} is defined as
\begin{eqnarray}
&\mathcal{L}_{actor}=\mathbb{E}_{
	\substack{a\sim\pi_\theta\\
		s\sim\mathcal{B}
	}
}
D_{KL}[\pi_\theta (a|s,\bar{z})\Vert \frac{\exp(Q_\xi(s,a,\bar{z}))}{Z_\theta (s)}],
\nonumber
\end{eqnarray}
where $Z_\theta (s)=\int \exp(Q_\xi(s,a,\bar{z})) da$  is a normalizing constant. Actor loss function minimizes the distance between policy $\pi_\theta (a|s,\bar{z})$ and value-directed action weight specified as
$\exp(Q_\xi(s,a,\bar{z}))/Z_\theta (s)$ which assigns more weight to action with higher value.

Joint estimation of parameters in inference network, policy and value function is conducted with
MAML. First,
sample RL task $b_i=\{s_j,a_j\}_{j=1}^N\sim \mathcal{B}^i$ and context $c^i\sim S_c(\mathcal{B}^i)$
where $c^i=\{s_j,a_j,s_j^\prime,r_j\}_{j=1}^N$.
Then 
sample hidden state $z\sim q_\phi(z|c^i)$.
Compute loss functions:
$\mathcal{L}^i_{actor}=\mathcal{L}_{actor}(b^i,z)$,
$\mathcal{L}^i_{critic}=\mathcal{L}_{critic}(b^i,z)$ and $\mathcal{L}_{KL}^i=\beta D_{KL}(q_\phi(z|c^i)\Vert r(z))$.
MAML updates on parameters are as follows:
\begin{eqnarray}
&\phi\leftarrow \phi-\alpha_1\bigtriangledown_\phi \sum_i (\mathcal{L}^i_{critic}+\mathcal{L}_{KL}^i),\nonumber\\
&\theta\leftarrow \theta-\alpha_2\bigtriangledown_\theta \sum_i
\mathcal{L}^i_{actor},\nonumber\\
&\xi\leftarrow\xi-\alpha_3\bigtriangledown_\xi\sum_i \mathcal{L}^i_{critic}.
\nonumber
\end{eqnarray}
Actor loss function is used to update policy parameter and critic loss function is used to update inference network and value function.

\subsubsection{MetaRL with DDPG}

\textbf{MetaGenRL} \cite{Kirsch2019} improves generalization to vastly different few-shot RL tasks through combining multiple agents to train a meta-learner. MetaGenRL considers DDPG (Deep Deterministic Policy Gradient) actor-critic model \cite{Silver2014,Lillicrap2016} which applies DQN (deep Q network) to continuous action domain. \cite{Kirsch2019} demonstrates that DDPG supports generalization to vastly different environments. 
MetaGenRL explicitly separates adaptation of policy $\pi_\theta$ and value $Q_\xi$ model which contributes to improvement in generalization capability.
Application of second-order gradients improves sampling efficiency in RL.
Critic loss is
\begin{eqnarray}
&\mathcal{L}_{critic}=\sum_{s,a,r,s^\prime}[Q_\xi(s,a)-(r+\gamma \bar{V} (s^\prime,\pi_\theta (s^\prime)))]^2.\nonumber
\end{eqnarray}
Critic loss is minimized to estimate value parameter $\xi$.
Policy parameter $\theta$ can be updated by taking the gradient of
$\mathbb{E}_\tau [\sum_{t=1}^H \log \pi_\theta (a|s) A(\tau,V,t)]$,
where $A(\tau,V,t)$ is a GAE (generalized advantage estimate) and $V:S\to\mathbb{R}$ is a value function. 
In MetaGenRL, meta-objective function is specified as a neural network $L_\alpha(\tau,\pi_\theta,V)$, where $\pi_\theta$ is an auxiliary input and $\alpha$ is neural network parameter. During adaptation, $\xi$ is updated by minimizing critic loss $\mathcal{L}_{critic}$,
\begin{eqnarray}
&\theta^\prime \leftarrow \theta-\bigtriangledown_\theta L_\alpha(\tau,x(\theta),V),\nonumber\\
&\alpha\leftarrow \alpha+\bigtriangledown_\alpha Q_\xi(s,\pi_{\theta^\prime}(s)).\nonumber
\end{eqnarray}

\subsubsection{Continual MetaRL with Dynamic Environments}

Robots learn to survive dynamic environments through continual meta-RL method \textbf{GrBAL} (Gradient-Based Adaptive Learner) \cite{Nagabandi2019}.
Transition probability estimator is $\hat{p}_\theta (s^\prime|s,a)$ where
$\hat{\theta}$ is a maximum likelihood estimate.
In meta-learning procedure, a general equation of task-specific parameter update is
\begin{eqnarray}
\phi_{\mathcal{T}}=u_\psi (\mathcal{D}_{\mathcal{T}}^{tr},\theta),
\nonumber
\end{eqnarray}
where $\theta$ is meta-parameter, and
$u_\psi (\mathcal{D}_{\mathcal{T}}^{tr},\theta)
=\theta-\psi\bigtriangledown_\theta\mathcal{L}(\mathcal{D}^{tr}_{\mathcal{T}},\theta)$ in MAML.
Parameters are estimated by minimizing the meta-loss function:
\begin{eqnarray}
\min_{\theta,\psi} \mathbb{E}_{\mathcal{T}\sim \rho(\mathcal{T})}
[\mathcal{L}(\mathcal{D}^{val}_{\mathcal{T}},\phi_{\mathcal{T}})].
\nonumber
\end{eqnarray}
In total, $M$ trajectories are sampled from a novel task.
We use the experience of $K$ time steps as training data to predict next $K$ time steps which is testing data.
Let $\mathcal{E}$ denote the set of all dynamic real-world environments.
Under these concrete settings, meta-objective is
\begin{eqnarray}
\min_{\theta,\psi}\mathbb{E}_{\tau_{\mathcal{E}}(t-M,t+K)\sim\mathcal{D}}[\mathcal{L}(\tau_{\mathcal{E}}(t,t+K),\theta^\prime_{\mathcal{E}})],
\nonumber
\end{eqnarray}
where $\tau_{\mathcal{E}}(t-M,t+K)$ is the task trajectories data between timestamps $t-M$ and $t+K$, $\tau_{\mathcal{E}}(t,t+K)$ is the task validation trajectories data between $t$ and $t+K$ and $\tau_{\mathcal{E}}(t-M,t-1)$ is the task training trajectories data. Task validation data between $t$ and $t+K$ is taken to update meta-parameters: initial values of task-specific parameters and base learner hyperparameters. Task training data between $t-M$ and $t-1$ is taken to update task-specific parameters. 
Update on task-specific parameters in base learner is written as
\begin{eqnarray}
&\phi_{\mathcal{E}}=u_\psi (\tau_{\mathcal{E}}(t-M,t-1),\theta).
\nonumber
\end{eqnarray}
Here base learner update on task-specific parameter using task training can be written as
GrBAL:
\begin{eqnarray}
\phi_{\mathcal{E}}=\theta_{\mathcal{E}}+\psi\bigtriangledown_{\theta_{\mathcal{E}}} \frac{1}{M}\sum_{k=t-M}^{t-1} \log \hat{p}_{\theta_{\mathcal{E}}}(s_{k+1}|s_k,a_k),
\nonumber
\end{eqnarray}
where $\psi$ is the step size.
Recurrence-Based Adaptive Learner (\textbf{ReBAL})
update parameters $\psi$ using a recurrent model.
Meta-loss function using task validation data is 
\begin{eqnarray}
\mathcal{L}(\tau_{\mathcal{E}}(t,t+K),\phi_{\mathcal{E}})\equiv -\frac{1}{K}\sum_{k=t}^{t+K}
\log \hat{p}_{\phi_{\mathcal{E}}}(s_{k+1}|s_k,a_k).
\nonumber
\end{eqnarray}

Another meta-RL framework based upon continual self-improvement is proposed in \cite{Schmidhuber1990}. \cite{Schmidhuber1990}
performs supervised learning augmented by artificial curiosity \cite{Alet2020} and boredom. 
Meta-loss minimization creates dynamic attention which guides algorithm exploration in 
artificial curiosity. 
Boredom refers to avoidance of environments which have been perfectly solved.
Layered meta-learning in section \ref{Layers-Meta-Learning}
can be applied in meta-RL tasks as well. 
Internal feedback corresponding to base learner comes from stepwise interaction between agents and known environments under the influence of agent actions. 
External feedback corresponding to meta-learner is from all interactions between agents and unknown environments, which shapes the whole embedded credit assignment mechanism.
Evaluation of internal and external feedback leads to update on base learner and meta-learners for adaptation to unseen tasks.

\subsubsection{Meta-RL with Coevolution}

Coevolution \cite{Schmidhuber1999} involves cooperation and competition between agents and environments. 
In \textbf{POET} (Paired Open-Ended Trailblazer) \cite{Wang2019a},
more complex environment is generated from coevolution between agent and environment, in which POET trains an accurate RL model for a difficult environment which cannot be solved otherwise by training from scratch. 
In the $i$th sampled trajectory, reward from interaction between agent and environment is $R(\theta_i)$ parametrized by $\theta_i\sim p_{\theta,\sigma} (\theta_i)=N(\theta,\sigma^2)$. Optimization to estimate hyperparameter $\theta$ is
\begin{eqnarray}
\max_{\theta,\sigma} \mathbb{E}_{w\sim p_{\theta,\sigma} (w)} R(w) \approx \frac{1}{n}\sum_{i=1}^n R(\theta_i).
\end{eqnarray}
To simplify, variance in sampled trajectories $\sigma$ can be set to be $1$.
POET consists of three components: generation of more complex environment $R(\theta)$ from coevolution between agent $\theta$ and environment $R$, train an RL model for current agent and environment pair, and adaptation of current agent to varying environment.
POET belongs to \textbf{AI-GA} (AI-generating algorithm) \cite{Clune2019} for self-supervised learning \cite{Schmidhuber1990}. In general, AI-GA consists of three components: meta-learning architecture, learner algorithms and generation of more complex environment.

\subsubsection{MetaRL with Task Decomposition}

Task decomposition \cite{Whiteson2005} decomposes complex tasks into a hierarchy of easier subtasks under a pre-specified or learned dependence structure. 
Meta-learning with task decomposition makes  adaptation in complex tasks more accurate and more efficient. 
Dependencies rely upon the sequence of subtasks following a natural logical order, where the next subtask can only be performed after the previous one is finished.
The execution of a subtask may also depend upon the knowledge acquired from several other subtasks. 
Task decomposition follows either an observable logical hierarchy for the subtasks or a dependency structure estimated from unsupervised learning. 
Plan Parsing \cite{Barrett1994} may be performed in a top-down or bottom-up fashion. 
An optimal task decomposition plan may be learned progressively using genetic programming optimization, choosing an optimal action at every step. 
LfD (Learning from demonstration) may apply TACO (Task Decomposition via Temporal Alignment for Control) \cite{Shiarlis2018} to decompose complex tasks so that robots can learn subtasks more easily from demonstrations.
Among different subtasks with few-shot data, in order to estimate parameters more accurately, parameter sharing may be assumed for subtasks and data from these subtasks can be aggregated to estimate shared parameter.
Many-layered learning \cite{Utgoff2002} proposes STL (Stream-To-Layers) which learns a hierarchical structure from online streaming data based upon knowledge dependency.

\textbf{MSGI} (Meta-Learner with Subtask Graph Inference) \cite{Sohn2020} models dependency structure in subtasks using the subtask graph $G$ with subtask precondition $G_c$ and subtask reward $G_r$.
Transition model with subtask graph is factorized into
\begin{eqnarray}
p(s^\prime|s,a)=\prod_i p_{G^i_c}(s_i^\prime|s,a).\nonumber
\end{eqnarray} 
Reward with subtask graph is the summation of reward from all subtasks
\begin{eqnarray}
R(s,a)=\sum_i R_{G^i_r}(s,a).\nonumber
\end{eqnarray}
Subtask precondition is estimated using decision tree where critical information required for the subtask is found via dimension reduction and model selection. For an unseen task, subtask graph is learned before adaptation is made with MAML.

\subsection{Meta-Imitation Learning}
\label{meta-imitation}

\begin{figure}[htpb]
	\centering
	\includegraphics[width=0.8in]{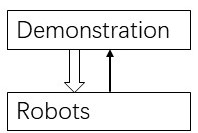}
	\caption{Meta-imitation learning. Find a policy such that robots act like demonstration. Interaction between robots and environment is the same as in demonstration. Robots clone policy and reward in demonstration.}
	\label{meta-imitate}
\end{figure}
In complex unstructured environments where there is no explicit task or reward definition, we can learn through imitation learning. 
With only one or few demonstrations, meta-learning is introduced to achieve fast adaptation of imitation model to changing environment or varying target policy, as in figure \ref{meta-imitate}. Similar here, meta-learning framework is applied to solve few-shot imitation task with deep imitation learning model.
Meta-imitation learning applies similar past experiences to a novel task in sparse-reward situations.
Applications of meta-imitation learning concentrate upon robotics where robots acquire skills from one visual demonstration.
\cite{Finn2017a} proposes a one-shot visual imitation learning model built upon MAML.
Few-shot meta-imitation learning is realized through MAML updates on parameters in novel settings.
Demonstration trajectory data is denoted as
$\tau=\{o_1,a_1,\cdots,o_H,a_H\}$ which consists of
observation-action pairs.
Similarly, an imitation task is defined as
\begin{eqnarray}
\mathcal{T}_i=[\tau\sim \pi^*_i,\mathcal{L}(a_1,\cdots,a_H,\hat{a}_1,\cdots,\hat{a}_H),H],
\nonumber
\end{eqnarray}
where $H$ is the finite time horizon, $\pi_i^*$ is an expert policy we target to imitate, and
$\mathcal{L}$ is an imitation loss function.
Loss function is mean squared error if actions are continuous, and cross-entropy otherwise.
Generally imitation loss function is specified as 
behavior-cloning loss:
\begin{eqnarray}
&\mathcal{L}_{\mathcal{T}_i}(f_\phi)=
\sum_{\tau^{(j)}\sim\mathcal{T}_i}\sum_{t=1}^H
\Vert f_\phi(o_t^{(j)})-a_t^{(j)}\Vert_2^2,
\nonumber
\end{eqnarray}
where $\tau^{(j)}$ is a trajectory from RL task $\mathcal{T}_i$, and $f_\phi$ is a CNN policy mapping from observation to action.
Policy CNN is parameterized by weights $W$ and bias $b$ of last layer, that is $\phi=(W,b)$.
Then behavior-cloning loss function is rewritten as
\begin{eqnarray}
&\mathcal{L}_{\mathcal{T}_i}^*(f_\phi)=\sum_{\tau^{(j)}\sim\mathcal{T}_i}\sum_t
\Vert Wy_t^{(j)}+b-a_t^{(j)}\Vert_2^2.
\nonumber
\end{eqnarray}
Here in \cite{Finn2017a}, policy CNN $f_\phi$ is trained on one demonstration from target policy $\pi_i^*$ and tested on another demonstration from the same target policy $\pi_i^*$ to compute testing error.
Meta-objective with MAML update on parameters is written as
\begin{eqnarray}
\min_{\theta,W,b}\sum_{\mathcal{T}_i\sim p(\mathcal{T})}\mathcal{L}^*_{\mathcal{T}_i}
(f_{\phi_i}),
\text{where}~
\mathcal{L}^*_{\mathcal{T}_i}
(f_{\phi_i})=\mathcal{L}^*_{\mathcal{T}_i}
(f_{\theta-\alpha\bigtriangledown_\theta \mathcal{L}^*_{\mathcal{T}_i}(f_\theta)}).
\nonumber
\end{eqnarray}
After MAML update on parameters, model imitates one demonstration from a new expert policy. Additionally,
in cases without expert action to imitate,
imitation loss function is written as
\begin{eqnarray}
\mathcal{L}^*_{\mathcal{T}_i}(f_\phi)=
\sum_{\tau^{(j)}\sim\mathcal{T}_i}\sum_t
\Vert Wy_t^{(j)}+b\Vert_2^2.
\nonumber
\end{eqnarray}
Policy CNN $f_\phi$ is directly estimated from one  demonstration video.

A similar meta-imitation model for the case where robots learn from both human and robot demonstrations is proposed in \cite{Yu2018}. 
Human demonstration data is denoted as $\mathcal{D}_{\mathcal{T}_i}^h$.
Robot demonstration data is
$\mathcal{D}_{\mathcal{T}_i}^r$.
Human video $d^h\sim \mathcal{D}_{\mathcal{T}_i}^h$.
Robot video $d^r=\{o_1,\cdots,o_H,s_1,\cdots,s_H,a_1,\cdots,a_H\}\sim \mathcal{D}_{\mathcal{T}_i}^r$.
MAML update on task-specific policy parameter is
\begin{eqnarray}
\phi_{\mathcal{T}}=\theta-\alpha\bigtriangledown_\theta \mathcal{L}_\psi (\theta,d^h),
\nonumber
\end{eqnarray}
where $\mathcal{L}_\psi$ parameterized by $\psi$ is the adaptation objective.
Behaviorial cloning loss function is
\begin{eqnarray}
&\mathcal{L}_{BC}(\phi,d^r)
=\sum_t\log \pi_\phi (a_t|o_t,s_t).
\nonumber
\end{eqnarray}
Meta-objective is the combination of MAML policy parameter adaptation and behavioral cloning loss function:
\begin{eqnarray}
\min_{\theta,\psi}\sum_{\mathcal{T}\sim p(\mathcal{T})}\sum_{d^h\in\mathcal{D}^h_{\mathcal{T}}}\sum_{d^r\in\mathcal{D}^r_{\mathcal{T}}}
\mathcal{L}_{BC}(\phi_{\mathcal{T}},d^r).
\nonumber
\end{eqnarray}
Then MAML update on meta-parameter is
\begin{eqnarray}
(\theta,\psi)\leftarrow (\theta,\psi)-
\beta\bigtriangledown_{\theta,\psi}\mathcal{L}_{BC}(\phi_{\mathcal{T}},d^r).
\nonumber
\end{eqnarray}
From the probabilistic perspective,
$\phi_{\mathcal{T}}$ is an MAP estimate of $\phi$ based upon approximate inference on $\log p(\phi|\mathcal{D}^{tr}_{\mathcal{T}},\theta)$ \cite{Grant2018}.

High-fidelity imitation model
\textbf{MetaMimic} proposed in \cite{Paine2018} stores memory of off-policy RL and 
replays memory for imitation learning. Policy model is specified to be a
wide deep neural network with batch normalization.
Usually larger deeper neural network policy model in off-policy RL brings better adaptation to novel tasks.

Experience replay memory is denoted as $\mathcal{M}$. Memory items are constructed in the following procedure. 
Action $a_t\sim\pi(o_t,d_{t+1})$, where $d_{t+1}$ is the demonstration video that leads the imitation.
Execute $a_t$. Then observe $o_{t+1}$ and task reward $r_t^{task}$.
Compute imitation reward $r_t^{imitate}=g(o_{t+1},d_{t+1})$, which measures the similarity between learned observation and video demonstration.
The following item $(o_t,a_t,o_{t+1},r_t^{task},r_t^{imitate},d_{t+1})$ is stored in memory $\mathcal{M}$.

One demonstration is denoted as $d=\{d_1,\cdots,d_H\}\sim p(d)$.
Imitation policy is specified as $\pi_\theta$ parameterized by $\theta$.
Action $a_t=\pi_\theta(o_t,d)$.
Environment renderer $\mathcal{E}(\pi)$ maps action to observation
$\mathcal{E}(\pi)=\{o_1,\cdots,o_H\}$.
Policy parameter $\theta$ is estimated by maximizing imitation reward:
\begin{eqnarray}
\max_\theta \mathbb{E}_{d\sim p(d)}\left[\sum_{t=1}^H \gamma^t g\left[d_t,\mathcal{E}(\pi_\theta(o_t,d))_t\right]\right],
\nonumber
\end{eqnarray}
where $\gamma>0$ is the discount factor, and $g$ is a similarity measure. By maximizing imitation reward, similarity between 
learned observation and demonstration is maximized.
Imitation reward is the Euclidean distance between learned observation and demonstration
\begin{eqnarray}
r_t^{imitate}
=\beta_1\exp(-\Vert o_{t+1}^{image}-d_{t+1}^{image}\Vert_2^2)
+\beta_2\exp(-\Vert o_{t+1}^{body}-d_{t+1}^{body}\Vert_2^2),
\nonumber
\end{eqnarray}
where $o_{t}^{body}$ is the body position coordinate and velocity, and $o_{t}^{image}$ is the image pixel capturing interactions between body and objects.

\textbf{WTL} (\textbf{Watch-Try-Learn}) \cite{Zhou2019a} is a meta-imitation learning (MIL) model which combines MIL with trial-and-error RL.
It includes a trial procedure which generates a policy from one demonstration and a retrial process that compares trials with demonstration to identify a successful policy.
Identifiability issue may arise in which policy is not uniquely determined by one demonstration.

A trajectory from RL task $\mathcal{T}_i$ is denoted as $\tau_i=[(s_1,a_1,r_i(s_1,a_1)),\cdots,(s_H,a_H,r_i(s_H,a_H))]$.
Demonstration trajectory is denoted as
$d=[(s_1,a_1),\cdots,(s_H,a_H)]$.
Training demonstrations $\{d_{i,k}\}$ are sampled from $\mathcal{D}_i$.
Trial policy $\pi_\theta(a|s,\{d_{i,k}\})$ is proposed based upon one demonstration.
Trial objective to estimate trial policy $\pi_\theta$ is 
\begin{eqnarray}
\mathcal{L}(\theta,\mathcal{D}_i)=
\mathbb{E}_{\{d_{i,k}\}\sim\mathcal{D}_i}
\mathbb{E}_{d^{test}_i\sim \mathcal{D}_i\backslash \{d_{i,k}\}}
\mathbb{E}_{(s_t,a_t)\sim d^{test}_i}
[-\log \pi_{\theta} (a_t|s_t,\{d_{i,k}\})].
\nonumber
\end{eqnarray}
MAML is applied to update $\theta$ with trial objective:
\begin{eqnarray}
\theta\leftarrow\theta-\alpha\bigtriangledown_\theta\mathcal{L}(\theta,\mathcal{D}_i).
\nonumber
\end{eqnarray}
Trial trajectories are generated from trial policy
$\{\tau_{i,l}\}\sim \pi_\theta(a|s,\{d_{i,k}\})$.
Demonstration-trial trajectory pairs $(\{d_{i,k}\},\{\tau_{i,l}\})$ are sampled from
$\mathcal{D}^*_i$.
A retrial policy integrates generated trials with demonstration to learn a successful policy
$\pi_\phi(a|s,\{d_{i,k}\},\{\tau_{i,l}\})$.
Retrial objective to estimate a successful policy
$\pi_\phi$ is
\begin{eqnarray}
\mathcal{L}(\phi,\mathcal{D}_i,\mathcal{D}^*_i)=\mathbb{E}_{(\{d_{i,k}\},\{\tau_{i,l}\})\sim\mathcal{D}^*_i}\mathbb{E}_{d^{test}_i\sim\mathcal{D}_i\backslash\{d_{i,k}\}}
\mathbb{E}_{(s_t,a_t)\sim d^{test}_i}[-\log\pi_\phi(a_t|s_t,\{d_{i,k}\},\{\tau_{i,l}\})].
\nonumber
\end{eqnarray}
MAML is applied to update $\phi$ with retrial objective:
\begin{eqnarray}
\phi\leftarrow\phi-\beta\bigtriangledown_\phi \mathcal{L}(\phi,\mathcal{D}_i,\mathcal{D}^*_i).
\nonumber
\end{eqnarray}

\subsection{Online Meta-learning}
\label{online-meta-learning}

\begin{figure}[htpb]
	\centering
	\includegraphics[width=2in]{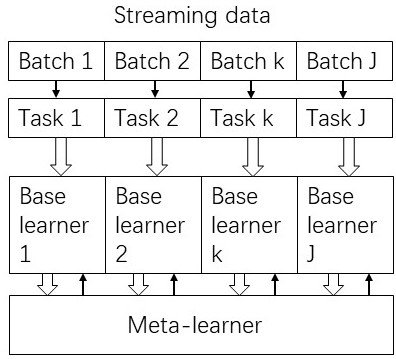}
	\caption{Online meta-learning. Each data batch in streaming data is regarded as a task dataset. Base learner is applied to each task and meta-learner updates base learner across tasks.}
	\label{online}
\end{figure}

Online learning is applied to streaming data where  a small batch of data is obtained each time and fast adaptation is required for timely reaction to continually changing environment. 
In online meta-learning, a small batch of streaming data obtained each time is regarded as a task dataset, as in figure \ref{online}.
In \cite{Finn2019}, \textbf{FTML}
(Follow The Meta Leader) is constructed by 
integrating a classical online algorithm follow the leader (FTL) with MAML.
FTL is proved to guarantee strong performance for smooth convex loss functions.

For task $\mathcal{T}_i$, MAML updates parameter $\theta$ using
\begin{eqnarray}
U_i(\theta)=\theta-\alpha\bigtriangledown_\theta \mathcal{L}_{\mathcal{T}_i}(h_\theta).
\nonumber
\end{eqnarray}
Regret function from FTL measures the distance between current value and optimal target value
\begin{eqnarray}
R(\theta)=\sum_{i=1}^J
\mathcal{L}_{\mathcal{T}_i}(h_{U_i(\theta)})
-\min_\theta\sum_{i=1}^J \mathcal{L}_{\mathcal{T}_i}(h_{U_i(\theta)}).
\label{regret}
\end{eqnarray}
By estimating optimal task-specific parameter that minimizes regret function, current value is closest to optimum. 
Since optimum is a fixed target, minimization of regret function is equivalent to minimizing first component in equation (\ref{regret}).
Tasks are streaming data and follow a natural time order so that
FTML updates task-specific parameters by minimizing the summation of task-loss up to task $\mathcal{T}_i$:
\begin{eqnarray}
\phi_{i+1}=\text{argmin}_\theta \sum_{k=1}^i
\mathcal{L}_{\mathcal{T}_k}(h_{U_k(\theta)}).
\nonumber
\end{eqnarray}
In order to solve this optimization online, we
apply practical online gradient computation:
\begin{eqnarray}
d_i(\theta)=\bigtriangledown_\theta \mathbb{E}_{k\sim v^t} \mathcal{L}(\mathcal{D}^{val}_k,U_k(\theta)),\text{where}~U_k(\theta)\equiv\theta-\alpha\bigtriangledown_\theta\mathcal{L}(\mathcal{D}^{tr}_k,\theta),
\nonumber
\end{eqnarray}
and $v^t$ is a sampling distribution from tasks $\mathcal{T}_1,\cdots,\mathcal{T}_i$.

In robotics, robots need continually rapid and precise adaptation to online streaming data especially when simulated mechanical failures or extreme environmental conditions suddenly occur. In such out-of-distribution tasks, fast adaptation of trained problem-solving model is required. 
In \cite{Nagabandi2019a},
deep online meta-learning and nonparametric Bayes are integrated to provide a feasible solution for robots to survive. In meta-learning, we concern adaptation of deep learning models to unseen few-shot in-distribution or out-of-distribution tasks. Task-specific parameters reflect task-specific features of each unseen task. Meta-parameters reflect features shared by multiple tasks. We can see that the whole meta-learning system is around the task distribution. This online meta-learning utilizes nonparametric Bayesian theories to compute posterior task distribution.
This method regards tasks to be countably infinitely many and assumes task distribution to be one of nonparametric Bayesian distribution types. After obtaining task posterior distribution, we can judge whether each unseen task is in-distribution or out-of-distribution and obtain posterior distributions of task-specific parameters and meta-parameters. 

%

Task distribution is $P(\mathcal{T}_t=\mathcal{T}_i|\pmb{x}_t,y_t
)$, where the set of all tasks is a countable infinite set.
Expectation-maximization (EM) is applied to
maximize log likelihood:
\begin{eqnarray}
\mathcal{L}=\mathbb{E}_{\mathcal{T}_i\sim p(\mathcal{T}_t|\pmb{x}_t,y_t)}
[\log p_{\phi_t(\mathcal{T}_i)}(y_t|\pmb{x}_t,\mathcal{T}_i)],
\nonumber
\end{eqnarray}
where $\phi_t(\mathcal{T}_i)$ is the task-specific parameter. 
Posterior distribution of task is
\begin{eqnarray}
p(\mathcal{T}_t=\mathcal{T}_i|\pmb{x}_t,y_t)\propto
p_{\phi_t(\mathcal{T}_i)}(y_t|\pmb{x}_t,\mathcal{T}_t=\mathcal{T}_i)P(\mathcal{T}_t=\mathcal{T}_i).
\nonumber
\end{eqnarray}
Nonparameteric Bayesian prior
Chinese restaurant process (CRP) is used as the prior distribution on task:
\begin{eqnarray}
&P(\mathcal{T}_t=\mathcal{T}_i)=
\frac{\sum_{k=1}^{t-1}P(\mathcal{T}_k=\mathcal{T}_i)}{t-1+\alpha},\nonumber\\
&P(\mathcal{T}_t=\mathcal{T}_{new})=\frac{\alpha}{t-1+\alpha},
\nonumber
\end{eqnarray}
where $\alpha>0$.
Task prior distribution CRP is
\begin{eqnarray}
P(\mathcal{T}_t=\mathcal{T}_i)=\left\{\sum_{k=1}^{t-1}P(\mathcal{T}_k=\mathcal{T}_i)+\delta(\mathcal{T}_k=\mathcal{T}_{new})\alpha\right\}.\nonumber
\nonumber
\end{eqnarray}
Meta-parameter $\theta$ is the prior of task-specific parameter $\phi_t$.
One gradient update of $\phi_t$ from prior $\theta$ is built upon FTML in \cite{Finn2019} where loss function depends upon all trained tasks before $t$
\begin{eqnarray}
\phi_{t+1}(\mathcal{T}_i)=\theta-\beta\sum_{k=0}^t [P(\mathcal{T}_k=\mathcal{T}_i|\pmb{x}_k,y_k)
\cdot\bigtriangledown_{\phi_k(\mathcal{T}_i)}
\log p_{\phi_k(\mathcal{T}_i)}(y_k|\pmb{x}_k)].\nonumber
\end{eqnarray}
MAML update on task-specific parameter $\phi_{t}(\mathcal{T}_i)$ is 
\begin{eqnarray}
\phi_{t+1}(\mathcal{T}_i)=\phi_{t}(\mathcal{T}_i)-\beta P(\mathcal{T}_t=\mathcal{T}_i|\pmb{x}_t,y_t)
\cdot \bigtriangledown_{\phi_t(\mathcal{T}_i)}
\log p_{\phi_t(\mathcal{T}_i)}(y_t|\pmb{x}_t).\nonumber
\end{eqnarray}
MAP estimator from posterior task distribution is
\begin{eqnarray}
\mathcal{T}^*=\text{argmax}_{\mathcal{T}_i}
p_{\phi_{t+1}(\mathcal{T}_i)}(y_t|\pmb{x}_t,\mathcal{T}_t=\mathcal{T}_i).
\nonumber
\end{eqnarray}
Prediction is made based upon task-specific parameter $\phi_{t+1}(\mathcal{T}^*)$.
Estimation of task-specific parameters and meta-parameters depends upon the MAP estimator from posterior task distribution. MAP estimator of task is a task representative of the whole task distribution. Posterior distribution of task-specific parameters and meta-parametes can be derived from posterior task distribution. 

Another application of meta-online learning is fast online adaptation of neural network model.
\cite{Harrison2018} proposes \textbf{ALPaCA} (Adaptive Learning for Probabilistic Connectionist Architectures).
The last layer of a neural network is specified as Bayesian linear regression which we need to update in fast adaptation.
Posterior distribution of output label from last layer is
\begin{eqnarray}
p(y|\pmb{x},\mathcal{D}^*_t)=\int p(y|\pmb{x},\theta)p(\theta|\mathcal{D}^*_t)
d\theta,
\nonumber
\end{eqnarray}
where  $\mathcal{D}^*_t=\{(\pmb{x}_i,y_i)\}_{i=1}^t$ is training data including streaming data before $t$.
A surrogate function $q_\xi (y|\pmb{x}_{t+1},\mathcal{D}^*_t)$ is an approximation to the posterior distribution of output label.
To estimate surrogate function $\xi$, we
minimize KL distance
\begin{eqnarray}
\min_{\xi}D_{KL}(p(y|\pmb{x}_{t+1},\mathcal{D}^*_t)\Vert q_\xi (y|\pmb{x}_{t+1},\mathcal{D}^*_t)).
\nonumber
\end{eqnarray}
Bayesian linear regression at last layer of neural network is
$y_t^T=\phi^T(\pmb{x}_t;w)K+\epsilon_t$,
where 
$\epsilon_t\sim\mathcal{N}(0,\Sigma_\epsilon)$
and $\Sigma_\epsilon$ is assumed known.
Rewrite regression in matrix form:
\begin{eqnarray}
Y=\Phi K+E,
\nonumber
\end{eqnarray}
where $K$ is an unknown parameter and $\Lambda$ is the precision matrix of pre-trained $\Phi$.
Online adaptation of parameters in Bayesian linear regression can be explicitly expressed.
Online adaptation of parameters in Bayesian linear regression is through the
minimization of loss function:
\begin{eqnarray}
&&\mathcal{L}(\bar{K}_0,\Lambda_0,w,\theta^*)
\nonumber\\
&=&-\mathbb{E}_{\pmb{x},y,\mathcal{D}^*\sim p(\pmb{x},y,\mathcal{D}^*|\theta^*)}[\log q_\xi (y|\pmb{x},\mathcal{D}^*)].
\nonumber
\end{eqnarray}
Online adaptation of parameters in Bayesian linear regression are explicitly expressed as follows
\begin{eqnarray}
&\Lambda^{-1}_t\leftarrow \Lambda^{-1}_{t-1}
-\frac{(\Lambda^{-1}_{t-1}\phi_t)(\Lambda^{-1}_{t-1}\phi_t)^T}{1+\phi^T_t\Lambda^{-1}_{t-1}\phi_t},\nonumber\\
&Q_t\leftarrow \phi_ty_t^T+Q_{t-1},~
\bar{K}_t\leftarrow \Lambda^{-1}_tQ_t,~\bar{y}_{t+1}\leftarrow K_t^T\phi_{t+1},\nonumber\\
&\text{and}~\Sigma_{t+1}\leftarrow (1+\phi^T_{t+1}\Lambda^{-1}_t\phi_{t+1})\Sigma_\epsilon.
\nonumber
\end{eqnarray}

\subsection{Unsupervised Meta-learning}
\label{unsupervised-meta-learning}


Unsupervised meta-learning applies to cases without labelled data.
Unsupervised clustering can provide estimated labels for unlabelled data \cite{Hsu2019} then supervised meta-learning can be applied regarding estimated labels as true labels. On the other hand, unsupervised learning allows parameter adaptation in the inner loop to be independent of data labels \cite{Metz2019}.
\cite{Khodadadeh2018} proposes 
\textbf{UMTRA} (Unsupervised Meta-learning with Tasks constructed by Random Sampling and Augmentation).
Unlabeled data is denoted as $\mathcal{U}$. In UMTRA, training data is unlabelled and unsupervised learning method is applied.
In random labelling, $x_1,\cdots,x_N$ is randomly sampled from training data and is labeled as $(x_1,1),\cdots,(x_N,N)$, which implies that there are $N$ classes with one image per class. Every sampled image is itself a class. 
Validation data is constructed from applying data augmentation function to training data
$x_i^*=\mathcal{A}(x_i)$. Data augmentation is performed by generating another image in each class and generated image should be indistinguishable from the real image in that class. That is, in validation data, there are two images per class.
Then MAML is applied on this task data with $N$ image classes and $2$ images under each class. Few-shot image classification performance of UMTRA is not as good as MAML and Prototypical Network in supervised setting.

In MAML, update on task-specific parameter in the inner loop base learner is through supervised learning using labelled data. 
In \cite{Metz2019}, unsupervised learning is utilized in base learner to conduct parameter adaptation to novel tasks. In \cite{Metz2019}, meta-learner still uses supervised learning to update meta-parameters.
Base model is specified as a feedforward multilayer perceptron parameterized by $\phi_i$.
Adaptation of parameters in the inner loop uses unlabeled data and is formulated as
\begin{eqnarray}
\phi_{k+1}= 
d(\phi_k,\pmb{x}_i;\theta),
\nonumber
\end{eqnarray} 
where $d$ is the unsupervised update rule of task-specific parameter $\phi_i$, and $\theta$ is the meta-parameter, $\phi_{k+1}$ is the updated task-specific parameter after $k$ iterations. On each task, base learner updates task-specific parameter using unlabelled task data. 
In meta-learner, labelled data are divided into training data and validation data. 
For example in meta-learner training data, coefficients are estimated using ridge regression
\begin{eqnarray}
\hat{v}=\text{argmin}_v\left\{\Vert y-v^T\pmb{x}\Vert^2+\lambda \Vert v\Vert^2\right\}.
\nonumber
\end{eqnarray}
Meta-objective is defined to be the cosine similarity between observed labels and predicted labels on meta-learner validation data:
\begin{eqnarray}
\text{MetaObjective}(\theta)=\text{CosDist}(y,\hat{v}^T\pmb{x}).
\nonumber
\end{eqnarray}

Meta-parameter is estimated through the optimization of the meta-objective
\begin{eqnarray}
\theta^*=\text{argmin}_\theta \mathbb{E}_{task} \left\{\sum_i \text{MetaObjective}(\phi_i) \right\}.
\nonumber
\end{eqnarray}

For unlabeled data, another perspective is to first apply unsupervised clustering to estimate labels and then apply supervised meta-learning. 
In \textbf{CACTUs} (clustering to automatically construct tasks for unsupervised meta-learning) \cite{Hsu2019}, unsupervised learning is first applied to estimate data labels which map data to embedding space. 
Learned cluster $\mathcal{C}_k$ centroid is denoted as $c_k$.
Partition from unsupervised clustering and cluster centroids are jointly estimated using
\begin{eqnarray}
\mathcal{P},\{c_k\}=\text{argmin}_{\{\mathcal{C}_k\},\{c_k\}}\sum_{k=1}^n \sum_{z\in\mathcal{C}_k}\Vert z-c_k\Vert_A^2,
\nonumber
\nonumber
\end{eqnarray}
which minimizes the distance between embeddings and cluster centroids.
BiGAN, DeepCluster and Oracle are typical unsupervised clustering methods, which can be combined with MAML or ProtoNets to construct CACTUs.
Oracle+ProtoNets achieves $62.29\%$ accuracy on 5-way 5-shot miniImageNet, comparable to the performance of supervised meta-learning methods.
Performance of unsupervised deep clustering models directly affect prediction performance of CACTUs. With better unsupervised deep clustering methods used to estimate data labels, few-shot image classification prediction accuracy of CACTUs will be even higher.



\subsection{Meta-Learning for Machine Cognition}
\label{Meta-Learning-for-Machine-Cognition}

For multi-agent problems, interactions between agents include competition and cooperation, which can be modelled with game theory under information asymmetry or RL under limited resources.
Other homogenous agents may not be the same from the cognition of one agent. 
Different agents may view the same task to be different and reach vastly different solutions.
Meta-learning can be applied to generalize a model for this group of agents to another group. Meta-learning can also generalize the model for an agent in this group to another agent in the same group.

Machine theory of mind \cite{Rabinowitz2018} proposes \textbf{ToMnet} (Theory of Mind neural network) using meta-learning to understand cognition of others based upon self-cognition. Under meta-learning framework, others' cognition is a few adaptations from self-cognition. ToMnet is used to predict the future behavior of opponents based upon prior observations for decision making in multi-agent AI model. 
Each prediction in ToMnet combines a pre-trained neural network model upon individual agents and 
a neural network model upon the current mental status of this agent. 
Prediction from ToMnet is approximately optimal from the perspective of Bayesian online learning. 
As like all meta-learning algorithms such as base learner and meta-learner framework where base learner captures individual information and meta-learner learns shared information, ToMnet combines prior belief on all agents and online agent-specific information to make efficient adaptation.

\subsection{Meta-Learning for Statistical Theories}
\label{Meta-Learning-for-Statistical-Theories}

In R2D2 \cite{Bertinetto2019} and MetaOptNet \cite{Lee2019}, statistical learners such as SVM and logistic regression are used as base learner under meta-learning framework. They achieve high prediction accuracy in few-shot image classification. This is one positive aspect of integrating statistical learning and meta-learning. Since meta-learning focuses upon improving generalization capability of machine learning algorithms, it can also be applied to broaden the scope of statistical methods.
\cite{Wu2019} proposes \textbf{MetaVAE} (Meta Variational Autoencoders) which improves generalization of traditional statistical variational inference methods to new generative distributions. 
Let $x$ denote input training data, $z$ denote learned feature from input, $p_{\mathcal{D}_i}(x)$ denote marginal data distribution. Meta-distribution $p_{\mathcal{M}}$ is a distribution over all marginal data distributions $p_{\mathcal{D}_i}(x)$. 
Meta-inference model is a mapping from data distribution and input training data to learned feature vector $g_{\phi}:p_{\mathcal{M}}\times x\to z$.
The objective function is to optimize
\begin{eqnarray}
\max_\phi \mathbb{E}_{p_{\mathcal{D}_i}\sim p_{\mathcal{M}}} \left[\mathbb{E}_{p_{\mathcal{D}_i}(x)}\left[\mathbb{E}_{g_\phi(p_{\mathcal{D}_i},x)} \log \frac{p_{\theta_i}(x,z)}{g_\phi (p_{\mathcal{D}_i},x)(z)}\right]\right].\nonumber
\end{eqnarray}
MetaVAE is formulated as
\begin{eqnarray}
\max_\phi \mathbb{E}_{p_{\mathcal{D}_i}\sim p_{\mathcal{M}}}\left[\max_{\theta_i} \mathcal{L}_{\phi,\theta_i}(p_{\mathcal{D}_i})\right],\nonumber\\
\text{where}~\mathcal{L}_{\phi,\theta_i}(p_{\mathcal{D}_i})=-D_{KL}(p_{\mathcal{D}_i}(x)g_\phi(p_{\mathcal{D}_i},x)\Vert p(z)p_{\theta_i}(x|z)).
\nonumber
\end{eqnarray}
Inference designed this way is robust to a large family of data distributions $p_{\mathcal{M}}$. Since parametric data distribution is considered here, it is possible to construct sufficient statistics of data that is used to build learned feature vector.

\section{Conclusion and Future Research}
\label{Conclusion-and-Future-Research}

Our survey provides an overview upon meta-learning research including black-box adaptation, similarity-based approach, base learner and meta-learner model, Bayesian meta-learning, meta-RL, meta-imitation learning, online meta-learning and unsupervised meta-learning. There are more branches of meta-learning covering integration with causal reasoning, change-point estimation, Bayesian deep learning and real-life applications to explore in the future. 

Meta-learning concentrates upon adaptation to out-of-distribution tasks. For in-distribution tasks, similarity structure between tasks is more explicit. Models for adaptation to in-distribution tasks should emphasize upon mining similarity structure to solve few-shot unseen tasks more efficiently. 
Methods developed for out-of-distribution tasks should maximize model generalization ability. There exists a tradeoff between model fitting and model generalization. Base learner focuses upon adaptation to each task and meta-learner concentrates upon  generalization, which offer a flexible combination to strike a balance in between. 
We can obtain an estimate of task distribution and judge whether an unseen task is in-distribution or out-of-distribution. 

In terms of application, meta-learning presents 
an opportunity for improving model adaptivity across changing environments.
For theoretical research, meta-learning provides a flexible framework of integrating different methods combining advantages while complementing disadvantages of each other.
Statistical models are less prone to over-fitting and can be combined with machine learning methods under meta-learning framework.
Meta-learning allows us to partition tasks into levels, apply base learner, meta-learners and other learners at different levels constructing an integrated solution for a complex mission. Communication between base learner and meta-learners at different levels is not limited to initial values, aggregating gradients, objectives and learned optimizers. 
More complex model specifications can be designed to suit the need of real-life applications.

Meta-learning is not just about solving few-shot in-distribution or out-of-distribution tasks with deep learning models. However, with the amount of data available nowadays and wide applications of deep learning models, most application problems can be solved with sufficiently large amount of data and sufficiently complex deep models to obtain satisfactory prediction accuracy. We do not see as many few-shot tasks. Actually it depends on how we define 'few-shot'. With higher prediction accuracy, the number of cases where deep models fail is small. These failing cases constitute a 'few-shot' task. As mentioned earlier, with meta-learning framework or components embedded, predictive performance of deep models on in-distribution tasks is still high, and predictive performance of deep models on out-of-distribution tasks is improved. Meta-learning framework or components can usually accelerate training of deep models and improve performance of deep models on out-of-distribution tasks. 


What remains unclear about the objective of meta-learning is whether we should allow meta-learning framework to evolve by itself to more complex tasks and corresponding solution to complex tasks. It is a promising field, since we can actually solve complex tasks which are unsolvable by training from scratch. But uncontrolled evolution of meta-learning framework may create too high autonomy in machine intelligence bringing ethical concerns. From the perspective of optimization, meta-learning framework can be viewed as an optimization technique. What it presents seems to be solving complex tasks using genetic algorithm for finding global optimization. Evolution of complex tasks can be seen as a meta-learner accumulating problem-solving experience by solving easier and simpler tasks. Then meta-learner guides base learner to find a solution much closer to global optimum. This is actually very similar to application of meta-learning in AutoML where deep neural network model grows to solve more complex problems and reach higher prediction accuracy. In the end, we have a deep neural network model with highest prediction accuracy, even higher than human-designed/tuned deep neural network model. In the searching for the global optimum deep neural network model, current model with highest prediction accuracy is a moving target for meta-learner to beat. Task is taken to be beating current model with highest prediction accuracy. Task evolves to be more complex along the searching process of AutoML. Solution is taken to be a deep neural network model that can beat the current best model. With progressive evolution of task, solution also becomes more complex. As a result, AI-GAs can be viewed as AutoML in deep RL, where AutoML is for modelling policy in deep RL.

Meta-learning seeks solving vastly different few-shot tasks with deep learning models. 
Meta-learning is a flexible framework with clear objectives but no clear form of specific models or protocols. Many meta-learning modelling routes may be combined to form an integrated model. Meta-learning can also be integrated into other machine learning frameworks or deep learning models. This is not a paper about methodology classification, but a paper about introducing meta-learning frameworks and applications with famous meta-learning methods as examples. 

\section*{Acknowledgment}

Thanks to Debasmit Das, Louis Kirsch and Luca Bertinetto (in alphabetical order) for useful and valuable comments on this manuscript.

\bibliographystyle{unsrt}



\begin{thebibliography}{100}
	
	\bibitem{Schmidthuber}
	J{\"{u}}rgen Schmidthuber.
	\newblock {Evolutionary principles in self-referential learning (Diploma
		Thesis)}.
	\newblock 1987.
	
	\bibitem{Schmidhuber2015}
	J{\"{u}}rgen Schmidhuber.
	\newblock {On Learning to Think: Algorithmic Information Theory for Novel
		Combinations of Reinforcement Learning Controllers and Recurrent Neural World
		Models}.
	\newblock 2015.
	
	\bibitem{Gers1999}
	Felix~A. Gers, Jurgen Schmidhuber, and Fred Cummins.
	\newblock {Learning to forget: Continual prediction with LSTM}.
	\newblock {\em IEE Conference Publication}, 2(470):850--855, 1999.
	
	\bibitem{Younger2001}
	Steven Younger, Sepp Hochreiter, and Peter Conwell.
	\newblock {Meta-learning with backpropagation}.
	\newblock {\em Proceedings of the International Joint Conference on Neural
		Networks}, pages 2001--2006, 2001.
	
	\bibitem{Ravi2019a}
	Sachin Ravi and Hugo Larochelle.
	\newblock {Optimization as a model for few-shot learning}.
	\newblock In {\em 5th International Conference on Learning Representations,
		ICLR - Conference Track Proceedings}, pages 1--11, 2017.
	
	\bibitem{Clune2019}
	Jeff Clune.
	\newblock {AI-GAs: AI-generating algorithms, an alternate paradigm for
		producing general artificial intelligence}.
	\newblock 2019.
	
	\bibitem{Wang2019a}
	Rui Wang, Joel Lehman, Jeff Clune, and Kenneth~O. Stanley.
	\newblock {Paired open-ended trailblazer (POET): Endlessly generating
		increasingly complex and diverse learning environments and their solutions}.
	\newblock pages 1--28, 2019.
	
	\bibitem{Pan2010}
	Sinno~Jialin Pan and Qiang Yang.
	\newblock {A survey on transfer learning}.
	\newblock {\em IEEE Transactions on Knowledge and Data Engineering},
	22(10):1345--1359, 2010.
	
	\bibitem{Hutter2019}
	Frank Hutter.
	\newblock {\em {Automated machine learning}}.
	\newblock 2019.
	
	\bibitem{Yu2018}
	Tianhe Yu, Chelsea Finn, Annie Xie, Sudeep Dasari, Tianhao Zhang, Pieter
	Abbeel, and Sergey Levine.
	\newblock {One-shot imitation from observing humans via domain-adaptive
		meta-learning}.
	\newblock {\em 6th International Conference on Learning Representations, ICLR -
		Workshop Track Proceedings}, 2018.
	
	\bibitem{Guan2020}
	Jiechao Guan, Zhiwu Lu, Tao Xiang, and Ji-Rong Wen.
	\newblock {Few-Shot Learning as Domain Adaptation: Algorithm and Analysis}.
	\newblock 2020.
	
	\bibitem{Finn2017}
	Chelsea Finn, Pieter Abbeel, and Sergey Levine.
	\newblock {Model-agnostic meta-learning for fast adaptation of deep networks}.
	\newblock {\em 34th International Conference on Machine Learning, ICML},
	3:1856--1868, 2017.
	
	\bibitem{Koch2015}
	Gregory Koch and Gregory Koch.
	\newblock {\em {Siamese Thesis}}.
	\newblock PhD thesis, 2015.
	
	\bibitem{Vinyals2016}
	Oriol Vinyals, Charles Blundell, Timothy Lillicrap, Koray Kavukcuoglu, and Daan
	Wierstra.
	\newblock {Matching networks for one shot learning}.
	\newblock {\em Advances in Neural Information Processing Systems}, pages
	3637--3645, 2016.
	
	\bibitem{Snell2017}
	Jake Snell, Kevin Swersky, and Richard Zemel.
	\newblock {Prototypical networks for few-shot learning}.
	\newblock {\em Advances in Neural Information Processing Systems}, pages
	4078--4088, 2017.
	
	\bibitem{Li2017}
	Zhenguo Li, Fengwei Zhou, Fei Chen, and Hang Li.
	\newblock {Meta-SGD: Learning to Learn Quickly for Few-Shot Learning}.
	\newblock 2017.
	
	\bibitem{Sung2018}
	Flood Sung, Yongxin Yang, Li~Zhang, Tao Xiang, Philip~H.S. Torr, and Timothy~M.
	Hospedales.
	\newblock {Learning to compare: relation network for few-shot learning}.
	\newblock {\em Proceedings of the IEEE Computer Society Conference on Computer
		Vision and Pattern Recognition}, pages 1199--1208, 2018.
	
	\bibitem{Gordon2018}
	Jonathan Gordon, John Bronskill, Matthias Bauer, Sebastian Nowozin, and
	Richard~E Turner.
	\newblock {VERSA: Versatile and efficient few-shot learning}.
	\newblock {\em Advances in Neural Information Processing Systems}, pages 1--9,
	2018.
	
	\bibitem{Khodadadeh2018}
	Siavash Khodadadeh, Ladislau B{\"{o}}l{\"{o}}ni, and Mubarak Shah.
	\newblock {Unsupervised meta-learning for few-shot image classification}.
	\newblock 2018.
	
	\bibitem{Wang2019b}
	Yaqing Wang and Quanming Yao.
	\newblock {Few-shot Learning: A Survey}.
	\newblock {\em Arxiv}, pages 1--41, 2019.
	
	\bibitem{Munkhdalai2018}
	Tsendsuren Munkhdalai, Xingdi Yuan, Soroush Mehri, and Adam Trischler.
	\newblock {Rapid adaptation with conditionally shifted neurons}.
	\newblock {\em 35th International Conference on Machine Learning, ICML},
	8:5898--5909, 2018.
	
	\bibitem{Li2019}
	Ke~Li and Jitendra Malik.
	\newblock {Learning to optimize}.
	\newblock In {\em 5th International Conference on Learning Representations,
		ICLR - Conference Track Proceedings}, 2017.
	
	\bibitem{Munkhdalai2017}
	Tsendsuren Munkhdalai and Hong Yu.
	\newblock {Meta networks}.
	\newblock In {\em 34th International Conference on Machine Learning, ICML},
	volume~5, pages 3933--3943, 2017.
	
	\bibitem{Schweighofer2003}
	Nicolas Schweighofer and Kenji Doya.
	\newblock {Meta-learning in reinforcement learning}.
	\newblock {\em Neural Networks}, 16(1):5--9, 2003.
	
	\bibitem{Jaafra2018}
	Yesmina Jaafra, Jean~Luc Laurent, Aline Deruyver, and Mohamed~Saber Naceur.
	\newblock {A Review of Meta-Reinforcement Learning for Deep Neural Networks
		Architecture Search}.
	\newblock pages 1--29, 2018.
	
	\bibitem{Gupta2018}
	Abhishek Gupta, Benjamin Eysenbach, Chelsea Finn, and Sergey Levine.
	\newblock {Unsupervised Meta-Learning for Reinforcement Learning}.
	\newblock pages 1--15, 2018.
	
	\bibitem{Wang2018}
	Jane~X. Wang, Zeb Kurth-Nelson, Dharshan Kumaran, Dhruva Tirumala, Hubert
	Soyer, Joel~Z. Leibo, Demis Hassabis, and Matthew Botvinick.
	\newblock {Prefrontal cortex as a meta-reinforcement learning system}.
	\newblock {\em Nature Neuroscience}, 21(6):860--868, 2018.
	
	\bibitem{Humplik2019}
	Jan Humplik, Alexandre Galashov, Leonard Hasenclever, Pedro~A. Ortega, Yee~Whye
	Teh, and Nicolas Heess.
	\newblock {Meta reinforcement learning as task inference}.
	\newblock 2019.
	
	\bibitem{Rakelly2019}
	Kate Rakelly, Aurick Zhou, Deirdre Quiilen, Chelsea Finn, and Sergey Levine.
	\newblock {Efficient off-policy meta-reinforcement learning via probabilistic
		context variables}.
	\newblock {\em 36th International Conference on Machine Learning, ICML}, pages
	9291--9301, 2019.
	
	\bibitem{Dasgupta2019}
	Ishita Dasgupta, Jane Wang, Silvia Chiappa, Jovana Mitrovic, Pedro Ortega,
	David Raposo, Edward Hughes, Peter Battaglia, Matthew Botvinick, and Zeb
	Kurth-Nelson.
	\newblock {Causal Reasoning from Meta-reinforcement Learning}.
	\newblock 2019.
	
	\bibitem{Nagabandi2019}
	Anusha Nagabandi, Ignasi Clavera, Simin Liu, Ronald~S. Fearing, Pieter Abbeel,
	Sergey Levine, and Chelsea Finn.
	\newblock {Learning to adapt in dynamic, real-world environments through
		meta-reinforcement learning}.
	\newblock {\em 7th International Conference on Learning Representations, ICLR},
	pages 1--17, 2019.
	
	\bibitem{Duan2017}
	Yan Duan, Marcin Andrychowicz, Bradly Stadie, Jonathan Ho, Jonas Schneider,
	Ilya Sutskever, Pieter Abbeel, and Wojciech Zaremba.
	\newblock {One-shot imitation learning}.
	\newblock In {\em Advances in Neural Information Processing Systems}, pages
	1088--1099, 2017.
	
	\bibitem{Finn2019b}
	Chelsea Finn.
	\newblock {Meta-learning: from few-shot learning to rapid reinforcement
		learning}.
	\newblock {\em 36th International Conference on Machine Learning ICML Tutorial
		session}, 2019.
	
	\bibitem{Finn2017a}
	Chelsea Finn, Pieter Abbeel, and Sergey Levine.
	\newblock {Model-agnostic meta-learning for fast adaptation of deep networks}.
	\newblock {\em 34th International Conference on Machine Learning, ICML 2017},
	3:1856--1868, 2017.
	
	\bibitem{Zhou2019a}
	Allan Zhou, Eric Jang, Daniel Kappler, Alex Herzog, Mohi Khansari, Paul
	Wohlhart, Yunfei Bai, Mrinal Kalakrishnan, Sergey Levine, and Chelsea Finn.
	\newblock {Watch, Try, Learn: Meta-Learning from Demonstrations and Reward}.
	\newblock 2019.
	
	\bibitem{Russell1991}
	Stuart Russell and Eric Wefald.
	\newblock {Principles of metareasoning}.
	\newblock {\em Artificial Intelligence}, 49(1-3):361--395, 1991.
	
	\bibitem{Schmidhuber1994}
	J{\"{u}}rgen~H. Schmidhuber.
	\newblock {On Learning how to Learn Learning Strategies}.
	\newblock {\em Technical Report FKI-198-94}, pages 1--20, 1994.
	
	\bibitem{Schmidhuber1998}
	J{\"{u}}rgen Schmidhuber, Jieyu Zhao, and Nicol~N. Schraudolph.
	\newblock {Reinforcement Learning with Self-Modifying Policies}.
	\newblock {\em Learning to Learn}, pages 293--309, 1998.
	
	\bibitem{Schmidhuber1993}
	J{\"{u}}rgen Schmidhuber.
	\newblock {A Neural Network the Embeds its own Meta-Levels}.
	\newblock In {\em IEEE International Conference on Neural Networks}, pages
	407--412, 1993.
	
	\bibitem{Caruana1995}
	Rich Caruana.
	\newblock {Learning many related tasks at the same time with backpropagation}.
	\newblock Technical report, 1995.
	
	\bibitem{Bengio2000}
	Yoshua Bengio.
	\newblock {Gradient-based optimization of hyperparameters}.
	\newblock {\em Neural Computation}, 12(8):1889--1900, 2000.
	
	\bibitem{Jones2001}
	Donald~R. Jones.
	\newblock {A taxonomy of global optimization methods based on response
		surfaces}.
	\newblock {\em Journal of Global Optimization}, 21(4):345--383, 2001.
	
	\bibitem{Stanley2002}
	Kenneth~O. Stanley and Risto Miikkulainen.
	\newblock {Evolving neural networks through augmenting topologies}.
	\newblock {\em Evolutionary Computation}, 10(2):99--127, 2002.
	
	\bibitem{Pellerin2004}
	Eric Pellerin, Luc Pigeon, and Sylvain Delisle.
	\newblock {A meta-learning system based on genetic algorithms}.
	\newblock {\em Data Mining and Knowledge Discovery: Theory, Tools, and
		Technology VI}, 2004.
	
	\bibitem{Kordik2010}
	Pavel Kord{\'{i}}k, Jan Koutn{\'{i}}k, Jan Drchal, Oleg Kov{\'{a}}ř{\'{i}}k,
	Miroslav {\v{C}}epek, and Miroslav {\v{S}}norek.
	\newblock {Meta-learning approach to neural network optimization}.
	\newblock {\em Neural Networks}, 23(4):568--582, 2010.
	
	\bibitem{Bergstra2011}
	James Bergstra, Remi Bardenet, Yoshua Bengio, and Balazs Kegl.
	\newblock {Algorithms for hyper-parameter optimization}.
	\newblock {\em 25th Annual Conference on Neural Information Processing Systems
		(NIPS)}, 2011.
	
	\bibitem{Miranda2012}
	Pericles~B.C. Miranda, Ricardo~B.C. Prudencio, Andre Carlos~P.L.F. {De
		Carvalho}, and Carlos Soares.
	\newblock {Multi-objective optimization and Meta-learning for SVM parameter
		selection}.
	\newblock {\em Proceedings of the International Joint Conference on Neural
		Networks}, 2012.
	
	\bibitem{Bergstra2012}
	James Bergstra and Yoshua Bengio.
	\newblock {Random search for hyper-parameter optimization}.
	\newblock {\em Journal of Machine Learning Research}, 13:281--305, 2012.
	
	\bibitem{Maclaurin2015}
	Dougal Maclaurin, David Duvenaud, and Ryan~P. Adams.
	\newblock {Gradient-based hyperparameter optimization through reversible
		learning}.
	\newblock feb 2015.
	
	\bibitem{AliEslami2016}
	S.~M. {Ali Eslami}, Nicolas Heess, Theophane Weber, Yuval Tassa, David
	Szepesvari, Koray Kavukcuoglu, and Geoffrey~E. Hinton.
	\newblock {Attend, infer, repeat: Fast scene understanding with generative
		models}.
	\newblock {\em Advances in Neural Information Processing Systems}, pages
	3233--3241, 2016.
	
	\bibitem{DeSouto2008}
	Marcilio~C.P. {De Souto}, Ricardo~B.C. Prud{\^{e}}ncio, Rodrigo~G.F. Soares,
	Daniel~S.A. {De Araujo}, Ivan~G. Costa, Teresa~B. Ludermir, and Alexander
	Schliep.
	\newblock {Ranking and selecting clustering algorithms using a meta-learning
		approach}.
	\newblock {\em Proceedings of the International Joint Conference on Neural
		Networks}, pages 3729--3735, 2008.
	
	\bibitem{Lemke2010}
	Christiane Lemke and Bogdan Gabrys.
	\newblock {Meta-learning for time series forecasting and forecast combination}.
	\newblock {\em Neurocomputing}, 73(10-12):2006--2016, 2010.
	
	\bibitem{Leyva2015}
	Enrique Leyva, Antonio Gonz{\'{a}}lez, and Ra{\'{u}}l P{\'{e}}rez.
	\newblock {A set of complexity measures designed for applying meta-learning to
		instance selection}.
	\newblock {\em IEEE Transactions on Knowledge and Data Engineering},
	27(2):354--367, 2015.
	
	\bibitem{Bosc2016}
	Tom Bosc.
	\newblock {Learning to learn neural networks}.
	\newblock Technical report, 2016.
	
	\bibitem{Baker2019}
	Bowen Baker, Otkrist Gupta, Nikhil Naik, and Ramesh Raskar.
	\newblock {Designing neural network architectures using reinforcement
		learning}.
	\newblock {\em 5th International Conference on Learning Representations, ICLR-
		Conference Track Proceedings}, 2017.
	
	\bibitem{Real2020}
	Esteban Real, Chen Liang, David~R. So, and Quoc~V. Le.
	\newblock {AutoML-Zero: Evolving Machine Learning Algorithms From Scratch}.
	\newblock 2020.
	
	\bibitem{Liu2018c}
	Chenxi Liu, Barret Zoph, Maxim Neumann, Jonathon Shlens, Wei Hua, Li~Jia Li,
	Li~Fei-Fei, Alan Yuille, Jonathan Huang, and Kevin Murphy.
	\newblock {Progressive Neural Architecture Search}.
	\newblock {\em Lecture Notes in Computer Science (including subseries Lecture
		Notes in Artificial Intelligence and Lecture Notes in Bioinformatics)}, 11205
	LNCS:19--35, 2018.
	
	\bibitem{Weng2018}
	L~Weng.
	\newblock {Meta-Learning: Learning to Learn Fast}, 2018.
	
	\bibitem{Santoro2016a}
	Adam Santoro, Sergey Bartunov, Matthew Botvinick, Daan Wierstra, and Timothy
	Lillicrap.
	\newblock {Meta-Learning with Memory-Augmented Neural Networks}.
	\newblock {\em 33rd International Conference on Machine Learning, ICML 2016},
	4:2740--2751, 2016.
	
	\bibitem{Nichol2018}
	Alex Nichol, Joshua Achiam, and John Schulman.
	\newblock {On first-order meta-learning algorithms}.
	\newblock {\em arXiv}, pages 1--11, 2018.
	
	\bibitem{Lake2015}
	Brenden~M. Lake, Ruslan Salakhutdinov, and Joshua~B. Tenenbaum.
	\newblock {Human-level concept learning through probabilistic program
		induction}.
	\newblock {\em Science}, 350(6266):1332--1338, 2015.
	
	\bibitem{Yoon2018a}
	Jaesik Yoon, Taesup Kim, Ousmane Dia, Sungwoong Kim, Yoshua Bengio, and Sungjin
	Ahn.
	\newblock {Bayesian model-agnostic meta-learning}.
	\newblock In {\em Advances in Neural Information Processing Systems}, pages
	7332--7342, 2018.
	
	\bibitem{Duan2016}
	Yan Duan, John Schulman, Xi~Chen, Peter~L. Bartlett, Ilya Sutskever, and Pieter
	Abbeel.
	\newblock {RL{\^{}}2: Fast Reinforcement Learning via Slow Reinforcement
		Learning}.
	\newblock pages 1--14, 2016.
	
	\bibitem{Wang2016}
	Jane~X Wang, Zeb Kurth-Nelson, Dhruva Tirumala, Hubert Soyer, Joel~Z Leibo,
	Remi Munos, Charles Blundell, Dharshan Kumaran, and Matt Botvinick.
	\newblock {Learning to reinforcement learn}.
	\newblock pages 1--17, 2016.
	
	\bibitem{Levine2017}
	Nir Levine, Tom Zahavy, Daniel~J. Mankowitz, Aviv Tamar, and Shie Mannor.
	\newblock {Shallow updates for deep reinforcement learning}.
	\newblock {\em Advances in Neural Information Processing Systems}, pages
	3136--3146, 2017.
	
	\bibitem{Stadie2018}
	Bradly~C. Stadie, Ge~Yang, Rein Houthooft, Xi~Chen, Yan Duan, Yuhuai Wu, Pieter
	Abbeel, and Ilya Sutskever.
	\newblock {Some considerations on learning to explore via meta-reinforcement
		learning}.
	\newblock In {\em Advances in Neural Information Processing Systems}, pages
	9280--9290, 2018.
	
	\bibitem{Nagabandi2019a}
	Anusha Nagabandi, Chelsea Finn, and Sergey Levine.
	\newblock {Deep online learning via meta-learning: Continual adaptation for
		model-based RL}.
	\newblock In {\em 7th International Conference on Learning Representations,
		ICLR}, 2019.
	
	\bibitem{Lee2019a}
	Alex~X. Lee, Anusha Nagabandi, Pieter Abbeel, and Sergey Levine.
	\newblock {Stochastic latent actor-critic: Deep reinforcement learning with a
		latent variable model}.
	\newblock pages 1--19, 2019.
	
	\bibitem{Bengio2011}
	Yoshua Bengio.
	\newblock {Deep Learning of Representations for Unsupervised and Transfer
		Learning}, 2011.
	
	\bibitem{Garg2018}
	Vikas~K. Garg and Adam Kalai.
	\newblock {Supervising unsupervised learning}.
	\newblock {\em Advances in Neural Information Processing Systems}, pages
	4991--5001, 2018.
	
	\bibitem{Hsu2019}
	Kyle Hsu, Sergey Levine, and Chelsea Finn.
	\newblock {Unsupervised learning via meta-learning}.
	\newblock {\em 7th International Conference on Learning Representations, ICLR},
	2019.
	
	\bibitem{Metz2019}
	Luke Metz, Jascha Sohl-Dickstein, Niru Maheswaranathan, and Brian Cheung.
	\newblock {Meta-learning update rules for unsupervised representation
		learning}.
	\newblock {\em 7th International Conference on Learning Representations, ICLR},
	pages 1--27, 2019.
	
	\bibitem{Berthelot2019}
	David Berthelot, Ian Goodfellow, Colin Raffel, and Aurko Roy.
	\newblock {Understanding and improving interpolation in autoencoders via an
		adversarial regularizer}.
	\newblock {\em 7th International Conference on Learning Representations ICLR},
	2019.
	
	\bibitem{Caron2018}
	Mathilde Caron, Piotr Bojanowski, Armand Joulin, and Matthijs Douze.
	\newblock {Deep clustering for unsupervised learning of visual features}.
	\newblock In {\em European Conference on Computer Vision (ECCV)}, pages
	139--156, 2018.
	
	\bibitem{Chen2016}
	Xi~Chen, Yan Duan, Rein Houthooft, John Schulman, Ilya Sutskever, and Pieter
	Abbeel.
	\newblock {InfoGAN: Interpretable representation learning by information
		maximizing generative adversarial nets}.
	\newblock {\em Advances in Neural Information Processing Systems}, pages
	2180--2188, 2016.
	
	\bibitem{Harrison2018}
	James Harrison, Apoorva Sharma, and Marco Pavone.
	\newblock {Meta-Learning Priors for Efficient Online Bayesian Regression}.
	\newblock 2018.
	
	\bibitem{Finn2019}
	Chelsea Finn, Aravind Rajeswaran, Sham Kakade, and Sergey Levine.
	\newblock {Online Meta-Learning}.
	\newblock 2019.
	
	\bibitem{Triantafillou2019a}
	Eleni Triantafillou, Tyler Zhu, Vincent Dumoulin, Pascal Lamblin, Utku Evci,
	Kelvin Xu, Ross Goroshin, Carles Gelada, Kevin Swersky, Pierre-Antoine
	Manzagol, and Hugo Larochelle.
	\newblock {Meta-dataset: A dataset of datasets for learning to learn from few
		examples}.
	\newblock 2019.
	
	\bibitem{Lake2011a}
	Brenden~M Lake, Ruslan Salakhutdinov, Jason Gross, and Joshua~B Tenenbaum.
	\newblock {One shot learning of simple visual concepts}.
	\newblock {\em In {\{}Proceedings of the 33rd Annual Conference of the
		Cognitive Science Society{\}}}, 2011.
	
	\bibitem{Russakovsky2015}
	Olga Russakovsky, Jia Deng, Hao Su, Jonathan Krause, Sanjeev Satheesh, Sean Ma,
	Zhiheng Huang, Andrej Karpathy, Aditya Khosla, Michael Bernstein,
	Alexander~C. Berg, and Li~Fei-Fei.
	\newblock {ImageNet Large Scale Visual Recognition Challenge}.
	\newblock {\em International Journal of Computer Vision}, 115(3):211--252,
	2015.
	
	\bibitem{Lee2019}
	Kwonjoon Lee, Subhransu Maji, Avinash Ravichandran, and Stefano Soatto.
	\newblock {Meta-learning with differentiable convex optimization}.
	\newblock {\em Proceedings of the IEEE Computer Society Conference on Computer
		Vision and Pattern Recognition}, 2019-June:10649--10657, 2019.
	
	\bibitem{Ren2018}
	Mengye Ren, Eleni Triantafillou, Sachin Ravi, Jake Snell, Kevin Swersky,
	Joshua~B. Tenenbaum, Hugo Larochelle, and Richard~S. Zemel.
	\newblock {Meta-learning for semi-supervised few-shot classification}.
	\newblock {\em 6th International Conference on Learning Representations, ICLR -
		Conference Track Proceedings}, pages 1--15, 2018.
	
	\bibitem{Krizhevsky2010}
	Alex Krizhevsky.
	\newblock {Convolutional deep belief networks on CIFAR-10}.
	
	\bibitem{Oreshkin2018}
	Boris~N. Oreshkin, Pau Rodriguez, and Alexandre Lacoste.
	\newblock {Tadam: Task dependent adaptive metric for improved few-shot
		learning}.
	\newblock In {\em Advances in Neural Information Processing Systems}, pages
	721--731, 2018.
	
	\bibitem{Bertinetto2019}
	Luca Bertinetto, Philip~H.S. Torr, Jo{\~{a}}o Henriques, and Andrea Vedaldi.
	\newblock {Meta-learning with differentiable closed-form solvers}.
	\newblock {\em 7th International Conference on Learning Representations, ICLR},
	pages 1--15, 2019.
	
	\bibitem{Santorini1993}
	M.~Marcus, B.~Santorini, and M.~Marcinkiewicz.
	\newblock {Building a Large Annotated Corpus of English: The Penn Treebank}.
	\newblock {\em Computational linguistics}, 19(2):313, 1993.
	
	\bibitem{Welinder2010}
	Peter Welinder, Steve Branson, Takeshi Mita, Catherine Wah, and Florian
	Schroff.
	\newblock {Caltech-ucsd Birds 200}.
	\newblock {\em Caltech-UCSD Technical Report}, pages 1--15, 2010.
	
	\bibitem{Wichrowska2017}
	Olga Wichrowska, Niru Maheswaranathan, Matthew~W. Hoffman, Sergio~G{\'{o}}mez
	Colmenarejo, Misha Denii, Nando {De Freitas}, and Jascha Sohl-Dickstein.
	\newblock {Learned optimizers that scale and generalize}.
	\newblock {\em 34th International Conference on Machine Learning, ICML}, pages
	5744--5753, 2017.
	
	\bibitem{Triantafillou2017}
	Eleni Triantafillou, Richard Zemel, and Raquel Urtasun.
	\newblock {Few-shot learning through an information retrieval lens}.
	\newblock {\em Advances in Neural Information Processing Systems}, pages
	2256--2266, 2017.
	
	\bibitem{Domhan2015}
	Tobias Domhan, Jost~Tobias Springenberg, and Frank Hutter.
	\newblock {Speeding up automatic hyperparameter optimization of deep neural
		networks by extrapolation of learning curves}.
	\newblock {\em IJCAI International Joint Conference on Artificial
		Intelligence}, 2015-Janua(Ijcai):3460--3468, 2015.
	
	\bibitem{Baker2018}
	Bowen Baker, Otkrist Gupta, Ramesh Raskar, and Nikhil Naik.
	\newblock {Accelerating neural architecture search using performance
		prediction}.
	\newblock {\em 6th International Conference on Learning Representations, ICLR
		2018 - Workshop Track Proceedings}, 2, 2018.
	
	\bibitem{Kaiser2019}
	Lukasz Kaiser, Aurko Roy, Ofir Nachum, and Samy Bengio.
	\newblock {Learning to remember rare events}.
	\newblock {\em 5th International Conference on Learning Representations, ICLR
		2017 - Conference Track Proceedings}, 2019.
	
	\bibitem{Qiao2018}
	Siyuan Qiao, Chenxi Liu, Wei Shen, and Alan Yuille.
	\newblock {Few-Shot Image Recognition by Predicting Parameters from
		Activations}.
	\newblock {\em Proceedings of the IEEE Computer Society Conference on Computer
		Vision and Pattern Recognition}, pages 7229--7238, 2018.
	
	\bibitem{Mishra2018}
	Nikhil Mishra, Mostafa Rohaninejad, Xi~Chen, and Pieter Abbeel.
	\newblock {A simple neural attentive meta-learner}.
	\newblock {\em 6th International Conference on Learning Representations, ICLR -
		Conference Track Proceedings}, 2018.
	
	\bibitem{Li2020}
	Aoxue Li, Weiran Huang, Xu~Lan, Jiashi Feng, Zhenguo Li, and Liwei Wang.
	\newblock {Boosting Few-Shot Learning With Adaptive Margin Loss}.
	\newblock 2020.
	
	\bibitem{Gidaris2018}
	Spyros Gidaris and Nikos Komodakis.
	\newblock {Dynamic Few-Shot Visual Learning Without Forgetting}.
	\newblock In {\em Proceedings of the IEEE Computer Society Conference on
		Computer Vision and Pattern Recognition}, pages 4367--4375, 2018.
	
	\bibitem{Nichol2018a}
	Alex Nichol and John Schulman.
	\newblock {Reptile: a scalable metalearning algorithm}.
	\newblock {\em arXiv}, pages 1--11, 2018.
	
	\bibitem{Liu2019}
	Yanbin Liu, Juho Lee, Minseop Park, Saehoon Kim, Eunho Yang, Sung~Ju Hwang, and
	Yi~Yang.
	\newblock {Learning to propagate labels: Transductive propagation network for
		few-shot learning}.
	\newblock {\em 7th International Conference on Learning Representations, ICLR},
	pages 1--11, 2019.
	
	\bibitem{Rusu2019}
	Andrei~A. Rusu, Dushyant Rao, Jakub Sygnowski, Oriol Vinyals, Razvan Pascanu,
	Simon Osindero, and Raia Hadsell.
	\newblock {Meta-learning with latent embedding optimization}.
	\newblock {\em 7th International Conference on Learning Representations, ICLR},
	pages 1--17, 2019.
	
	\bibitem{Edwards2016}
	Harrison Edwards and Amos Storkey.
	\newblock {Towards a neural statistician}.
	\newblock pages 1--13, 2016.
	
	\bibitem{Grant2018}
	Erin Grant, Chelsea Finn, Sergey Levine, Trevor Darrell, and Thomas Griffiths.
	\newblock {Recasting gradient-based meta-learning as hierarchical bayes}.
	\newblock {\em 6th International Conference on Learning Representations, ICLR -
		Conference Track Proceedings}, pages 1--13, 2018.
	
	\bibitem{Feurer2015}
	Matthias Feurer, Aaron Klein, Katharina Eggensperger, Jost~Tobias Springenberg,
	Manuel Blum, and Frank Hutter.
	\newblock {Efficient and robust automated machine learning}.
	\newblock {\em Advances in Neural Information Processing Systems},
	2015-Janua:2962--2970, 2015.
	
	\bibitem{Xing2019}
	Chen Xing, Negar Rostamzadeh, Boris~N. Oreshkin, and Pedro~O. Pinheiro.
	\newblock {Adaptive Cross-Modal Few-Shot Learning}.
	\newblock 2019.
	
	\bibitem{Zagoruyko2016}
	Sergey Zagoruyko and Nikos Komodakis.
	\newblock {Wide Residual Networks}.
	\newblock {\em British Machine Vision Conference 2016, BMVC 2016},
	2016-Septe:87.1--87.12, 2016.
	
	\bibitem{Das2020}
	Debasmit Das and C.~S.George Lee.
	\newblock {A Two-Stage Approach to Few-Shot Learning for Image Recognition}.
	\newblock {\em IEEE Transactions on Image Processing}, 29:3336--3350, 2020.
	
	\bibitem{Bock1988}
	R~Vilalta and Y~Drissi.
	\newblock {A perspective view and survey of meta-learning}.
	\newblock {\em Artificial Intelligence Review}, pages 77--95, 2002.
	
	\bibitem{Li2019a}
	Jeffrey Li, Mikhail Khodak, Sebastian Caldas, and Ameet Talwalkar.
	\newblock {Differentially Private Meta-Learning}.
	\newblock 2019.
	
	\bibitem{Li2020a}
	Debang Li, Junge Zhang, and Kaiqi Huang.
	\newblock {Learning to Learn Cropping Models for Different Aspect Ratio
		Requirements}.
	\newblock {\em Cvpr}, 2020.
	
	\bibitem{Fallah2019}
	Alireza Fallah, Aryan Mokhtari, and Asuman Ozdaglar.
	\newblock {On the Convergence Theory of Gradient-Based Model-Agnostic
		Meta-Learning Algorithms}.
	\newblock 2019.
	
	\bibitem{Liu2019a}
	Hao Liu, Richard Socher, and Caiming Xiong.
	\newblock {Taming MAML: Efficient unbiased meta-reinforcement learning}.
	\newblock {\em 36th International Conference on Machine Learning, ICML 2019},
	2019-June:7156--7169, 2019.
	
	\bibitem{Song2019}
	Xingyou Song, Wenbo Gao, Yuxiang Yang, Krzysztof Choromanski, Aldo Pacchiano,
	and Yunhao Tang.
	\newblock {ES-MAML: Simple Hessian-Free Meta Learning}.
	\newblock 2019.
	
	\bibitem{Rothfuss2019}
	Jonas Rothfuss, Tamim Asfour, Dennis Lee, Ignasi Clavera, and Pieter Abbeel.
	\newblock {PrOMP: Proximal meta-policy search}.
	\newblock In {\em 7th International Conference on Learning Representations,
		ICLR}, 2019.
	
	\bibitem{Guo2020}
	Jianzhu Guo, Xiangyu Zhu, Chenxu Zhao, Dong Cao, Zhen Lei, and Stan~Z. Li.
	\newblock {Learning Meta Face Recognition in Unseen Domains}.
	\newblock 2020.
	
	\bibitem{Choi2020}
	Myungsub Choi, Janghoon Choi, Sungyong Baik, Tae~Hyun Kim, and Kyoung~Mu Lee.
	\newblock {Scene-Adaptive Video Frame Interpolation via Meta-Learning}.
	\newblock 2020.
	
	\bibitem{Antoniou2019}
	Antreas Antoniou, Amos Storkey, and Harrison Edwards.
	\newblock {How to train your MAML}.
	\newblock In {\em 7th International Conference on Learning Representations,
		ICLR}, 2019.
	
	\bibitem{Lee2019c}
	Jessica Lee, Deva Ramanan, and Rohit Girdhar.
	\newblock {MetaPix: Few-Shot Video Retargeting}.
	\newblock 2019.
	
	\bibitem{Wang2018a}
	Ting~Chun Wang, Ming~Yu Liu, Jun~Yan Zhu, Andrew Tao, Jan Kautz, and Bryan
	Catanzaro.
	\newblock {High-Resolution Image Synthesis and Semantic Manipulation with
		Conditional GANs}.
	\newblock {\em Proceedings of the IEEE Computer Society Conference on Computer
		Vision and Pattern Recognition}, pages 8798--8807, 2018.
	
	\bibitem{NESTEROV1983}
	Y.~NESTEROV.
	\newblock {A method for solving the convex programming problem with convergence
		rate O(1/k{\^{}}2)}.
	\newblock {\em American Mathematical Society}, 27(2):372--376, 1983.
	
	\bibitem{Duchi2010}
	John Duchi, Elad Hazan, and Yoram Singer.
	\newblock {Adaptive subgradient methods for online learning and stochastic
		optimization}.
	\newblock Technical report, 2010.
	
	\bibitem{Zeiler2012}
	Matthew~D. Zeiler.
	\newblock {ADADELTA: An Adaptive Learning Rate Method}.
	\newblock dec 2012.
	
	\bibitem{Kingma2015}
	Diederik~P. Kingma and Jimmy~Lei Ba.
	\newblock {Adam: A method for stochastic optimization}.
	\newblock In {\em 3rd International Conference on Learning Representations,
		ICLR - Conference Track Proceedings}. International Conference on Learning
	Representations, ICLR, 2015.
	
	\bibitem{SchmidhuberIdsia1997}
	J{\"{u}}rgen Schmidhuber.
	\newblock {Discovering Neural Nets with Low Kolmogorov Complexity and High
		Generalization Capability}.
	\newblock {\em Neural Networks}, 10(5):857--873, 1997.
	
	\bibitem{Kim2018}
	Jaehong Kim, Sangyeul Lee, Sungwan Kim, Moonsu Cha, Jung~Kwon Lee, Youngduck
	Choi, Yongseok Choi, Dong-Yeon Cho, and Jiwon Kim.
	\newblock {Auto-meta: Automated gradient based meta learner search}.
	\newblock pages 1--8, 2018.
	
	\bibitem{Liu2016}
	Qiang Liu and Dilin Wang.
	\newblock {Stein variational gradient descent: A general purpose Bayesian
		inference algorithm}.
	\newblock {\em Advances in Neural Information Processing Systems}, pages
	2378--2386, 2016.
	
	\bibitem{Liu2017}
	Yang Liu, Prajit Ramachandran, Qiang Liu, and Jian Peng.
	\newblock {Stein variational policy gradient}.
	\newblock {\em Uncertainty in Artificial Intelligence - Proceedings of the 33rd
		Conference, UAI 2017}, 2017.
	
	\bibitem{Finn2018}
	Chelsea Finn, Kelvin Xu, and Sergey Levine.
	\newblock {Probabilistic model-agnostic meta-learning}.
	\newblock 2018.
	
	\bibitem{Schmidhuber2003}
	J{\"{u}}rgen Schmidhuber.
	\newblock {Goedel Machines: Self-Referential Universal Problem Solvers Making
		Provably Optimal Self-Improvements}.
	\newblock 2003.
	
	\bibitem{Yao1993}
	Xin Yao.
	\newblock {A Review of Evolutionary Artificial Neural Networks}.
	\newblock {\em International Journal of Intelligent Systems}, 4:203--222, 1993.
	
	\bibitem{Stadie2018a}
	Bradly Stadie, Ge~Yang, Rein Houthooft, Xi~Chen, Yan Duan, Yuhuai Wu, Pieter
	Abbeel, and Hya Sutskever.
	\newblock {The importance of sampling in meta-reinforcement learning}.
	\newblock {\em 32nd Conference on Neural Information Processing Systems (NIPS),
		Montreal, Canada}, 2018.
	
	\bibitem{Fakoor2020}
	Rasool Fakoor, Pratik Chaudhari, Stefano Soatto, and Alexander Smola.
	\newblock {Meta-Q-Learning}.
	\newblock In {\em ICLR}, 2020.
	
	\bibitem{Brockman2016}
	Greg Brockman, Vicki Cheung, Ludwig Pettersson, Jonas Schneider, John Schulman,
	Jie Tang, and Wojciech Zaremba.
	\newblock {OpenAI gym}.
	\newblock jun 2016.
	
	\bibitem{Haarnoja2018}
	Tuomas Haarnoja, Aurick Zhou, Pieter Abbeel, and Sergey Levine.
	\newblock {Soft actor-critic: Off-policy maximum entropy deep reinforcement
		learning with a stochastic actor}.
	\newblock {\em 35th International Conference on Machine Learning, ICML 2018},
	5:2976--2989, 2018.
	
	\bibitem{Kirsch2019}
	Louis Kirsch, Sjoerd van Steenkiste, and J{\"{u}}rgen Schmidhuber.
	\newblock {Improving Generalization in Meta Reinforcement Learning using
		Learned Objectives}.
	\newblock 2019.
	
	\bibitem{Silver2014}
	David Silver, Guy Lever, Nicolas Heess, Thomas Degris, Daan Wierstra, and
	Martin Riedmiller.
	\newblock {Deterministic policy gradient algorithms}.
	\newblock {\em 31st International Conference on Machine Learning, ICML 2014},
	1:605--619, 2014.
	
	\bibitem{Lillicrap2016}
	Timothy Lillicrap, Jonathan Hunt, Alexander Pritzel, Nicolas Heess, Tom Erez,
	Yuval Tassa, David Silver, and Daan Wierstra.
	\newblock {Continuous Control with Deep Reinforcement Learning}.
	\newblock {\em ICLR}, 2016.
	
	\bibitem{Schmidhuber1990}
	J{\"{u}}rgen Schmidhuber.
	\newblock {Making the World Differentiable : On Using Self-Supervised Fully
		Recurrent Neural Networks for Dynamic Reinforcement Learning and Planning in
		Non-Stationary Environments ( TR FKI-126-90 )}.
	\newblock {\em Neural Networks}, pages 1--26, 1990.
	
	\bibitem{Alet2020}
	Ferran Alet, Martin~F. Schneider, Tomas Lozano-Perez, and Leslie~Pack
	Kaelbling.
	\newblock {Meta-learning curiosity algorithms}.
	\newblock 2020.
	
	\bibitem{Schmidhuber1999}
	J{\"{u}}rgen Schmidhuber.
	\newblock {Artificial curiosity based on discovering novel algorithmic
		predictability through coevolution}.
	\newblock {\em Proceedings of the 1999 Congress on Evolutionary Computation,
		CEC 1999}, 3:1612--1618, 1999.
	
	\bibitem{Whiteson2005}
	Shimon Whiteson, Nate Kohl, Risto Miikkulainen, and Peter Stone.
	\newblock {Evolving soccer keepaway players through task decomposition}.
	\newblock {\em Machine Learning}, 59(1-2):5--30, 2005.
	
	\bibitem{Barrett1994}
	Anthony Barrett and Daniel~S. Weld.
	\newblock {Task-decomposition via plan parsing}.
	\newblock {\em Proceedings of the National Conference on Artificial
		Intelligence}, 2:1117--1122, 1994.
	
	\bibitem{Shiarlis2018}
	Kyriacos Shiarlis, Markus Wulfmeier, Sasha Salter, Shimon Whiteson, and Ingmar
	Posner.
	\newblock {TACO: Learning task decomposition via temporal alignment for
		control}.
	\newblock {\em 35th International Conference on Machine Learning, ICML 2018},
	10:7419--7431, 2018.
	
	\bibitem{Utgoff2002}
	P.~E. Utgoff and D.~J. Stracuzzi.
	\newblock {Many-layered learning}.
	\newblock {\em Proceedings - 2nd International Conference on Development and
		Learning, ICDL 2002}, pages 141--146, 2002.
	
	\bibitem{Sohn2020}
	Sungryull Sohn, Hyunjae Woo, Jongwook Choi, and Honglak Lee.
	\newblock {Meta Reinforcement Learning with Autonomous Inference of Subtask
		Dependencies}.
	\newblock 2020.
	
	\bibitem{Paine2018}
	Tom~Le Paine, Sergio~G{\'{o}}mez Colmenarejo, Ziyu Wang, Scott Reed, Yusuf
	Aytar, Tobias Pfaff, Matt~W. Hoffman, Gabriel Barth-Maron, Serkan Cabi, David
	Budden, and Nando de~Freitas.
	\newblock {One-shot high-fidelity imitation: Training large-scale deep nets
		with RL}.
	\newblock 2018.
	
	\bibitem{Rabinowitz2018}
	Neil~C. Rabinowitz, Frank Perbet, H.~Francis Song, Chiyuan Zhang, and Matthew
	Botvinick.
	\newblock {Machine Theory of mind}.
	\newblock {\em 35th International Conference on Machine Learning, ICML 2018},
	10:6723--6738, 2018.
	
	\bibitem{Wu2019}
	Mike Wu, Kristy Choi, Noah Goodman, and Stefano Ermon.
	\newblock {Meta-Amortized Variational Inference and Learning}.
	\newblock 2019.
	
\end{thebibliography}

\end{document}